\documentclass{article}
\usepackage[preprint, nonatbib]{neurips_2022}
\usepackage{slantsc}
\usepackage{amsfonts}
\usepackage[utf8]{inputenc} 
\usepackage[T1]{fontenc}    
\usepackage{hyperref}       
\usepackage{url}            
\usepackage{booktabs}       
\usepackage{amsfonts}       
\usepackage{nicefrac}       
\usepackage{microtype}      
\usepackage{xcolor}         
\usepackage{authblk} 
\usepackage{enumitem}
\usepackage{pifont} 
\usepackage[maxbibnames=99]{biblatex}
\usepackage{textcomp}
\usepackage{amsmath}
\usepackage{amssymb}
\usepackage{multirow}
\usepackage{makecell}

\usepackage{caption}
\renewcommand{\raggedright}{\leftskip=0pt \rightskip=0pt plus 0cm}

\usepackage{amsmath}

\usepackage{graphicx}
\usepackage{float} 
\usepackage{subcaption}

\usepackage{hyperref}
\usepackage{array}
\usepackage{changepage}
\usepackage{multirow}
\newcommand{\specialcell}[2][c]{
  \begin{tabular}[#1]{@{}c@{}}#2\end{tabular}}
\def\eg{\emph{e.g.}} 
\def\ie{\emph{i.e}} 
\newcommand{\Tref}[1]{Table~\ref{#1}}

\newcommand{\Fref}[1]{Fig.~\ref{#1}}
\newcommand{\Sref}[1]{Sec.~\ref{#1}}

\newcommand{\eref}[1]{Eq.~(\ref{#1})}
\newcommand{\fref}[1]{Fig.~\ref{#1}}
\newcommand{\sref}[1]{Sec.~\ref{#1}}

\newcommand*{\affaddr}[1]{#1} 
\newcommand*{\affmark}[1][*]{\textsuperscript{#1}}
\newcommand*{\email}[1]{\texttt{#1}}
\usepackage{colortbl}

\title{NeIF: Representing General Reflectance as Neural Intrinsics Fields for Uncalibrated Photometric Stereo}
\author{%
\textbf{Zongrui~Li\affmark[1]}, \textbf{Qian~Zheng\affmark[2]}, \textbf{Feishi~Wang\affmark[3]}, \textbf{Boxin~Shi\affmark[3,4]}, \textbf{Gang Pan\affmark[2]}, \textbf{Xudong Jiang\affmark[1]}\\
\affaddr{\affmark[1]School of Electrical and Electronic Engineering, Nanyang Technological University, Singapore}\\
\affaddr{\affmark[2]College of Computer Science and Technology, Zhejiang University, Hangzhou, China}\\
\affaddr{\affmark[3]School of Computer Science, Peking University, Beijing, China}\\
\affaddr{\affmark[4]Institute for Artificial Intelligence, Peking University, Beijing, China} \\
\email{\{zongrui001,EXDJiang\}@ntu.edu.sg, \{qianzheng, gpan\}@zju.edu.cn, \{wangfeishi,shiboxin\}@pku.edu.cn}\\
}
\begin{document}
\maketitle
\begin{abstract}
Uncalibrated photometric stereo (UPS) is challenging due to the inherent ambiguity brought by unknown light. 
Existing solutions alleviate the ambiguity by either explicitly associating reflectance to light conditions or resolving light conditions in a supervised manner.
This paper establishes an implicit relation between light clues and light estimation and solves UPS in an unsupervised manner.
The key idea is to represent the reflectance as four neural intrinsics fields, \ie, position, light, specular, and shadow, based on which the neural light field is implicitly associated with light clues of specular reflectance and cast shadow. 
The unsupervised, joint optimization of neural intrinsics fields can be free from training data bias as well as accumulating error, and fully exploits all observed pixel values for UPS.
Our method achieves a superior performance advantage over state-of-the-art UPS methods on public and self-collected datasets, under regular and challenging setups. The code will be released soon.
\end{abstract}

\section{Introduction}
\label{tag:introduction}
Photometric stereo (PS)~\cite{woodham1980} aims at recovering the surface normal from several light-varying images captured at a fixed viewpoint. As compared with other approaches (\eg, multi-view stereo~\cite{seitz2006comparison}, active sensor-based solutions~\cite{zhang2002novel}), photometric stereo is excellent at recovering fine-detailed surfaces and has been widely used for Hollywood movies \cite{chabert2006relighting}, industrial quality inspection \cite{weigl2015photometric}, and biometrics \cite{xie2013real}. Calibrating accurate lighting directions is crucial to the performance of photometric stereo methods \cite{xie2015practical}. However, lighting calibration is often tedious, dramatically restricting the applicability in the real-world. To this end, researchers develop uncalibrated photometric stereo (UPS) methods that estimate surface normal with unknown lights. 

Uncalibrated photometric stereo suffers from  General Bas-Relief (GBR) ambiguity\footnote{In Lambertian objects,  there is an inherent 9-parameter ambiguity in the surface normals and light source directions. In the case when the integrability constraint is applied to the surface normals, the ambiguity is reduced to the 3-parameter generalized bas-relief (GBR) ambiguity~\mbox{\cite{georghiades2003incorporating}}.}\cite{belhumeur1999} for an integrable surface.
Early solutions address the ambiguity by explicitly associating reflectance to light, \ie, adopting analytic reflectance models (\eg, Lambertian reflectance\footnote{Lambertian reflectance is the property that defines an ideal diffusely reflecting surface~\mbox{\cite{Koppal2020}}.} \cite{alldrin2007resolving,papadhimitri2014closed}, parametric specular reflection \cite{georghiades2003incorporating}, specular spikes \cite{yeung2014normal}, inter-reflection \cite{chandraker2005reflections}) or imposing priors from reflectance properties \cite{alldrin2007toward,holroyd2008photometric,higo2010consensus,shi2012elevation}. 
\begin{figure}[t]
    \centering
    \includegraphics[width=0.85\textwidth]{"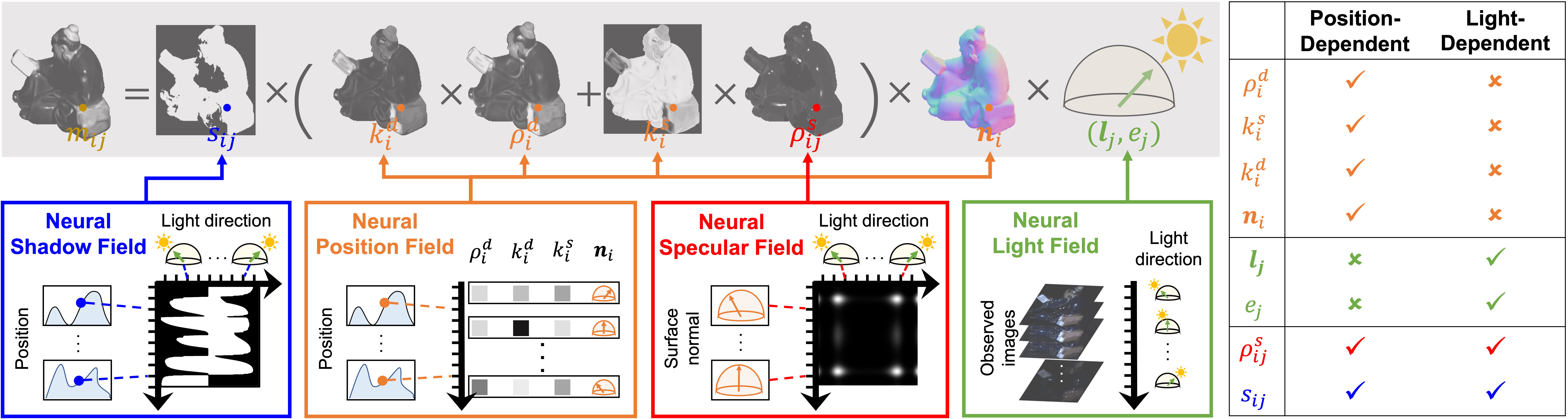"}
    \caption{Illustration of our neural intrinsics fields. Left-top: the rendering equation. Left-bottom: our four neural intrinsic fields, \ie, from left to right: shadow, position, specular, and light fields, respectively. Each sub-figure on the left-bottom illustrates the mutual information across dimensions of position-light, position, normal-light, and observed images, respectively. The left figure shows how the neural intrinsics fields are imposed to render a pixel. Right: a summary of our intrinsics w.r.t. the dependence on position or light. The definition of notations can be found from \eref{eq:general_brdf} and \eref{eq:field}. } 
    \label{fig:teaser}
    \vspace{-1em}
\end{figure}
Thus, due to the strong reliance on reflectance assumption, these methods can be less effective for unknown reflectance. Besides, these methods ignore the clue of cast shadow and even fail in shadow regions due to the shadow-free surface assumption. Further, most of them require the light intensity to be identical for robust estimation. Recently, deep learning-based approaches address the ambiguity by estimating light direction and intensity before recovering surface normal~\cite{chen2019self,chen2020learned,kaya2021uncalibrated,sarno2022neural}. They train a light estimation network using a large-scale amount of rendered data in a supervised manner. However, the training data bias \cite{mac2002problem} can hardly be eliminated
Because rendered training data inevitably contains the domain gap from the real ones and scarcely covers all surfaces with different geometry and reflectance in the real-world. Besides, such a two-step solution can bring accumulating errors for surface normal estimation. Further, all these methods assume the light intensity to distribute in a pre-defined range (\ie, $[0.2, 2]$). Training data bias, accumulating errors, and pre-defined light intensity range make previous PS methods sensitive to the lab setup of light, {\it i.e.}, if the setup slightly violates the light assumption ({\it e.g.}, non-ideal directional light, with little ambient light, out the range of pre-defined intensity), results of previous supervised PS methods degrades significantly.

To this end, we propose {\bf NeIF}, representing general reflectance as {\bf Ne}ural {\bf I}ntrinsics {\bf F}ields for uncalibrated photometric stereo. Our method differs from previous methods in three aspects: 1) it fully considers clues of specular reflectance and cast shadow from each observed pixel for light estimation so that it is expected to produce accurate estimation for both light conditions and surface normal;
2) it does not make explicit assumptions about the absolute range of reflectance or light so that it works with general surface reflectance and flexible light settings;
3) it infers light and surface normal in an unsupervised and joint manner. Therefore, it can better balance the accuracy between estimated light and surface normal even if the light assumption is not strictly satisfied, which is expected to facilitate the PS method to casual users.
Our key idea is to represent the general reflectance as four neural intrinsics fields (\ie, position, light, specular, and shadow, see \fref{fig:teaser}), implemented by four multi-layer perceptrons (MLPs). These four fields are connected based on the implicit relation (or dependence) of these intrinsics, optimized by reconstructing each pixel value from observed images which fully exploit mutual information across different dimensions, as shown in \fref{fig:teaser}.
The reconstruction error is backpropagated to the neural light field through neural specular and shadow fields so that clues of specular and shadow can be implicitly and fully considered for light estimation. 
Our contributions are summarized as:
\begin{itemize}[itemsep=0pt,topsep=-1pt,parsep=0pt]
    \item We represent general reflectance as four intrinsic neural fields to implicitly associate per-pixel reflectance to light, which solves uncalibrated photometric stereo by fully considering clues of specular reflectance and cast shadow for light estimation.
    \item We propose the NeIF, an uncalibrated photometric stereo method trained in an unsupervised and joint optimization manner, which works with general surface reflectance and flexible light settings, and is free from training data bias and accumulating errors. 
    \item We show that our method achieves superior performance over uncalibrated and unsupervised methods. We also demonstrate its excellent generalization capacity to data from different sources and robust performance with the challenging setup of sparse uncalibrated photometric stereo and casual scenarios.
\end{itemize}

\section{Related Work}
This section mainly reviews the latest works in neural reflectance representation and related works on unsupervised PS methods, UPS methods, and intrinsic decomposition. Readers may refer to~\mbox{\cite{xie2013real,shi2016benchmark,chen2019self,garces2022survey}} for a more comprehensive summary of neural reflectance representation, PS methods, UPS methods, and intrinsic decomposition, respectively.

\textbf{Neural Reflectance Representation in Multi-view Stereo.}
Recently, Neural Radiance Fields (NeRFs)~\mbox{\cite{mildenhall2020nerf}} implicitly represent the full 3D geometries and the reflectance of the objects through MLPs. Many subsequent works explore its application in various computer vision problems, such as 3D reconstruction~\mbox{\cite{mildenhall2020nerf,verbin2021ref,wang2021neus,oechsle2021unisurf,Srinivasan_2021_CVPR,zhang2021nerfactor,Zhang_2021_CVPR,Boss_2021_ICCV}} and relighting~\mbox{\cite{Srinivasan_2021_CVPR,zhang2021nerfactor,Zhang_2021_CVPR}}. 
While NeRF~\mbox{\cite{mildenhall2020nerf}} applies volume rendering~\mbox{\cite{kajiya1984ray}} to optimize the MLPs through reconstruction loss, they often end up with `fake reflectance'~\mbox{\cite{verbin2021ref}}(\ie, the emitters inside the objects contribute to specific reflectance effects) and inaccurate geometries due to the lack of an implicit surface representation~\mbox{\cite{xie2022neural}}. To solve this, works like~\mbox{\cite{wang2021neus,oechsle2021unisurf}} combine surface rendering~\mbox{\cite{kajiya1986rendering}} with volume rendering~\mbox{\cite{kajiya1984ray}}. Other works like~\mbox{\cite{verbin2021ref,zhang2021nerfactor,Zhang_2021_CVPR,Boss_2021_ICCV,Srinivasan_2021_CVPR}} adopt explicit representation on surface's reflectance to mimic realistic reflectance, which improve the utility and accuracy of NeRF~\mbox{\cite{mildenhall2020nerf}}. However, all mentioned methods work in multi-view stereo, which varies in the view direction with the static light. Unlike those methods, we reconstruct the surface normal of an object, instead of the full 3D structures, under a setup of photometric stereo that varies in the light directions with a fixed view direction. We also apply surface rendering~\mbox{\cite{kajiya1986rendering}} to explicitly model the interaction between light and objects, producing a more realistic approximation of the physical-based intrinsics of an object with different materials.

\textbf{Unsupervised Photometric Stereo.}
Classical methods solve the calibrated photometric stereo problem without knowing the ground truth surface normal. Therefore, we classify them as unsupervised methods. The least square-based algorithm~\cite{woodham1980} provides the simplest solution, which assumes the object to be Lambertian. It is generally served as a baseline method due to its stability, but its strong assumption on the reflectance model makes it fail for the non-Lambertian surface. The following works either regard the non-Lambertian reflectance components as the outliers~\cite{barsky20034, chung2008efficient, miyazaki2010median, wu2010robust, ikehata2012robust, wu2006dense} or apply analytic reflectance models including Torrance-Sparrow~\cite{georghiades2003incorporating}, the Ward model~\cite{chung2008efficient,goldman2009shape,ackermann2012photometric}, specular spike~\mbox{\cite{yeung2014normal}}, {\it etc.} to consider the non-Lambertian effects. Those analytic models explicitly consider different reflectance effects based on the microfacet theories~\cite{shi2013bi}. However, the performance of those methods can only deal with limited types of materials. 
There are also more advanced methods that utilize the general reflectance features such as isotropy~\cite{alldrin2008photometric} and monotonicity~\cite{higo2010consensus}. Those methods give a reliable estimation for objects with a broad range of materials. 

With the progress of deep learning, many learning-based frameworks have been proposed for calibrated photometric stereo.
Taniai {\it et al.}~\cite{taniai2018neural} proposed the first unsupervised learning-based photometric stereo method through a rendering equation. However, their reflectance model does not separately consider shadow, specular highlights, and diffuse components. 
Recently, \mbox{Li {\it et al.}~\cite{li2022neural}} proposed a framework that estimates the spatially varying BRDF (svBRDF) and the surface normal simultaneously with known lights. Considering the ambiguity in UPS, we handle cast shadow and specularity differently by exploiting the clues in those effects.
Particularly, the reconstruction loss between the rendered images and the ground truth images can backpropagate from the shadow field to the light field for robust light estimation, while their shadow handling module cannot be differentiated at the light's input. Moreover, their estimated shadow map is sensitive to the normal map at a specific epoch, which may introduce accumulation errors to the normal map estimation, while ours does not have that concern due to our light-normal joint optimization.

\textbf{Uncalibrated Photometric Stereo.}
Previous works hold the Lambertian assumption and exploit extra clues from reflectance, such as half-vector symmetry~\cite{lu2017symps}, albedo clustering~\cite{shi2010self}, specular spikes~\cite{yeung2014normal}, or make additional assumptions of light source distribution, such as ring light~\cite{zhou2010ring}, symmetry light~\cite{minami2022symmetric}, or uniform-distributed light sources~\cite{yuille1997shape, alldrin2007resolving, shi2010self, wu2013calibrating, papadhimitri2014closed, lu2017symps}, to resolve three parameters GBR ambiguity~\cite{belhumeur1999}. 
However, all these methods require the light intensity to be identical, which is inapplicable in the real-world datasets such as {\sc DiLiGenT}~\cite{shi2016benchmark}, {\sc Apple \& Gourd}~\cite{alldrin2008photometric}, and {\sc Light Stage Data Gallery}~\cite{chabert2006relighting}. 
Cho {\it et al.}~\cite{cho2018semi} put up a semi-calibrated method to deal with non-uniform light intensity, but they assume the light directions to be known. 
Quéau {\it et al.} \cite{queau2017non} address the photometric stereo problem under inaccurate lighting calibration, while the accuracy can significantly drop when non-Lambertian components become dominant. 

Recently, many deep learning methods have been proposed for uncalibrated photometric stereo. Chen {\it et al.}~\cite{chen2019self} propose a supervised uncalibrated framework, SDPS-Net, which can simultaneously estimate the light conditions (intensity and direction) and the surface normal. They suggest treating light estimation as a classification problem and separating the normal and light prediction to reduce the complexity. 
Their following work, GC-Net~\cite{chen2020learned}, improves the performance of SDPS-Net by adding shading as an extra channel to the input of the light estimation network. However, as a common problem for all supervised methods, an over-fitting problem may occur due to the training data bias \cite{mac2002problem}. In contrast, unsupervised methods do not have such a concern. 
Another benefit is that there is no need to synthesize training sets for unsupervised network training. 
To utilize the advantage of unsupervised methods, Kaya {\it et al.}~\cite{kaya2021uncalibrated} propose a compromised method that trains the light estimation network in a supervised manner (similar to \cite{chen2019self}) but estimates the surface normal in an unsupervised way (similar to~\cite{taniai2018neural}). 
However, they still suffer from training data bias during light estimation. Besides, all these methods make a strict assumption that the light intensity is distributed in a pre-defined range (\ie, [0.2, 2]) and suffer from accumulating error during normal estimation due to their two-step frameworks.
In contrast, our method neither makes a strict assumption on reflectance nor needs special light source distribution and jointly solves light conditions and surface normal in an unsupervised manner. 

\textbf{Intrinsics Decomposition.}
Intrinsics decomposition includes topics in a wide range of setting~\mbox{\cite{garces2022survey}}. Works involved are implemented under different types of light sources (\eg, directional light~\mbox{\cite{janner2017self}}, point light~\mbox{\cite{nestmeyer2020learning}}, or arbitrary lights~\mbox{\cite{yu2019inverserendernet,sengupta2019neural,zhou2019glosh}}), targeting on multifarious types of objects (\eg, single arbitrary object~\mbox{\cite{8908779}}, faces~\mbox{\cite{zollhofer2018state}}, or human body~\mbox{\cite{kanamori2019relighting}}) described in multiple material models (\ie, Blinn-Phong~\mbox{\cite{meka2018lime,zhang2021ners}}, general svBRDF~\mbox{\cite{deschaintre2018single,deschaintre2019flexible}}, etc.). We solved a sub-topic in intrinsics decomposition targeting on the single arbitrary object with the highest complexity in image formation models~\mbox{\cite{garces2022survey}}, where the surface normal, the object's material, and the light are unknown. We also assume the light source to be directional and distant.

\section{Method}
\label{section_3}
Given a set of observations $\boldsymbol{I} \triangleq (I_0, I_1, ..., I_f)$ of a static surface, illuminated by $f$ unknown directional illuminations distributing on the upper-hemisphere, uncalibrated photometric stereo aims at recovering light directions $\boldsymbol{L}\triangleq (\boldsymbol{l}_0, \boldsymbol{l}_1, ..., \boldsymbol{l}_f)$, light intensities $\boldsymbol{E}\triangleq (e_0, e_1, ..., e_f)$, and surface normal $\boldsymbol{N}\triangleq\{\boldsymbol{n}_i|i\in\mathbb{P}\}$, $\mathbb{P}$ is the set of all positions on the surface. 
The solution is achieved by solving the optimization problem, 
\begin{equation}
\begin{array}{c}
\label{eq:opt}
    \mathop{\arg\min}\limits_{\boldsymbol{L}, \boldsymbol{E}, \boldsymbol{N}}\sum_{i=1}^{\#\mathbb{P}}\sum_{j=1}^{f}{\text{D}(\bar{m}_{ij}, m_{ij})}, 
\end{array}
\end{equation}
where $\bar{m}_{ij}\in I_j$ is the observed pixel intensity at position $i$, \#$\mathbb{P}$ is the number of elements in $\mathbb{P}$\footnote{Without loss of generality, we put `\#' before a set symbol to represent its number of elements in this paper.}, $m_{ij}$ is the corresponding rendered pixel intensity, $\text{D}(\cdot,\cdot)$ is a metric describing their difference. Under an orthographic camera with the linear radiometric response, $m_{ij}$ is formulated as (simplified in a per-pixel form), 
\begin{equation}
\label{eq:general_brdf}
    m_{ij} = e_j \rho(\boldsymbol{n}_i, \boldsymbol{l}_j, \boldsymbol{v})\max(\boldsymbol{n}_i^\top \boldsymbol{l}_j, 0)
= e_j \rho_{ij}\max(\boldsymbol{n}_i^\top \boldsymbol{l}_j, 0),
\end{equation}
where $\boldsymbol{v}=[0,0,1]$ is the view direction pointing toward the viewer, $\rho_{ij}$ describes the general reflectance, $\max(\boldsymbol{n}_i^\top \boldsymbol{l}_j,0)$ represents the attach shadow. 

Unknown light brings two ambiguities when solving \eref{eq:opt}, \ie, shape-light ambiguity\footnote{Shape-light ambiguity is a more general form of GBR ambiguity in arbitrary non-Lambertian objects.}, which is denoted as an invertible matrix $\boldsymbol{G}\in\mathbb{R}^{3\times 3}$, and reflectance-light ambiguity, which is denoted as a non-zero scalar $c_j\in\mathbb{R}$,
\begin{equation}
\label{eq:amb}
    m_{ij} = e_j (c_jc_j^{-1})\rho_{ij}\max(\boldsymbol{n}_i^\top (\boldsymbol{G}\boldsymbol{G}^{-1})\boldsymbol{l}_j, 0).
\end{equation}

\subsection{Neural Intrinsics Fields}

\textbf{General reflectance decomposition.}
To exploit clues of specular reflectance and cast shadow for light estimation, we decompose the general reflectance as the cast shadow term $s_{ij}$ multiplying the bidirectional reflectance term,
\begin{equation}
\label{eq:field}
\rho_{ij}=s_{ij} (k^d_i\rho^d_i+k^s_i\rho^s_{ij}),
\end{equation}
where subscript `$i$' and `$j$' indicate position- (or normal-) and light-dependent factors, respectively; the cast shadow term $s_{ij}$ is either 0 or 1; $\rho^d_i, \rho^s_{ij}$ represent the diffuse and specular reflectance, and $k^d_i$ and $k^s_i$ are coefficients that balance out the effects of specular and diffuse reflectance\footnote{We think these coefficients of a point will not change under different lights, while they can be different at different positions.}.
\begin{figure}[t]
    \centering
    \includegraphics[width=0.85\textwidth]{"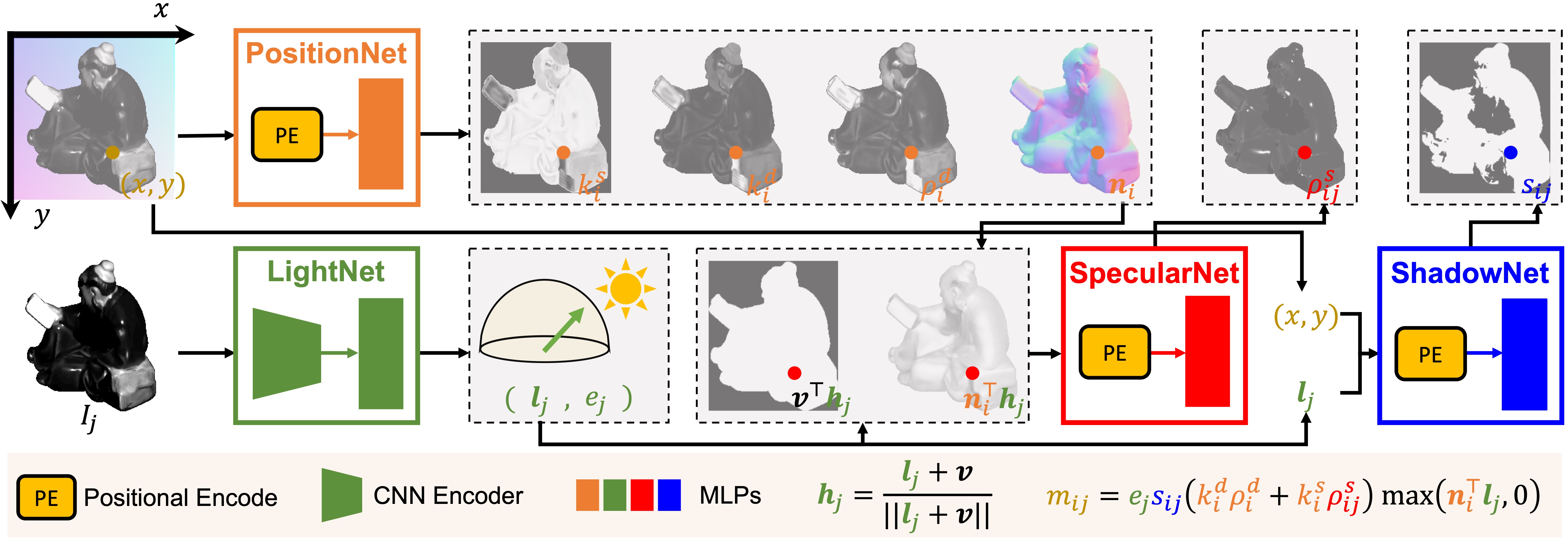"} 
    \caption{The framework of the proposed NeIF. PositionNet takes the input of positional code and outputs diffuse reflectance $\rho_i^d$, surface normal $\boldsymbol{n}_i$, and coefficients $k^d_i,k^s_i$. LightNet takes the observed image $I_j$ as the input and outputs light intensity $e_j$ and direction $\boldsymbol{l}_j$. SpecularNet takes $\boldsymbol{v}_j^{\top} \boldsymbol{h}_j$ and $\boldsymbol{n}^{\top} \boldsymbol{h}_j$ as the input and outputs specular reflectance $\rho^s_{ij}$. ShadowNet takes inputs of positional code and light direction $\boldsymbol{l}_j$ and output shadow indicator $s_{ij}$. All the intrinsics are used to render the observed pixel value $m_{ij}$ using \eref{eq:general_brdf} and \eref{eq:field}.} 
    \label{fig:overview}
    \vspace{-2.5em}
\end{figure}

\textbf{Neural fields of specular reflectance and cast shadow.}
Previous methods identify specular features from {\it specific} pixels and associate an {\it explicit} relation to light for light estimation. However, this scheme fails for the surface where the specular features are invisible, or the explicit relation is violated.
Besides, leaving out the clue of cast shadow can obstruct producing competitive performance. 

In contrast, we leverage {\it both} clues of specular and shadow for light estimation, which is achieved by building the neural specular field and neural shadow field and associating the fields to light conditions.
These neural fields make the utmost of {\it all} observed pixels and exploit the mutual information across different normal-light  (specular) and position-light (shadow) for light estimation.
Two MLPs, namely {\it SpecularNet} and {\it ShadowNet}, implement these neural fields, respectively. 
Instead of {\it explicitly} estimating light conditions from SpecularNet or ShadowNet, we take the estimated light direction as the input of these networks to achieve their {\it implicit} association to light conditions. Since the specular reflectance and cast shadow are normal- and position-dependent\footnote{Unlike~\mbox{\cite{li2022neural}}, our SpecularNet cannot estimate objects with multiple specular lobes. Due to the complexity of the UPS problem, a reflectance model with a smaller number of unknowns helps stabilize the training. Please refer to the \Sref{basis} in the appendix.}, we also feed the estimated surface normal and the positional code to them, respectively.
SpecularNet and ShadowNet output $\rho_{ij}^s$ and $s_{ij}$, respectively, as shown in \fref{fig:overview}. 

\textbf{Neural light field.}
There is mutual information across different observed images, \ie, observed images with similar appearances are expected to be illuminated by similar lights.
To fully consider and exploit this mutual information, we build the neural light field by concatenating a CNN encoder to an MLP, namely {\it LightNet}. The encoder extracts a light code from each observed image\footnote{We use light code instead of positional code because we experimentally find that the light code contains discriminative features of light intensity and direction.}, and the LightNet infers the corresponding light conditions (\ie, $e_j, \boldsymbol{l}_j$) from the light code, as shown in \fref{fig:overview}. 

\textbf{Neural position field.}
There is mutual information across different positions on a surface, \ie, the consistency of shape and diffuse reflectance in the spatial domain.
To fully consider and exploit this mutual information, we establish a neural position field, namely {\it PositionNet}, implemented by an MLP. The neural position field outputs position-dependent, light-independent factors\footnote{Since specular reflectance and cast shadow are both position-dependent and light-dependent, predicting them requires the input of light conditions, which increases the complexity of the neural position field. Therefore, we do not estimate them in the PositionNet, but predict them using SpecularNet and ShadowNet, respectively.}, \ie, $\boldsymbol{n}_i, \rho_i^d, k_i^d, k_i^s$. Particularly, factors like $k_i^d$, $k_i^s$ can generate svBRDF by assigning position-dependent values to different surface points even our specular field is not position-dependent (\ie, $k_i^s$ controls the specularity and $k_i^d$ controls the diffuse reflectance of a certain point). \footnote{Please refer to \Sref{svbrdf} in the appendix for a visual illustration of the estimated svBRDF.} The PositionNet takes the positional code as the input, as shown in \mbox{\fref{fig:overview}.}

\subsection{Optimizing Neural Intrinsics Fields}
We adopt the reconstruction loss function with the $\ell_1$ metric to optimize our NeIF, 
\begin{equation}
    \mathcal{L}_{\text{rec}} = \frac{1}{\text{\#} \mathbb{P} \times f}\sum_{i=1}^{\text{\#}\mathbb{P}}\sum_{j=1}^{f}{|\bar{m}_{ij} - e_js_{ij}(k^d_i\rho^d_i+k^s_i\rho^s_{ij})\max(\boldsymbol{n}_i^\top \boldsymbol{l}_j, 0)|},
\end{equation}

\textbf{Silhouette constraint.}
$\mathcal{L}_\text{rec}$ cannot resolve the shape-light ambiguity in an unsupervised manner due to the inherently severe ill-posedness. Therefore, we introduce the silhouette constraint (similar to those in~\cite{hashimoto2019uncalibrated, chen2020learned}) to stabilize the training of PositionNet~\footnote{Please refer to \Sref{SN} in the appendix for an explanation about how silhouette constraint solves shape-light ambiguity.}. To be specific, we use polynomial fitting with a moving window block to traverse and pre-compute the contour's normal of the given objects, represented as $\hat{\boldsymbol{N}}^\text{si}\triangleq\{\hat{\boldsymbol{n}}^\text{si}_k\in\mathbb{R}^{2\times1}, k\in\mathbb{S}\}$, $\mathbb{S}$ is the point set of the contour. 
We consider $\hat{\boldsymbol{N}}^\text{si}$ can guide the prediction of the azimuth of boundary normal (at the same positions) and introduce the silhouette loss function,
\begin{equation}
    \mathcal{L}_{\text{si}}=\sum_{k=1}^{\text{\#} \mathbb{S}}|{\text{Nor}}({\text C}({\boldsymbol{n}}_k)) - \hat{\boldsymbol{n}}^\text{si}_k|,
\end{equation}
where ${\boldsymbol{n}}_k\in\mathbb{R}^{1\times 3}$ represents the estimated surface normal at the positions of silhouette from PositionNet, ${\text C}(\cdot)$ cuts off the 3rd dimension of ${\boldsymbol{n}}_k$ (\ie, ${\text C}({\boldsymbol{n}}_k)\in\mathbb{R}^{1\times 2}$), and ${\text{Nor}(\cdot)}$ is the vector normalization operation. 

\textbf{Warm-up loss functions.}
To avoid local minimum and achieve faster convergence, we warm up the NeIF in early-stage during training with three additional loss functions. We use the azimuth of lighting direction estimated by YS97~\cite{yuille1997shape} to initialize the weights\footnote{We use YS97~\mbox{\cite{yuille1997shape}} as a weight initialization method on LightNet at the very beginning, with limited influence for subsequent training epochs due to its inaccurate result. To highlight our unsupervised learning manner, a better initialization (\eg, light estimation results of CW20~\mbox{\cite{chen2020learned}}) is not applied}. of LightNet,  
\begin{equation}
\mathcal{L}_\text{az}=\frac{1}{f}\sum_{j=1}^f|\text{Nor}(\text{C}(\boldsymbol{l}_j)) - \text{Nor}(\text{C}(\boldsymbol{l}^\text{az}_j))|_2, 
\end{equation}
where $\boldsymbol{l}^\text{az}_j$ are estimated light directions by~\cite{yuille1997shape}. We adopt the gradient penalty \cite{gulrajani2017improved} $\mathcal{L}_\text{gp}$ to stabilize the training of SpecularNet,
\begin{equation}
\mathcal{L}_\text{gp}=\frac{1}{\text{\#} \mathbb{P} \times f}\sum_{i=1}^{\text{\#} \mathbb{P}} \sum_{j=1}^{f} |\max(-\nabla_{\boldsymbol{n}^\top_i \boldsymbol{h}_j} \rho^s_{ij}, 0)|_2,
\end{equation}
The intuition is from Blinn-Phong model~\cite{blinn1977models}, where the specular reflectance is monotonically increasing {w.r.t} the $\boldsymbol{n}^\top_i \boldsymbol{h}_j$, \ie, $\nabla_{\boldsymbol{n}^\top_i \boldsymbol{h}_j} \rho_{ij}^s>0$. 
We also supervise the training of ShadowNet using pseudo shadow maps $\hat{\boldsymbol{S}}_j, j=1,2,..., f$\footnote{We present the details in \Sref{surface recon} of the appendix about how we calculate the $\hat{\boldsymbol{S}}_j$ by~\mbox{\cite{Cao_2021_CVPR}}.},
\begin{equation}
\mathcal{L}_\text{shadow} = \frac{1}{\text{\#}\mathbb{P}}\sum_{j=1}{f}|{\boldsymbol{S}}_j - \hat{\boldsymbol{S}_j}|_2,
\end{equation}
The pseudo shadow maps are obtained by binarizing the observed images, \ie, considering an observed pixel to be cast shadow if its intensity value is smaller than 0.2$\times$ the mean intensity value of this image.
After early-stage training, we discard these loss functions for a broad range of reflectance.

\textbf{NeIF training.}
We train NeIF with the warm-up loss function in first 10 epochs,
\begin{equation}
  \mathcal{L}_\text{warmup}=\mathcal{L}_\text{rec}+\lambda_\text{si}\mathcal{L}_\text{si}+\lambda_\text{az}\mathcal{L}_\text{az}+\lambda_\text{gp}\mathcal{L}_\text{gp}+\lambda_\text{shadow}\mathcal{L}_\text{shadow},
\end{equation}
where $\lambda_\text{si}=5, \lambda_\text{az}=0.1, \lambda_\text{gp}=10, \lambda_\text{shadow}=10$.
We then train NeIF until 500 epochs or converging with the loss function,
\begin{equation}
  \mathcal{L}_\text{NeIF}=\mathcal{L}_\text{rec}+\lambda_\text{si}\mathcal{L}_\text{si}+\lambda_\text{shadow}\mathcal{L}_\text{recShadow}.
\end{equation}
where $\mathcal{L}_\text{recShadow}$ is the another shadow map supervision loss to train the ShadowNet.
The loss function is the same to $\mathcal{L}_\text{shadow}$. 
The only difference is the calculation of $\hat{\boldsymbol{S}}_j$. For $\mathcal{L}_\text{recShadow}$, we calculate $\hat{\boldsymbol{S}}_j$ by rendering a depth map with the estimated $\boldsymbol{l}_j$.
The depth map is reconstructed from the estimated surface normal map $\boldsymbol{N}$ by method~\cite{Cao_2021_CVPR}. 
$\mathcal{L}_\text{recShadow}$ is used to align outputs of ShadowNet to those of PositionNet.

\textbf{Implementation details.}
We generate the positional code from the coordinate (in the image plane) by the same method in~\cite{mildenhall2020nerf}. With the assumption of isotropic reflectance, we simplify the input of SpecularNet from  $\{\boldsymbol{v}^\top \boldsymbol{l}_j, \boldsymbol{v}^\top\boldsymbol{n}_i, \boldsymbol{n}^\top_i\boldsymbol{l}_j\}$ to $\{\boldsymbol{v}^\top\boldsymbol{h}_j,  \boldsymbol{n}^\top_i\boldsymbol{h}_j\}$\footnote{$\boldsymbol{h}_j$ is the bisector of $\boldsymbol{l}_j$ and $\boldsymbol{v}$, $\boldsymbol{h}_j=\frac{\boldsymbol{l}_j+\boldsymbol{v}}{\|\boldsymbol{l}_j+\boldsymbol{v}\|}$.} \cite{lu2017symps} for easier training. 
Similar to the Cook-Torrance reflectance model \cite{cook1982reflectance}, we assume $k^d_i+k^s_i=1$ to reduce the number of unknowns.  
The CNN decoder with declining channels processes the down-sampled images with a dimension of $256\times256$. The LightNet takes flatten features to predict $\boldsymbol{l}_j, e_j$ in two different branches. 
For the ShadowNet, we also generate the positional code for $\boldsymbol{l}_j$ and concatenate it to the feature in the 9th layer. The output of ShadowNet is either 0 or 1, which is realized by a step function with a similar implementation in \cite{qin2020binary}.

\section{Experiments}
For network structure, device specification, and hyperparameters, reader can refer to \Sref{implementation} of the appendix for more details. For evaluation, we adopt the same metric in~\cite{chen2019self}, the scale-invariant relative error, to measure the accuracy of recovered light intensity as,
\begin{equation}
\label{li_err}
E_\text{int}=\frac{1}{f} \sum_{j=1}^{f}\left(\frac{\left|\eta e_{j}-\tilde{e}_{j}\right|}{\tilde{e}_{j}}\right).
\end{equation}
where, $\eta$ is calculated by solving $\mathop{\arg\min} _\eta\sum_{j=1}^{f}\left(\eta e_{j}-\tilde{e}_{j}\right)^{2}$ by least squares minimization. The metric to measure the accuracy of the predicted light directions and surface normal is the widely used mean angle error (MAE) in degree.

\begin{table}[t]
    \fontsize{6pt}{7pt}\selectfont
    \setlength{\tabcolsep}{2pt}
\caption{Quantitative comparison in terms of mean angular error for surface normal on {\sc DiLiGenT} benchmark~\cite{shi2016benchmark}. This table summarizes comparison methods. We report the mean of five random tests. Please refer to \Sref{std} of the appendix for the standard deviation of our method. `N.A.' represents not applicable as calibrated PS is with known $\boldsymbol\ell$ and $e$. `Semi' indicates certain methods leverage partial information of light. `\ding{51}' (or `\ding{55}') represents that certain methods (do not) adopt supervised learning for the estimation of surface normal $\boldsymbol{n}$, light direction $\boldsymbol{l}$, or light intensity $e$. `Identical' means certain methods require the light intensity of different illuminating to be identical.}
    \label{tab:normal_diligent}
    \centering
    \label{normal_comparison}
\begin{tabular}{c|cccc|cccccccccc|c}
    \hline
    Method & \specialcell{PS/\\UPS} & \specialcell{$\boldsymbol{n}$\\Supervision} & \specialcell{$\boldsymbol{l}$\\Supervision} & \specialcell{$e$\\Supervision} & \sc{Ball}  & \sc{Bear}  & \sc{Buddha} & \sc{Cat}   & \sc{Cow}   & \sc{Goblet} & \sc{Harvest} & \sc{Pot1}  & \sc{Pot2}  & \sc{Reading} & AVG    \\ \hline
    LS~\cite{woodham1980}         & PS   & \ding{55} & N.A. & N.A. & 4.10  & 8.39  & 14.92 & 8.41  & 25.60  & 18.50   & 30.62 & 8.89  & 14.65 & 19.80  & 15.39  \\
    TM18~\cite{taniai2018neural} & PS   & \ding{55} & N.A. & N.A. & \textbf{1.47}  & 5.79  & 10.36 & 5.44  & 6.32  & 11.47  & 22.59 & 6.09  & 7.76  & 11.03 & 8.83   \\
    LL22~\cite{li2022neural} & PS   & \ding{55} & N.A. & N.A. & 2.43  & \textbf{3.64}  & \textbf{8.04} & \textbf{4.86}  & \textbf{4.72}  & \textbf{6.68}  & \textbf{14.90} & \textbf{5.99}  & \textbf{4.97}  & \textbf{8.75} & \textbf{6.50}   \\
    \hline
    CH19~\cite{chen2019self} & UPS & \ding{51} & \ding{51}   & \ding{51} & 2.77  & 6.89  & 8.97  & 8.06  & 8.48  & 11.91  & 17.43 & 8.14  & 7.50  & 14.90 & 9.51  \\
    CW20~\cite{chen2020learned} & UPS & \ding{51} & \ding{51} & \ding{51} & 2.50  & 5.60  & \textbf{8.60}  & 7.90  & 7.80  & 9.60   & \textbf{16.20} & 7.20  & 7.10  & 14.90 & 8.71  \\
    CM20~\cite{cho2018semi}    & Semi& \ding{55} & Known & \ding{55} & 2.78 & 8.07 & 13.38 & 8.05 & 26.90 & 18.18 & 33.35 & 9.47 & 19.58 & 14.19 & 15.40 \\
    KK21~\cite{kaya2021uncalibrated} & UPS & \ding{55} & \ding{51} & \ding{51} & 3.78  & 5.96  & 13.14 & 7.91  & 10.85 & 11.94  & 25.49 & 8.75  & 10.17 & 18.22 & 11.62 \\
    SK22~\cite{sarno2022neural}   & UPS & \ding{51} & \ding{51} & \ding{51} & 3.46 & 5.48 & 10.00    & 8.94 & \textbf{6.04} & 9.78  & 17.97 & 7.76 & 7.10   & 15.02 & 9.15 \\
    \textbf{Ours} & UPS & \ding{55} & \ding{55} & \ding{55} & \textbf{1.17}  & \textbf{4.49}  & \textbf{8.73}  & \textbf{4.89}  & 6.27  & \textbf{9.53}  & 18.31  & \textbf{7.08}  & \textbf{5.85}  & \textbf{12.02}  & \textbf{7.83}  \\
    \hline
    \end{tabular}
    \vspace{-1.5em}
\end{table}

\subsection{Evaluation on Public Datasets}

Since the proposed NeIF is an unsupervised uncalibrated photometric stereo method, we compare its performance with state-of-the-art uncalibrated and unsupervised photometric stereo methods.
Three real-world datasets, including the {\sc DiLiGenT} benchmark dataset~\cite{shi2016benchmark}, {\sc Apple \& Gourd} dataset~\cite{alldrin2008photometric}, and {\sc Light Stage Data Gallery} dataset~\cite{chabert2006relighting}, are used for evaluation. \footnote{More ablation studies could be found in \Sref{"tag:ablation study"} of the appendix. Comprehensive estimation results on the public datasets can be found in \Sref{comparison} of the appendix.}

\begin{figure}[t]
\centering
\includegraphics[width=\textwidth]{"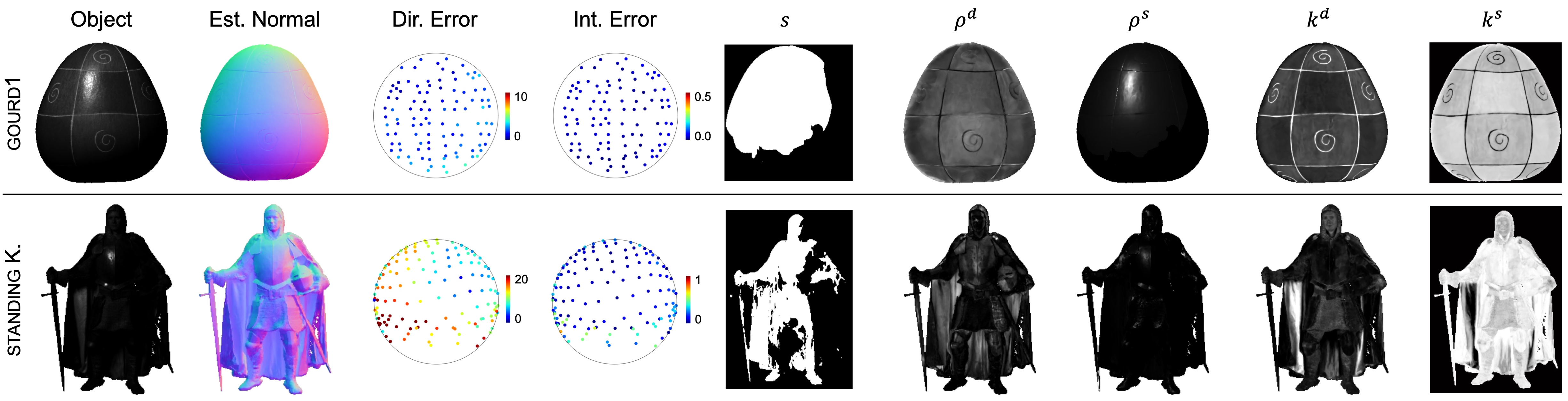"} 
\caption{Left-right: the reference image, estimated surface normal map from our method, the lighting direction error, light intensity error, the estimated shadow map, diffuse reflectance map, specular reflectance map, diffuse scaling coefficient map, and specular coefficient map. Two objects are {\sc Gourd1} from  {\sc Apple \& Gourd} dataset~\cite{alldrin2008photometric}, and {\sc Standing Knight} from {\sc Light Stage Data Gallery} dataset~\cite{chabert2006relighting}.} 
\label{fig:visual_apple}
\vspace{-1em}
\end{figure}

\textbf{Comparison for normal map.}
\Tref{tab:normal_diligent} lists relevant works for a comprehensive surface normal estimation comparison. 
As summarized in \Tref{tab:normal_diligent}, our method is the only method that addresses UPS without the supervision of $\boldsymbol{N}$, $\boldsymbol{L}$, or $\boldsymbol{E}$. However, our method achieves the best performance in concurrent UPS methods and maintains a considerable advantage over other competitors~\cite{chen2019self, chen2020learned}.
Handling the objects in {\sc DiLiGenT} dataset~\cite{shi2016benchmark} with different shapes and reflectance,
numbers from our method are either best or competitive (about 1$^\circ$ as compared with the best performing UPS method), which shows its good generalization capacity to general reflectance and various shapes. This is because our NeIF fully considers mutual information by building up intrinsics fields, and the implicit modeling facilitates general reflectance modeling. 
We also show the reconstructed normal map and other estimated factors for \mbox{{\sc Apple}} and \mbox{{\sc Standing Knight}} in \mbox{\fref{fig:visual_apple}}. Our method produces reliable estimation for most regions, thanks to the full exploitation of mutual information across different dimensions. However, there are inconsistent boundaries on the predicted shadow map, mainly due to the per-pixel estimation manner of ShadowNet. Those artifacts are hard to remove because ShadowNet functions in such a per-pixel way providing important clues for light estimation and stabilizing the training process.\footnote{We replace the ShadowNet with pseudo shadow maps, and the results show the necessity of the ShadowNet. Please refer to \Sref{statistical shadow handling} of the appendix for more details.}


\textbf{Comparison for light directions and intensities.}
As can be observed in \Tref{tab:light_diligent},  our method achieves a superior performance advantage over unsupervised methods (YS97~\cite{yuille1997shape} and PF14~\cite{papadhimitri2014closed}) while maintaining competitive performance as compared with supervised methods (CH19~\cite{chen2019self} and CW20~\cite{chen2020learned}). These supervised methods adopt two-step solutions and pre-defined intensities range, which suffer from accumulating error and data bias. Therefore, even though they achieve similar or better performance in terms of light conditions accuracy on published datasets, they are less effective in estimating surface normal or even failed in the intensities distribution deviating from the pre-defined range, as compared with our method (see \Tref{tab:normal_diligent}). Readers can refer to \Sref{comparison} of the appendix for more details about the light estimation results on \mbox{{\sc Apple \& Gourd} dataset~\cite{alldrin2008photometric}} and \mbox{{\sc Light Stage Data Gallery} dataset~\cite{chabert2006relighting}}, and \Sref{random scale} of the appendix for the randomly scaled light intensities experiments.


\setlength{\tabcolsep}{2pt}
\begin{table}[t]
    \fontsize{6pt}{7pt}\selectfont
    \centering
    \caption{Quantitative comparison in terms of mean angluar error for light direction and scale-invariant error for intensity on {\sc DiLiGenT} benchmark~\cite{shi2016benchmark}.}
    \label{tab:light_diligent}
    \resizebox{\linewidth}{!}{
    \begin{tabular}{c|cccccccccccccccccccc|cc}
    \hline
    & \multicolumn{2}{c}{\sc{Ball}} & \multicolumn{2}{c}{\sc{Bear}} & \multicolumn{2}{c}{\sc{Buddha}} & \multicolumn{2}{c}{\sc{Cat}} & \multicolumn{2}{c}{\sc{Cow}} & \multicolumn{2}{c}{\sc{Goblet}} & \multicolumn{2}{c}{\sc{Harvest}} & \multicolumn{2}{c}{\sc{Pot1}} & \multicolumn{2}{c}{\sc{Pot2}} & \multicolumn{2}{c|}{\sc{Reading}} & \multicolumn{2}{c}{AVG} \\ 
    Model         & dir. & int.  & dir. & int.  & dir. & int.  & dir. & int.  & dir.  & int.  & dir.  & int.  & dir.  & int.  & dir. & int.  & dir.  & int.  & dir.  & int.  & dir.  & int.  \\
    \hline
    YS97~\cite{yuille1997shape} & 12.41 & 0.334 & 14.06 & 0.260 & 11.68 & 0.300 & 13.75 & 0.318 & 15.79 & 0.251 & 15.24 & 0.316 & 59.41 & 0.586 & 12.99 & 0.322 & 12.58 & 0.283 & 13.08 & 0.266 & 18.10 & 0.320 \\
    PF14~\cite{papadhimitri2014closed}          & 4.90 & 0.036 & 5.24 & 0.098 & 9.76 & 0.053 & 5.31 & 0.059 & 16.34 & 0.074 & 33.22 & 0.223 & 24.99 & 0.156 & 2.43 & \textbf{0.017} & 13.52 & 0.044 & 21.77 & 0.122 & 13.75 & 0.088 \\
    CH19~\cite{chen2019self}         & 3.27 & 0.039 & 3.47 & 0.061 & 4.34 & 0.048 & 4.08 & 0.095 & 4.52  & 0.073 & 10.36 & 0.067 & 6.32  & 0.082 & 5.44 & 0.058 & 2.87  & 0.048 & \textbf{4.50}   & 0.105 & 4.92  & 0.068 \\
    CW20~\cite{chen2020learned}         & 1.75 & 0.027 & \textbf{2.44} & 0.101 & 2.86 & \textbf{0.032} & 4.58 & 0.075 & \textbf{3.15}  & \textbf{0.031} & \textbf{2.98}  & \textbf{0.042} & \textbf{5.74}  & 0.065 & \textbf{1.41} & 0.039 & \textbf{2.81}  & 0.059 & 5.47  & 0.048 & \textbf{3.32}  & 0.052 \\
    \hline
    \textbf{Ours} & \textbf{1.79} & \textbf{0.014}  & 3.54 & \textbf{0.011}  & \textbf{2.33} & 0.034 & \textbf{2.60} & \textbf{0.023} & 5.81  & 0.196 & 8.45  & 0.045  & 7.40  & \textbf{0.032} & 3.73 & 0.071 & 2.10  & \textbf{0.037} & 7.91  & \textbf{0.047} & 4.57  & \textbf{0.051} \\
    \hline
    \end{tabular}
    }
    \vspace{-2em}
\end{table}

\subsection{Evaluation on Challenging Setup}

\begin{figure}[t]
\centering
\includegraphics[width=0.9\textwidth]{"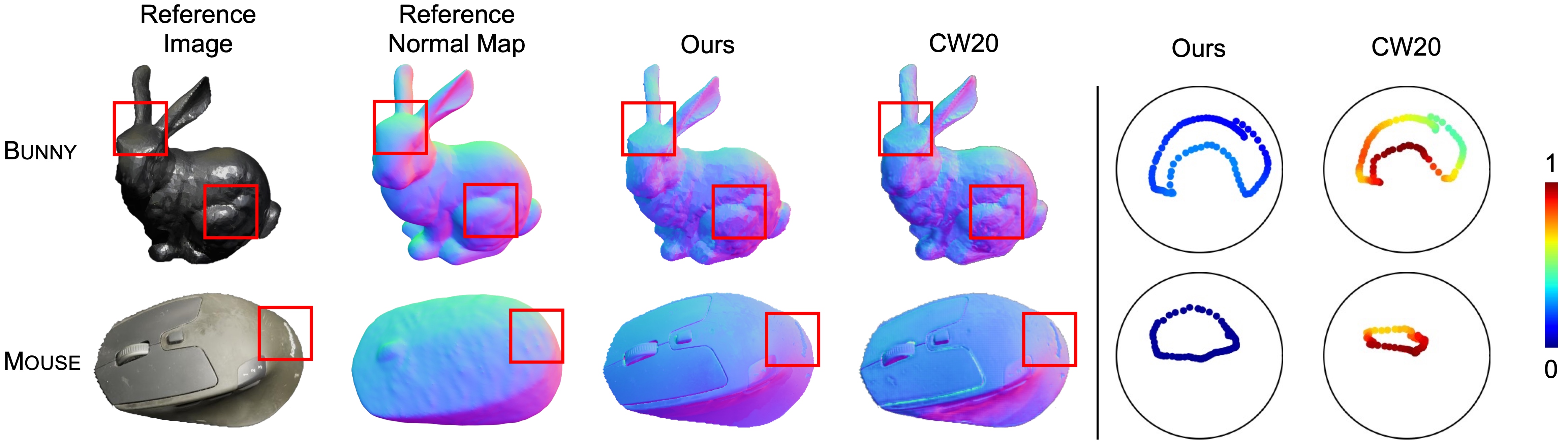"} 
\caption{Visual quality comparison in terms of normal map on \mbox{{\sc Bunny}} (row 1) and \mbox{{\sc Mouse}} (row 2). Left-right: mean of the observed images, coarse normal map download in \mbox{\url{https://sketchfab.com/}} that has similar shape with the objects for reference, normal map from our method, CW20~\mbox{\cite{chen2020learned}}, light estimation result from our method, CW20~\mbox{\cite{chen2020learned}}. The predicted lights are projected onto the XY-plane. The color of the subfigures (column 5-6) indicates the value of the light intensity.}
\label{fig:visual_natural}
\vspace{-1em}
\end{figure}

\textbf{Casual scenarios.} To further highlight the advantage of our method regarding unsupervised learning and light-normal joint optimization, we perform an experimental comparison on data collected in casual environments (\ie, the environment with the ambient light and non-ideal directional light source. Please refer to \Sref{NLUPS with ambient} of the appendix for more details about the experiment setting and data collecting process). 
We compare our method with the state-of-the-art UPS method, CW20~\mbox{\cite{chen2020learned}}. The estimated normal map for \mbox{{\sc Bunny}} and \mbox{{\sc Mouse}} are shown in \mbox{\Fref{fig:visual_natural}}\footnote{Another object, \mbox{{\sc Venus}}, is made up of spatially varying BRDF with anisotropic material (\ie, glass). The results are shown in \Sref{NLUPS with ambient} of the appendix.}. 
According to the result, CW20~\mbox{\cite{chen2020learned}} are sensitive to the data bias in supervised learning of the light estimation model.
The error of estimated light dramatically degrades the performance of normal estimation.
In contrast, we have a more reasonable estimation of the selected objects thanks to our unsupervised and light-normal joint optimization manner. The results indicate that our method can better balance the accuracy of the surface normal and light in general cases.

\textbf{Sparse Uncalibrated Photometric Stereo.} To further illustrate the superiority and robustness of our method, we investigate NeIF under sparse lights. We randomly select 10 or 16 images illuminated by different lights from \mbox{{\sc DiLiGenT}} dataset~\mbox{\cite{shi2016benchmark}} and test our method using these images. We report the mean results over 30 trials in \Sref{"tag:sparse ups"} of the appendix. According to the results, our method achieves competitive performance on normal estimation as compared with state-of-the-art sparse PS methods~\mbox{\cite{zheng2019spline,Chen_2018_ECCV,papadhimitri2014closed}} with known light (\ie, 11.62 for NeIF vs. 9.82 for ZJ19~\mbox{\cite{zheng2019spline}} under 10 lights, and 10.38 for NeIF vs. 9.00 for CH18~\mbox{\cite{Chen_2018_ECCV}} under 16 lights). Kindly note that we do not require any ground truth normal for supervision. Compared with the unsupervised UPS method (PF14~\mbox{\cite{papadhimitri2014closed}}), we achieve superior performance. Please refer to \Sref{"tag:sparse ups"} of the appendix for more details on quantity and quality comparison.

\section{Conclusion}
This paper proposes NeIF for UPS. By representing the general reflectance as four neural intrinsics fields, it implicitly imposes the light clues of specular reflectance and cast shadow for light estimation, which facilitates solving UPS with general reflectance. 
The proposed NeIF can fully exploit mutual information from all observed pixel values so that it produces stable estimation for UPS under challenging settings.
The unsupervised training manner of NeIF is beneficial to the generalization capacity of data from different sources.

\textbf{Limitations.} Although our method produces promising results for light conditions and surface normal estimation, it has several limitations. 
First, due to the challenging UPS task, we simplify the image formation model for stable training (\ie, we do not consider the inter-reflection and anisotropic materials; we assume the specular reflectance to be single lobe and bivariate; we only predict the gray-level intrinsics).
Second, as shown in \mbox{\Fref{fig:visual_apple}}, there are artifacts in predicted material intrinsics and shadow map due to the shape-light and reflectance-light ambiguities.
Third, the training process of our method is more time-consuming compared to concurrent inverse rendering methods~\mbox{\cite{li2022neural,taniai2018neural}} due to the joint optimization manner and the per-pixel estimation, which is inapplicable for real-time applications\footnote{Please refer to \Sref{computation} of the appendix for further analysis about the time consumption.}. At last, the image's noise will degrade our model's performance and other inverse rendering methods. Therefore, the observed images should be neither overexposure nor underexposure.

\newpage
\printbibliography

\newpage
\section{Appendix}
\appendix
In this appendix,
\begin{enumerate}[itemsep=0pt,topsep=0pt,parsep=0pt]
    \item \Sref{basis}: we switch our rendering model to be identical to~\cite{li2022neural}, where we use 9 specular basis instead of 1. (footnote 6 in the main paper).
    \item \Sref{correlation study}: we conduct correlation analysis to validate the effectiveness of the CNN extractor. (footnote 7 in the main paper).
    \item \Sref{svbrdf}: we visualize the BRDFs of different points to demonstrate that our method could generate spatially varying BRDF. (footnote 9 in the main paper).
    \item \Sref{SN}: we provide an intuitive explanation of how silhouette constraints work in our method. (footnote 10 in the main paper).
    \item \Sref{surface recon}: we present details of calculating $\boldsymbol{\hat{S}}_j$. (footnote 12 in the main paper).
    \item \Sref{std}: we provide the standard deviation for 10 objects on the {\sc DiLiGenT} benchmark~\cite{shi2016benchmark} under 5 random tests. (table 1's caption in the main paper).
    \item \Sref{implementation}: we provide details of the network structure, device specification, and training procedure. (Sec. 4 in the main paper).
    \item \Sref{"tag:ablation study"}: we conduct ablation studies to validate  effectiveness of our loss functions and network structures. (footnote 14 in the main paper).
    \item \Sref{statistical shadow handling}: we compare our method with those that remove the ShadowNet and use the shadow maps obtained from simple binarization of observed images (as in~\cite{li2022neural}, footnote 15 in the main paper).
    \item \Sref{comparison}: we show complete qualitative comparison regarding 10 objects from {\sc DiLiGenT} dataset~\cite{shi2016benchmark}, 3 objects from {\sc Light Stage Data Gallery~\cite{chabert2006relighting}}, and 3 objects from {\sc Apple \& Gourd}~\cite{alldrin2008photometric} in \sref{comparison}. (footnote 14 and Sec. 4.1 in the main paper). 
    \item \Sref{random scale}: we randomly scale the intensity of input images and provide an experimental comparison. (Sec. 4.1 in the main paper).
    \item \Sref{NLUPS with ambient}: we collect data in casual scenarios to validate the effectiveness of our method on unsupervised learning and light-normal joint optimization. (Sec. 4.2 in the main paper).
    \item \Sref{"tag:sparse ups"}: we conduct additional experiments to validate the effectiveness of our method for sparse uncalibrated photometric stereo. (Sec. 4.2 in the main paper).
    \item \Sref{computation}: we list the testing time of our method for objects in {\sc DiLiGenT} benchmark~\cite{shi2016benchmark}. We also explain why our model takes a long time to train. (footnote 17 in the main paper).
\end{enumerate}
\section{Multiple Specular Basis}
\label{basis}
We extend the number of the basis used in our SpecularNet from 1 to 9 and switch our rendering model to be identical to \cite{li2022neural}. Specifically, 
\begin{equation}
    m_{ij} =e_{j}(\rho^{d}_i + \mathbf{c}_i^{\top} D(\boldsymbol{h}_{ij}, \boldsymbol{n}_i))\max \left(\boldsymbol{n}_{i}^{\top} \boldsymbol{l}_{j}, 0\right).
\end{equation}
Where, $m_{ij}$ is the pixel value, $e_{j}$ is the light intensity of light $j$, $\rho_i^d$ is the diffuse reflectance at point $i$, $\boldsymbol{c}_i\triangleq (c_0, ..., c_k)$ is the specular weights generated by the PositionNet, $k$ is the number of basis, $\boldsymbol{h}_{ij}$ is the half-vector between the surface normal at point $i$ and the light direction $\boldsymbol{l}_j$, $\boldsymbol{n}_i$ is the surface normal of point $i$). The results are shown in \Tref{tab:multiple basis}. Although there are improvements in objects like {\sc Bear}, {\sc Pot1}, and {\sc Reading}, the overall performance drops significantly. This further illustrates that our rendering model is a necessary compromise for the complexity of the UPS problem.

\begin{table}[h]
  \centering
  \captionsetup{justification   = raggedright, singlelinecheck = false}
  \caption{Quantitative results in terms of mean angular error for surface normal, light direction, and scale-invariant error for intensity on {\sc DiLiGenT} benchmark ~\cite{shi2016benchmark}. `Ours (9 basis)' indicates using 9 specular basis in SpecularNet.}
  \resizebox{\linewidth}{!}{
    \begin{tabular}{c|ccccccccccc|c}
    \hline
    &       & \textsc{Ball}  & \textsc{ Bear}  & \textsc{ Buddha} & \textsc{ Cat}   & \textsc{ Cow}   & \textsc{ Goblet} & \textsc{ Harvest} & \textsc{ Pot1}  & {\sc Pot2}  & \textsc{ Reading} & AVG \\
    \hline
    \multirow{3}[0]{*}{Ours} & norm.  & 1.17  & 4.49  & 8.73  & 4.89  & 6.27  & 9.53  & 18.31  & 7.08  & 5.85  & 12.02  & 7.83\\
          & dirs. & 1.79  & 3.54  & 2.33  & 2.60  & 5.81  & 8.45  & 7.40  & 3.73  & 2.10  & 7.91  & 4.57\\
          & ints. & 0.01  & 0.01  & 0.03  & 0.02  & 0.20  & 0.04  & 0.03  & 0.07  & 0.04  & 0.05  & 0.05\\
    \hline
    \multirow{3}[0]{*}{Ours (9 basis)} & norm.  & 1.69  & 4.03  & 12.84  & 5.90  & 8.89  & 14.45  & 20.02  & 6.82  & 9.64  & 11.34  & 9.56 \\
          & dirs. & 2.50  & 3.01  & 7.34  & 4.94  & 8.69  & 21.53  & 12.60  & 3.88  & 6.97  & 6.41  & 7.79  \\
          & ints. & 0.011  & 0.019  & 0.025  & 0.036  & 0.074  & 0.072  & 0.059  & 0.026  & 0.030  & 0.051  & 0.040  \\
    \hline
    \end{tabular}%
    }
  \label{tab:multiple basis}%
\end{table}%

\newpage
\section{Correlation Analysis}
\label{correlation study}
\begin{figure}[h]
\centering
\includegraphics[width=0.8\textwidth]{"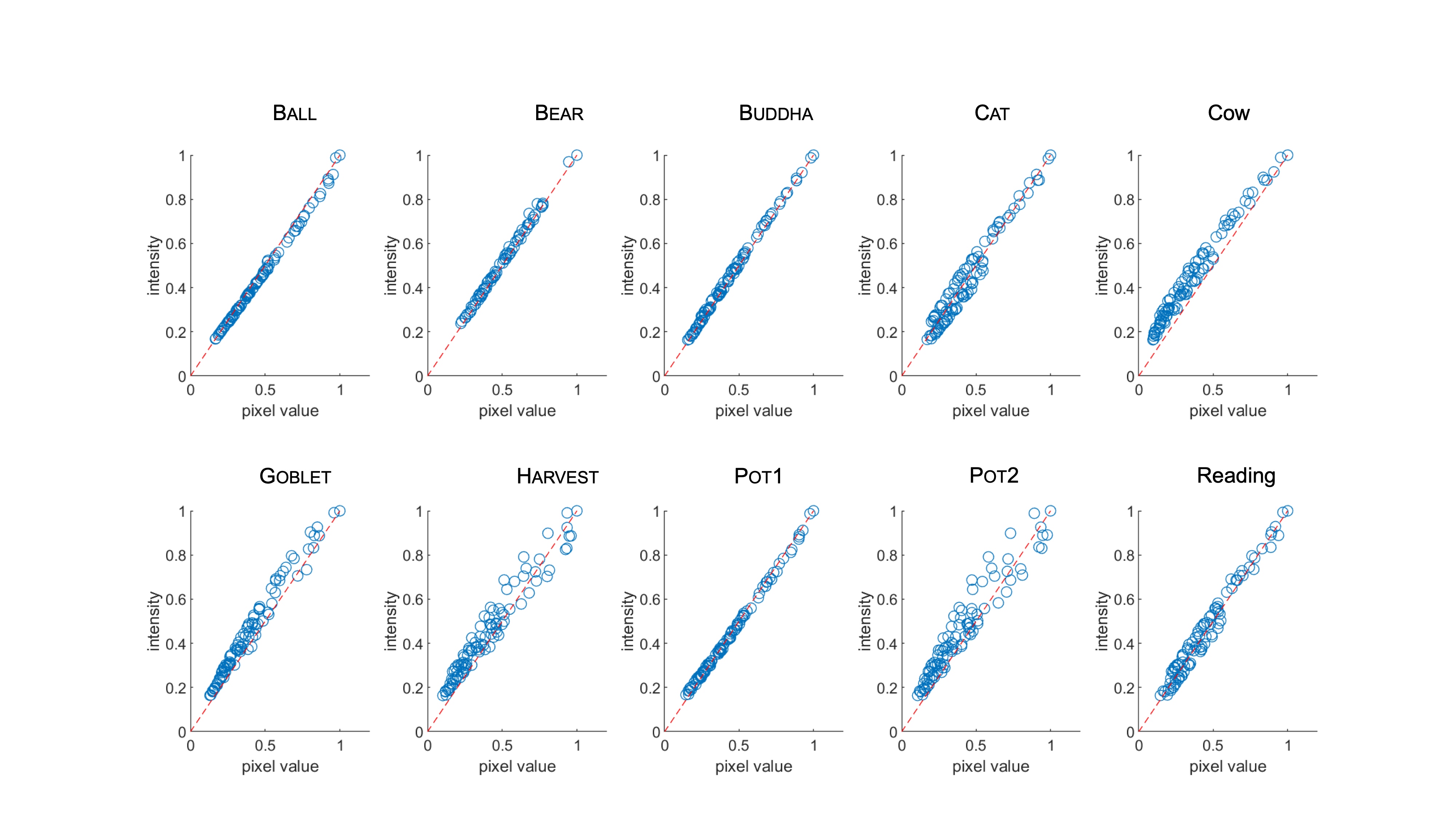"}
\captionsetup{justification   = raggedright, singlelinecheck = false}
\caption{Correlation analysis between the pixel value and the light intensity on each object from DiLiGenT dataset~\cite{shi2016benchmark}. The X-axis refers to the pixel value of the 90th percentile maximum of each 96 image. Y-axis indicates the corresponding light intensity of the corresponding image. The values are scaled into [0, 1] by min-max normalization. The mean correlation of the 10 objects is 0.98.}
\label{fig:light_intensity}
\end{figure}

While the positional code of the light index seems to be a good candidate as the input of LightNet, it is less effective to estimate light conditions than the observed image as it provides observable clues of intensity.
\fref{fig:light_intensity} reveal that the observed images provides important clues.
Therefore, we take the observed image as the input of LightNet.

\begin{figure}[h]
\centering
\includegraphics[width=0.7\textwidth]{"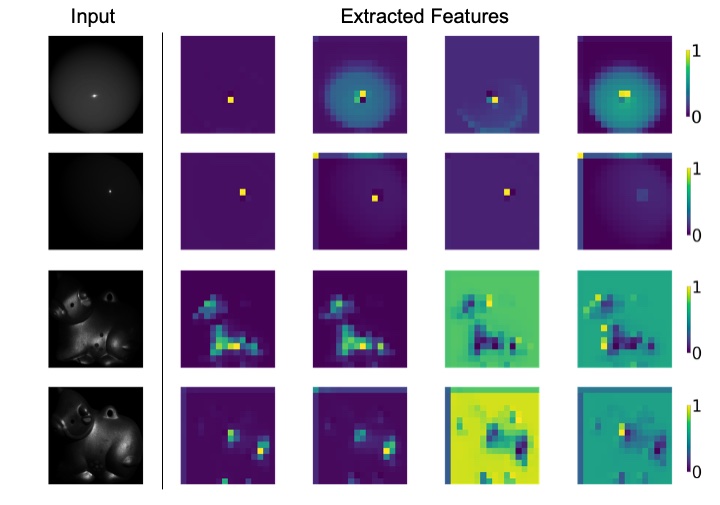"}
\captionsetup{justification   = raggedright, singlelinecheck = false}
\caption{Feature visualization of CNN encoder on {\sc Ball} and {\sc Cow}. Left-right: the input images of the CNN encoder, four feature maps from the last layer of the CNN encoder. Similar to~\cite{chen2020learned}, the feature maps are normalized into $\left[0, 1\right]$ for  better illustration.}
\label{fig:feature map}
\end{figure}

\textbf{Qualitative validation.} We extract feature maps from the last layer of our CNN encoder. As shown in \Fref{fig:feature map}, those feature maps contain observable specular features, indicating both the direction and the intensity of the light sources.\cite{chen2020learned}

\textbf{Quantitative validation.} We conduct correlation analysis to illustrate the correlation between the learned feature map and the lighting information. First, we use Euclidean Distance (\ie, $\ell_2$) to measure the distance between feature maps under two different lights, and we order these distances to get a $\frac{96\times95}{2}$-dimension vector $\boldsymbol{V}^f_i, i=1,2,3,4$. 
Similarly, we get two $\frac{96\times95}{2}$-dimension vectors by measuring the distance between every two lights, \ie, $\boldsymbol{V}^{ld}$ for light directions with the metric of cosine distance, and $\boldsymbol{V}^{li}$ for light intensities with the metric of $\ell_1$ difference.
The cosine similarity between these vectors is shown in \Tref{tab:corr}. We find that there are \textbf{at least one} feature map that is highly correlated to the ground truth light direction or light intensity, especially for objects like {\sc Ball} and {\sc Reading} that have sharp specular lope.

\begin{table}[h]
  \centering
  \captionsetup{justification   = raggedright, singlelinecheck = false}
  \caption{The cosine similarity between the distance matrix of the four extracted feature maps ($\boldsymbol{V}^f_i, i=1,2,3,4$) and the light directions/intensities. "max" indicates the maximum similarity among the feature maps.}
    \resizebox{\linewidth}{!}{
    \begin{tabular}{c|cccccccccccccccccccc}
        \hline
          & \multicolumn{2}{c}{\sc Ball} & \multicolumn{2}{c}{\sc Bear} & \multicolumn{2}{c}{\sc Buddha} & \multicolumn{2}{c}{\sc Cat} & \multicolumn{2}{c}{\sc Cow} & \multicolumn{2}{c}{\sc Goblet} & \multicolumn{2}{c}{\sc Harvest} & \multicolumn{2}{c}{\sc Pot1} & \multicolumn{2}{c}{\sc Pot2} & \multicolumn{2}{c}{\sc Reading} \\
          & dir.  & int.  & dir.  & int.  & dir.  & int.  & dir.  & int.  & dir.  & int.  & dir.  & int.  & dir.  & int.  & dir.  & int.  & dir.  & int.  & dir.  & int. \\
          \hline
    $\boldsymbol{V}^f_1$ & 0.84  & 0.93  & 0.83  & 0.69  & 0.96  & 0.81  & 0.93  & 0.81  & 0.94  & 0.76  & 0.87  & 0.80  & 0.82  & 0.89  & 0.92  & 0.67  & 0.90  & 0.77  & 0.89  & 0.91  \\
     $\boldsymbol{V}^f_2$ & 0.96  & 0.81  & 0.91  & 0.82  & 0.92  & 0.82  & 0.95  & 0.71  & 0.94  & 0.75  & 0.94  & 0.76  & 0.87  & 0.80  & 0.94  & 0.70  & 0.89  & 0.66  & 0.89  & 0.89  \\
     $\boldsymbol{V}^f_3$ & 0.85  & 0.90  & 0.86  & 0.74  & 0.90  & 0.89  & 0.94  & 0.79  & 0.94  & 0.76  & 0.92  & 0.80  & 0.86  & 0.86  & 0.89  & 0.62  & 0.89  & 0.71  & 0.90  & 0.85  \\
     $\boldsymbol{V}^f_4$ & 0.96  & 0.78  & 0.95  & 0.81  & 0.90  & 0.67  & 0.96  & 0.74  & 0.94  & 0.77  & 0.88  & 0.86  & 0.88  & 0.82  & 0.95  & 0.72  & 0.90  & 0.76  & 0.89  & 0.87  \\
    max.  & 0.96  & 0.93  & 0.95  & 0.82  & 0.96  & 0.89  & 0.96  & 0.81  & 0.94  & 0.77  & 0.94  & 0.86  & 0.88  & 0.89  & 0.95  & 0.72  & 0.90  & 0.77  & 0.90  & 0.91  \\
    \hline
    \end{tabular}%
  \label{tab:corr}%
  }
\end{table}%
\section{Spatially Varying BRDF}
\label{svbrdf}
\begin{figure}[h]
\centering
\includegraphics[width=0.5\textwidth]{"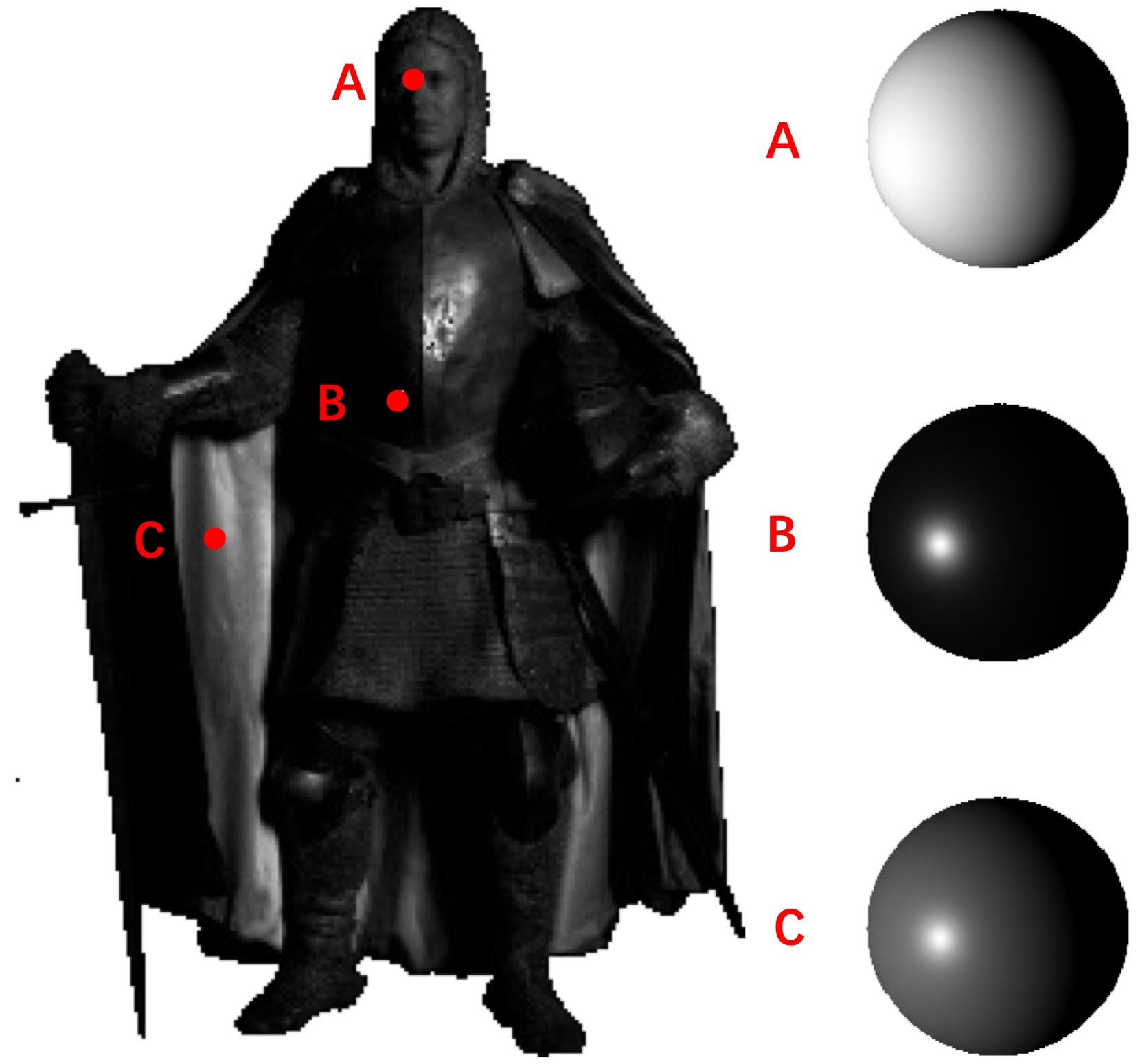"} 
\captionsetup{justification   = raggedright, singlelinecheck = false}
\caption{Visual illustration of BRDF at different positions. On the left is the observed image of {\sc Standing Knight}. We select three different points on the object and show our predicted BRDF spheres of those points on the right side. We normalized the image and the predicted BRDF from 0 to 1 for a better illustration. For the right, sub-figures top-bottom are: BRDF for the point on the face, BRDF for the point on the armor, and BRDF for the point on the cloak, respectively.}
\label{fig:svbrdf}
\end{figure}

We visualize the BRDF in different position of {\sc Knight Standing} in \Fref{fig:svbrdf}. Although we are restricted by our specular model, we can still generate spatially varying BRDF by assigning different scaling factors $k_d\in [0,1]$ to different spatial points. That is, $k_d$ controls the specularity of a certain point. The reflectance could be Lambertion if $k_d$ is 1 and non-Lambertian otherwise. For instance, we have a larger $k_d$ on the knight's face and cloak, but a smaller $k_d$ on the armor, generating different BRDF and shading effects shown in \Fref{fig:svbrdf}.

\newpage
\section{Silhouette Constraint}
\label{SN}
For Lambertian objects, the GBR ambiguity is represented as:
\begin{equation}
\left\{
\begin{array}{rr}
    \boldsymbol{I}^k =& \boldsymbol{B}^{\top}\boldsymbol{S}^k,\\
    \boldsymbol{B}^k =& \rho^d\boldsymbol{G}^{-\top}\boldsymbol{N},\\
    \boldsymbol{S}^k =& e^k \boldsymbol{G}\boldsymbol{l}^k,
\end{array}
\right.
\end{equation}
Where, $\boldsymbol{G}$ is the $3\times 3$ ambiguity matrix, $\rho^d $ is the albedo, $\boldsymbol{N}$ is the surface normal, $e^k$ is the light intensity, $\boldsymbol{l}^k$ is the light direction. We denote the object’s silhouette normal as $\boldsymbol{N}^{s}$, the fitted silhouette normal as $\bar{\boldsymbol{N}}^{s}$, and the predicted silhouette normal by PositionNet as $\bar{\boldsymbol{N}}^{s}$ that contains GBR ambiguity, \ie, $\bar{\boldsymbol{N}}^{s}=\boldsymbol{G}^{-\top}\boldsymbol{N}^{s}$. 

During training, the output of the LightNet is $\bar{e}^k$ and $\bar{\boldsymbol{l}}^k$. The GBR ambiguity is solved by: 
\begin{equation}
\mathop{\arg\min}_{\boldsymbol{G}}\sum_{k=1}^f(|\boldsymbol{I}^{k} - \bar{\boldsymbol{B}}^{{\mathrm{s}}\top}\bar{\boldsymbol{S}}^{k}| + \left|\operatorname{Nor}\left(\mathrm{C}\left(\bar{\boldsymbol{N}}^{\mathrm{s}}\right)\right)-\hat{\boldsymbol{N}}^{\mathrm{s}}\right|).
\end{equation}
Where, $f$ is the number of the light source, ${\text C}(\cdot)$ cuts off the 3rd dimension of ${\boldsymbol{n}}_k$ (\ie, ${\text C}({\boldsymbol{n}}_k)\in\mathbb{R}^{1\times 2}$), and ${\text{Nor}(\cdot)}$ is the vector normalization operation. This can be easily extended to the non-Lambertian case with the shape-light ambiguity. 

\begin{figure}[h]
\centering
\includegraphics[width=\textwidth]{"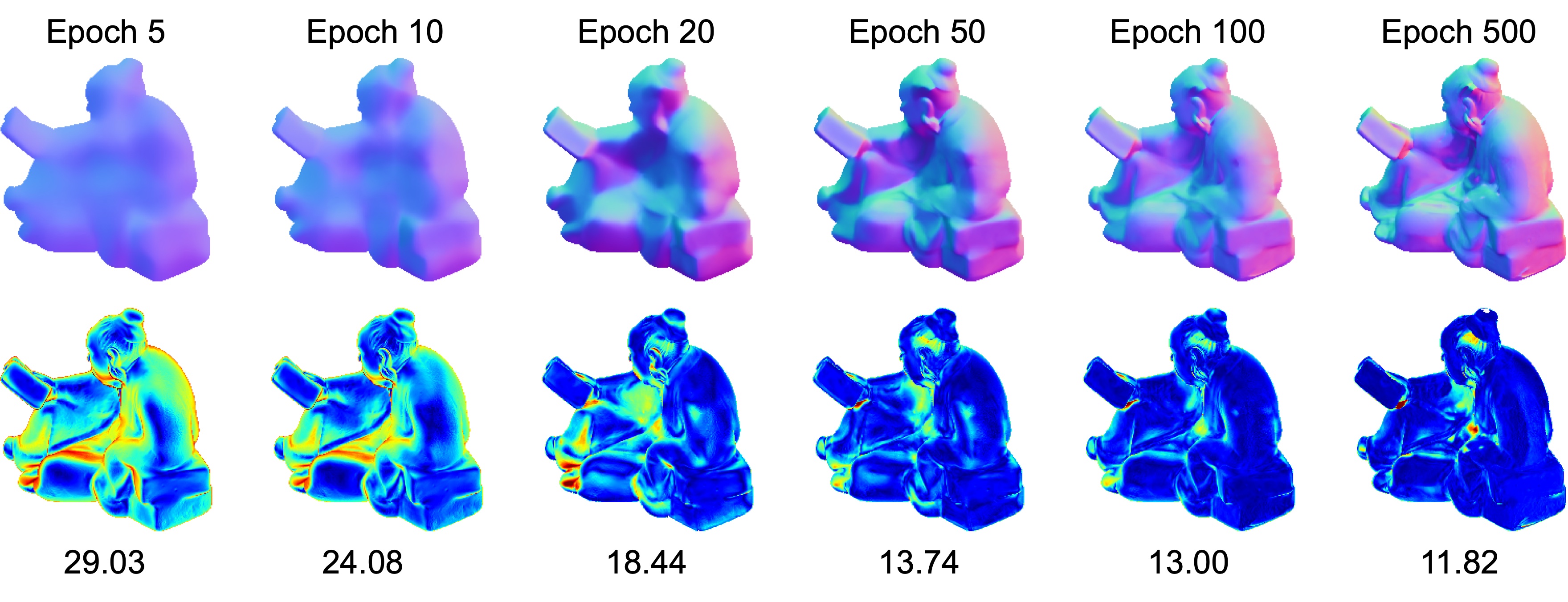"}
\captionsetup{justification   = raggedright, singlelinecheck = false}
\caption{Visual qualitative comparison of the surface normal for different training epochs. Top: the surface normal maps. Bottom: the error maps. Note that the silhouette normal takes effects from the boundary.}
\label{fig:boundary_normal}
\end{figure}

We also intuitively illustrate the effect of fitted silhouette normal in \fref{fig:boundary_normal}. The expression of the shape-light ambiguity during training is the synchronous rotation of estimated surface normal and light direction. While we fix the silhouette normal's azimuth angle, these constraints help mitigate the ambiguity by aligning the predicted surface normal to the fitted silhouette normal's azimuth, which suppresses the rotation of surface normal. 

\newpage
\section{Calculating $\boldsymbol{\hat{S}}_j$}
\label{surface recon}
The ShadowNet implicitly records the interaction between the depth map and the light direction. 
To align the estimated surface normal (output of PositionNet) to the estimated shadow (output of ShadowNet), we establish the relationship between the normal map and shadow map.
We achieve this by reconstructing the depth map from the predicted normal map and rendering the shadow map from the depth map and estimated light direction.

{\bf Depth map reconstruction.} 
Reconstructing the depth map is also called normal integration, and many algorithms are proposed to deal with it. 
We apply the orthogonal five-point plane fitting method in~\cite{Cao_2021_CVPR} for depth map reconstruction.
Because the normal map from our method might contain outliners due to our per-pixel manner solution, while method in~\cite{Cao_2021_CVPR}is more robust to outliers and sharp features than previous ones.
Specifically, this method formulates normal integration as an inverse problem of local plane fitting in the camera coordinates by minimizing the perpendicular point-to-plane distance. 

The method formulates the plane as 
\begin{equation}
    \mathbf{p}^\top \mathbf{n} + d = 0
\end{equation}
where $\mathbf{n}$ is the unit normal vector perpendicular to the plane, and $d$ is the distance from the coordinate origin to the plane. 
And the goal is to minimize the sum of the squared distance of all points to their nearby plane. 
\begin{equation}
    \min_{\mathbf{p}, d} \sum_{\mu \in \Omega_n} \sum_{\mathbf{p} \in \mathcal{N}(\mathbf{p(\mathbf{u})})} \left( \mathbf{p}^\top \mathbf{n}(\mathbf{u}) + d(\mathbf{u}) \right) ^ 2
\end{equation}
where each point $p$ is parameterized by the known projection of the 3D point on the image plane $\mathbf{u}$, and the 3D point's unknown position on the camera ray $z$. And $\mathcal{N}(\mathbf{p(\mathbf{u})})$ is defined as five nearest neighbor points, including $\mathbf{p}$ itself and its left / right / upper / lower counterparts. 

{\bf Shadow rendering.} We set out a light ray at the query point $\boldsymbol{x}=(u, v, w)$ from the depth map, $\boldsymbol{r}=\boldsymbol{x}+t\boldsymbol{l}$, where $\boldsymbol{l}$ is the estimated light direction and $t$ is a parameter for sampling points.
Then, a series of points (we sampled 32 pixel points) $\boldsymbol{X}=\{(u_i, v_i, w_i)|1 \leq i \leq 32\}$ on the ray can be sampled.
We then get 32 depth values $\{\hat{w}_i|1 \leq i \leq 32\}$ with the same $(u_i, v_i)$ in $\boldsymbol{X}$ from the depth map.
The query point $\boldsymbol{x}$ is determined to be either in the cast shadow or not through the equation:
\begin{equation}
    \hat{s} = H(\min\{w^i_l-\hat{w}^i|1 \leq i \leq 32\})
\end{equation}
Where $H(\cdot)$ is the Heaviside step function.

\section{Standard Deviation of Random Tests on {\sc DiLiGenT}~\cite{shi2016benchmark} Benchmark}
By changing the random seeds, we provide the standard deviation for 10 objects on the {\sc DiLiGenT}~\cite{shi2016benchmark} benchmark. The results are shown in \Tref{tab:mean}.
\label{std}
\begin{table}[h]
  \centering
  \captionsetup{justification=raggedright, singlelinecheck=false}
  \caption{Quantitative results in terms of mean angular error for surface normal, light direction, and scale-invariant error for intensity on {\sc DiLiGenT} benchmark ~\cite{shi2016benchmark}. This table summarizes two versions. Rows 2-4 are the reported results in the main paper. Row 5-7 are the mean of 5 random tests.}
  \resizebox{\linewidth}{!}{
    \begin{tabular}{c|ccccccccccc|cc}
    \hline
          &       & {\sc Ball}  & {\sc Bear}  & {\sc Buddha} & {\sc Cat}   & {\sc Cow}   & {\sc Goblet} & {\sc Harvest} & {\sc Pot1}  & {\sc Pot2}  & {\sc Reading} & AVG   & STD \\
    \hline
    \multirow{3}[0]{*}{Ours} & norm.  & 1.17  & 4.49  & 8.73  & 4.89  & 6.27  & 9.53  & 18.31  & 7.08  & 5.85  & 12.02  & 7.83  & 0.44  \\
          & dirs. & 1.79  & 3.54  & 2.33  & 2.60  & 5.81  & 8.45  & 7.40  & 3.73  & 2.10  & 7.91  & 4.57  & 0.77  \\
          & ints. & 0.014 & 0.011 & 0.034 & 0.023 & 0.196 & 0.045 & 0.032 	& 0.071 & 0.037 & 0.047 & 0.051 & 0.033   \\
          
    \hline
    \end{tabular}%
    }
  \label{tab:mean}%
\end{table}%

\newpage
\section{Implementation Details}
\label{implementation}
\begin{figure}[H]
\vspace{-1em}
\centering
\includegraphics[width=\textwidth]{"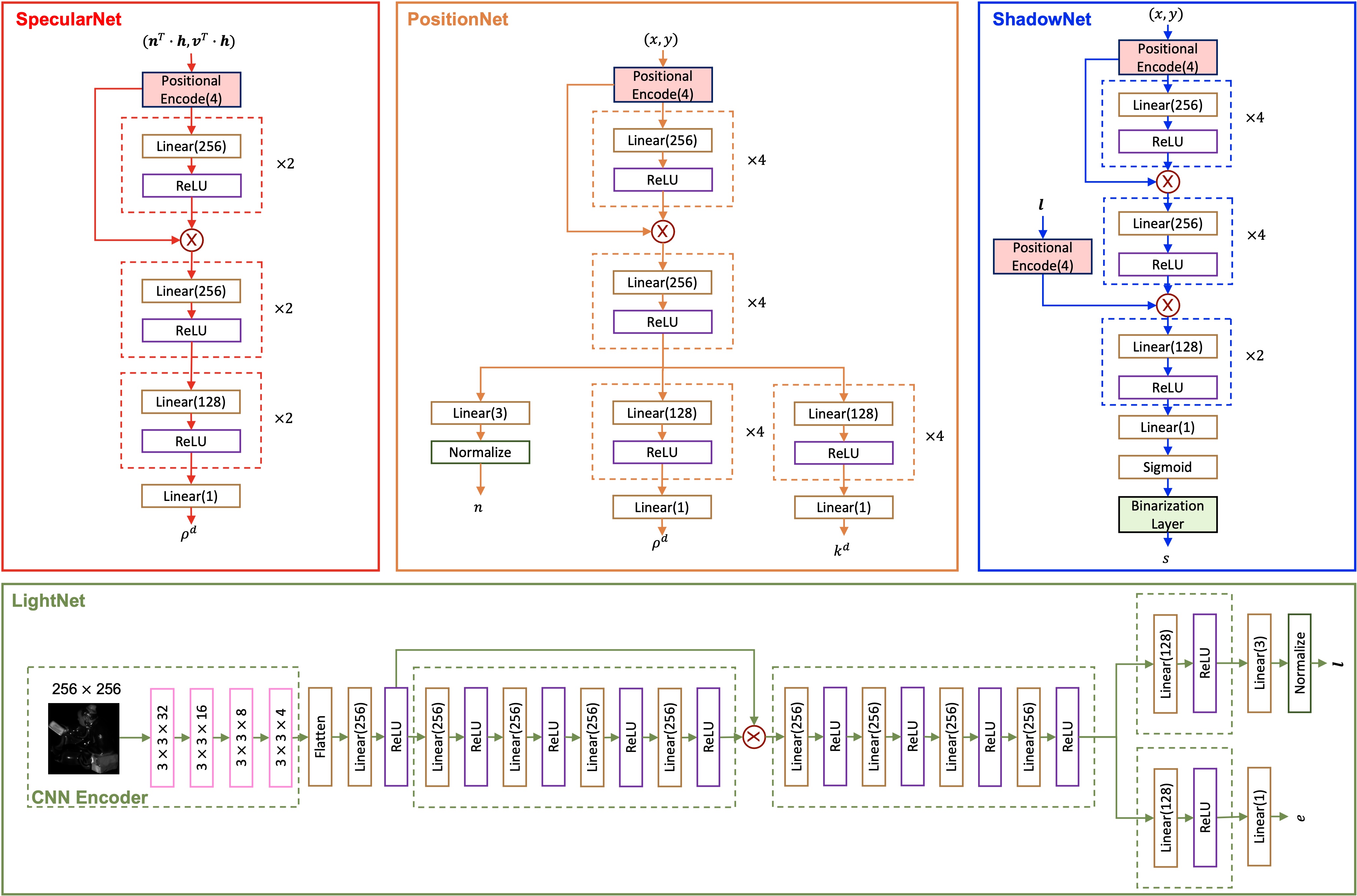"}
\captionsetup{justification   = raggedright, singlelinecheck = false}
\caption{Network architectures of SpecularNet, PositionNet, ShadowNet, and LightNet. }
\label{fig:network structure}
\vspace{-1em}
\end{figure}

In \fref{fig:network structure}, we display the network architectures of PositionNet, LightNet, ShadowNet, and SpecularNet. 

\textbf{Positional Encoding}. The positional encoding module is identical to NeRF~\cite{mildenhall2020nerf}\footnote{\url{https://github.com/facebookresearch/pytorch3d/tree/main/projects/nerf}}, which is based on the following equation \ref{equ:positional encoding},
\begin{equation}
\label{equ:positional encoding}
    E(p)=\left(\sin \left(2^{0} \pi p\right), \cos \left(2^{0} \pi p\right), \cdots, \sin \left(2^{L-1} \pi p\right), \cos \left(2^{L-1} \pi p\right)\right),
\end{equation}
where $L$ is the dimension of the positional code, $p$ is the encoded features.
The $p$ in our model is a vector, \ie,  $(\cos{<\boldsymbol{h}_j, \boldsymbol{n}_i>}, \cos{<\boldsymbol{h}_j, \boldsymbol{v}>}$) for SpecularNet, $(x, y)$ for PositionNet and ShadowNet, and $\boldsymbol{l}$ for ShadowNet. 
We compute each dimension's positional code $E(p)$ and concatenate these codes to get our positional code.

\textbf{Binarization Layer}. The binarization layer follows a similar design in~\cite{qin2020binary}. Considering a step function $B(\cdot)$,
$$
B(x)=\left\{
\begin{aligned}
     1 &  & x > 0.5 \\
     0 &  & x \leq 0.5 
\end{aligned}
\right.
$$
During training, the gradient of the binarization layer is defined as $\nabla_{x}B(x) = 1$ to make the ShadowNet trainable, which is similar to that in~\cite{qin2020binary}. 

\textbf{Training details.} Our main framework is implemented in PyTorch, while the pre-calculations (method~\cite{yuille1997shape} and silhouette fitting) are implemented in MATLAB. We use Adam as the optimizer with a learning rate $\alpha=5 \times 10^{-4}$ to train our framework in 500 epochs for each scene separately, and the warm-up stage takes up 10 epochs. The last 100 epochs use a lower learning rate $\alpha=5 \times 10^{-5}$ for fine-tunning. The batch size is $32$ for lighting and $256$ for spatially random sampling. After every epoch, the depth map is reconstructed according to the predicted normal map by method~\cite{Cao_2021_CVPR}. Each scene takes from 2 hours to 6 hours on one RTX 2080Ti 12GB GPU, depending on the resolution of the objects. For Sparse UPS, due to the fewer reconstruction terms to train our NeIF, we slightly increase the dimension of the positional encoding module from 4 to 6 that increases the frequency~\cite{vaswani2017attention} to stabilize the training, \ie, strengthening the role of positional code to reduce the variance of estimated intrinsics.
Besides, we drop $\mathcal{L}_\text{az}$ during early stage warm-up because YS97~\cite{yuille1997shape}  fails for the sparse inputs.

\section{Ablation Study}
\label{"tag:ablation study"}
In this section, we evaluate the effectiveness of the loss functions in Eq. (10) and Eq. (11) and learning strategies.
To be specific, we consider the ablation studies including,
\begin{itemize}[itemsep=0pt,topsep=-1pt,parsep=0pt]
    \item (1) without back-propagation from ShadowNet to LightNet; 
    \item (2) without back-propagation from SpecularNet to LightNet;
    \item (3) without both mentioned back-propagation paths; 
    \item (4) without early-stage supervised by YS97~\cite{yuille1997shape};
    \item (5) without gradient penalty in the early-stage.
\end{itemize}  

The quantitative results comparison is displayed in \Tref{tab:ablation_study}.
The comparison with cases (1), (2), and (3) indicates the important roles of shadow and specular for uncalibrated photometric stereo as they provide clues of light estimation.
The comparison with case (4) shows that early-stage weak supervision is necessary to prevent the neural networks from poor initialization. 
The comparison with case (5) illustrates the effectiveness of the gradient penalty. Without that, the prediction of $\rho_s$ could be less stable. 
\begin{table}[h]
    \vspace{-1em}
  \centering
  \caption{Quantitative comparison in terms of estimated normal, light direction, and light intensity errors on {\sc DiLiGenT} dataset~\cite{shi2016benchmark}. {\it w/o} $\rho_s$-back indicates without back-propagation from ShadowNet to LightNet; {\it w/o} $s$-back indicates without back-propagation from SpecularNet to LightNet; {\it w/o} YS97~\cite{yuille1997shape} indicates no YS97~\cite{yuille1997shape} weak-supervision (\ie, $\mathcal{L}_\text{az}$ in Eq. (10)) in the early-stage warm-up; {\it w/o} GP indicates no gradient penalty loss function (\ie, $\mathcal{L}_\text{gp}$ in Eq. (10)) in the early stage warm-up.}
  \resizebox{\linewidth}{!}{
    \begin{tabular}{c|ccccccccccc|c}
    \hline
          &       & \multicolumn{1}{c}{\sc Ball} & \multicolumn{1}{c}{\sc Bear} & \multicolumn{1}{c}{\sc Buddha} & \multicolumn{1}{c}{\sc Cat} & \multicolumn{1}{c}{\sc Cow} & \multicolumn{1}{c}{\sc Goblet} & \multicolumn{1}{c}{\sc Harvest} & \multicolumn{1}{c}{\sc Pot1} & \multicolumn{1}{c}{\sc Pot2} & \multicolumn{1}{c|}{\sc Reading} & \multicolumn{1}{c}{AVG} \\
          \hline
    \multirow{3}[0]{*}{Ours} & norm.  & 1.17  & 4.49  & 8.73  & 4.89  & 6.27  & 9.53  & 18.31  & 7.08  & 5.85  & 12.02  & 7.83\\
          & dirs. & 1.79  & 3.54  & 2.33  & 2.60  & 5.81  & 8.45  & 7.40  & 3.73  & 2.10  & 7.91  & 4.57\\
          & ints. & 0.014 & 0.011 & 0.034 & 0.023 & 0.196 & 0.045 & 0.032 	& 0.071 & 0.037 & 0.047 & 0.051 \\
    \hline
    \multirow{3}[0]{*}{(1) {\it w/o} $\rho_s$-back} & norm.  & 1.07  & 6.66  & 13.26  & 4.94  & 9.71  & 11.44  & 31.35  & 6.88  & 6.19  & 16.29  & 10.78  \\
          & dirs.  & 4.74  & 7.14  & 2.70  & 12.79  & 13.10  & 13.10  & 11.07  & 6.67  & 2.13  & 19.54  & 9.30  \\
          & ints.  & 0.011  & 0.011  & 0.099  & 0.030  & 0.148  & 0.043  & 0.477  & 0.044  & 0.034  & 0.054  & 0.095  \\
    \hline
    \multirow{3}[0]{*}{(2) {\it w/o} $s$-back} & norm.  & 1.04  & 5.32  & 12.59  & 5.28  & 6.41  & 12.03  & 17.20  & 7.36  & 6.22  & 11.20  & 8.47  \\
          & dir.  & 1.95  & 5.98  & 11.24  & 4.00  & 6.25  & 16.18  & 5.74  & 7.12  & 2.01  & 7.04  & 6.75  \\
          & int.  & 0.011  & 0.013  & 0.045  & 0.024  & 0.048  & 0.050  & 0.033  & 0.041  & 0.037  & 0.043  & 0.035  \\
    \hline
    \multirow{3}[0]{*}{ (3) \specialcell{{\it w/o} $\rho_s$-back\\{\it w/o} $s$-back}} & norm.  & 1.26  & 8.76  & 13.66  & 7.17  & 13.56  & 11.68  & 32.35  & 7.84  & 6.37  & 16.67  & 11.93  \\
          & dirs.  & 2.28  & 13.66  & 7.41  & 8.23  & 19.50  & 15.12  & 11.57  & 7.47  & 2.82  & 20.66  & 10.87  \\
          & ints.  & 0.011  & 0.028  & 0.106  & 0.027  & 0.200  & 0.048  & 0.277  & 0.017  & 0.042  & 0.071  & 0.083  \\
    \hline
    \multirow{3}[0]{*}{(4) {\it w/o} YS97 \cite{yuille1997shape}} & norm.  & 1.29  & 9.31  & 10.39  &  7.74     &  39.74     & 27.65  & 34.03  & 6.99       & 57.82  & 20.29  & 21.52  \\
          & dirs.  & 1.42  & 13.36  & 9.41  &  7.5     & 29.08  & 66.78  & 71.31  &      6.52 & 50.15  & 27.08  & 28.26  \\
          & ints.  & 0.013 & 0.039  &   0.031    &   0.224    &   0.045    &  0.389     &     0.270  & 0.056 &  0.448     & 0.124 &  0.164 \\
    \hline
    \multirow{3}[0]{*}{(5) {\it w/o} GP} & norm.  & 1.10  & 4.11  & 8.45  & 6.65  & 7.29  & 9.65  & 30.52  & 12.35  & 6.83  & 16.08  & 10.30  \\
          & dirs.  & 1.65  & 3.36  & 2.39  & 7.94  & 9.24  & 11.05  & 6.02  & 11.35  & 1.99  & 9.79  & 6.48  \\
          & ints.  & 0.028 & 0.011  &  0.030 & 0.197& 0.042 &    0.041   &   0.079    & 0.146  & 0.055 & 0.235 & 0.086  \\
    \hline
    \end{tabular}%
    }
  \label{tab:ablation_study}%
\vspace{-1.5em}
\end{table}%

\section{Statistical Shadow Handling}
\label{statistical shadow handling}
We remove the ShadowNet and train our method using the pseudo shadow maps for reconstruction loss. 
The pseudo shadow maps are obtained by binarizing the observed images, \ie, considering an observed pixel to be cast shadow if its intensity value is smaller than $0.2\times$ the mean intensity values of this image. The result is shown in \Tref{tab:pseudo shadow}. Although pseudo shadow map works well on objects like {\sc Ball}, {\sc Buddha}, and {\sc Cat}, the performance,  the accuracy of the light directions, dropped significantly on objects like {\sc Harvest}, {\sc Goblet}, and {\sc Reading}. This is because pseudo shadow maps do not provide any extra clues for the LightNet through the back-propagation.
\begin{table}[h]
  \vspace{-1.5em}
  \centering
  \captionsetup{justification   = raggedright, singlelinecheck = false}
  \caption{Quantitative comparison in terms of normal map error, light direction error, and light intensity error on {\sc DiLiGenT} dataset~\cite{shi2016benchmark}. `Ours w $\hat{s}$' indicates using pseudo shadow map instead of the ShadowNet for shadow handling.}
  \resizebox{\linewidth}{!}{
    \begin{tabular}{c|ccccccccccc|c}
    \hline
    & & {\sc Ball} & {\sc Bear} & {\sc Buddha} & {\sc Cat} & {\sc Cow} & {\sc Goblet} & {\sc Harvest} & {\sc Pot1} & {\sc Pot2} & {\sc Reading} & {AVG} \\
    \hline
    \multirow{3}[0]{*}{Ours w $\hat{s}$} & norm.  & 1.68  & 6.52  & 8.33  & 4.87  & 7.92  & 18.32  & 39.08  & 8.02  & 8.20  & 12.91  & 11.59  \\
    & dirs.  & 2.60  & 5.00  & 2.32  & 1.34  & 6.82  & 26.61  & 14.48  & 9.28  & 6.69  & 12.25  & 8.74  \\
    & ints.  & 0.014  & 0.012  & 0.030  & 0.024  & 0.067  & 0.056  & 0.530  & 0.037  & 0.026  & 0.060  & 0.086  \\
    \hline
    \multirow{3}[0]{*}{Ours} & norm.  & 1.17  & 4.49  & 8.73  & 4.89  & 6.27  & 9.53  & 18.31  & 7.08  & 5.85  & 12.02  & 7.83\\
          & dirs. & 1.79  & 3.54  & 2.33  & 2.60  & 5.81  & 8.45  & 7.40  & 3.73  & 2.10  & 7.91  & 4.57\\
          & ints. & 0.014 & 0.011 & 0.034 & 0.023 & 0.196 & 0.045 & 0.032 	& 0.071 & 0.037 & 0.047 & 0.051 \\
    \hline
    \end{tabular}%
    }
  \label{tab:pseudo shadow}%
\end{table}%

\newpage
\section{Qualitative and Quantitative Comparison}
\label{comparison}
\subsection{Results on {\sc DiLiGenT} dataset~\cite{shi2016benchmark}}

\begin{figure}[H]
\begin{subfigure}{\linewidth}
    \centering
    \includegraphics[width=0.8\linewidth]{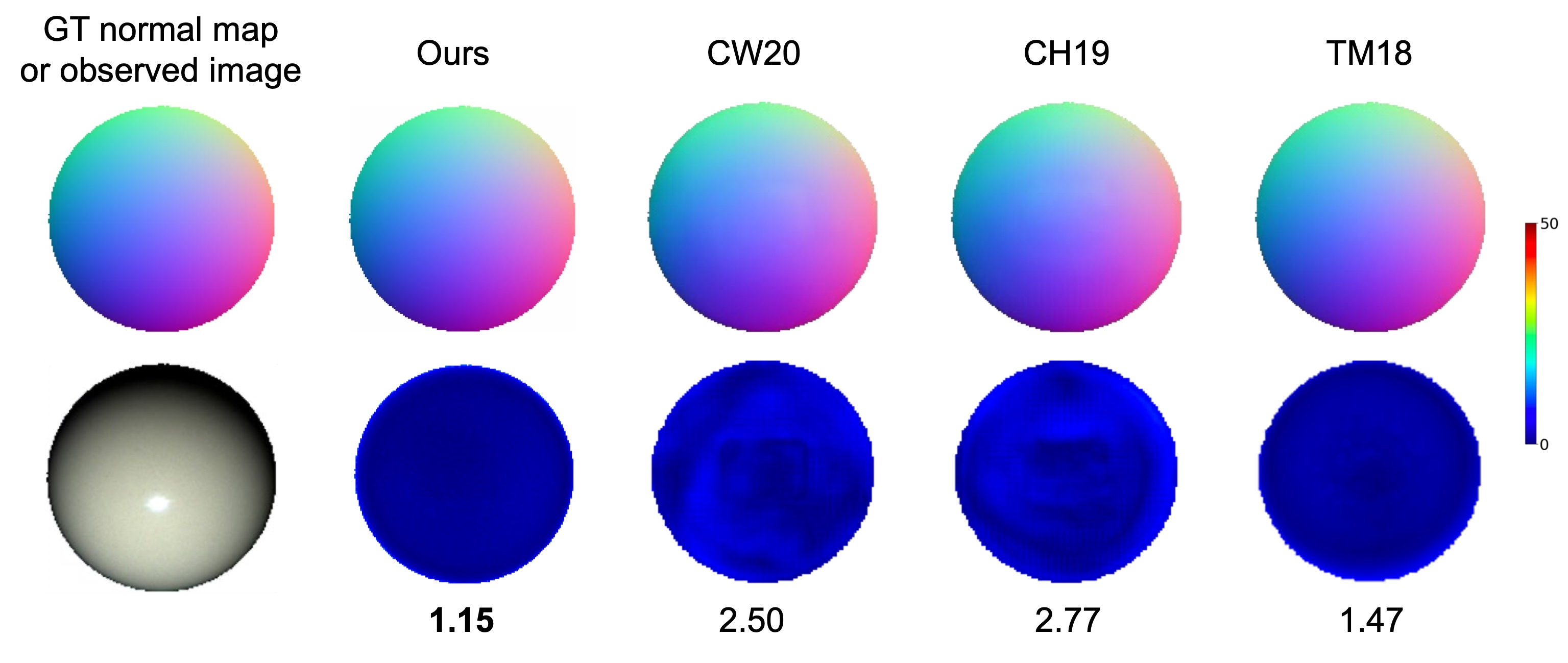}
    \caption{\sc Ball}
\end{subfigure}
\begin{subfigure}{\linewidth}
    \centering
    \includegraphics[width=0.8\linewidth]{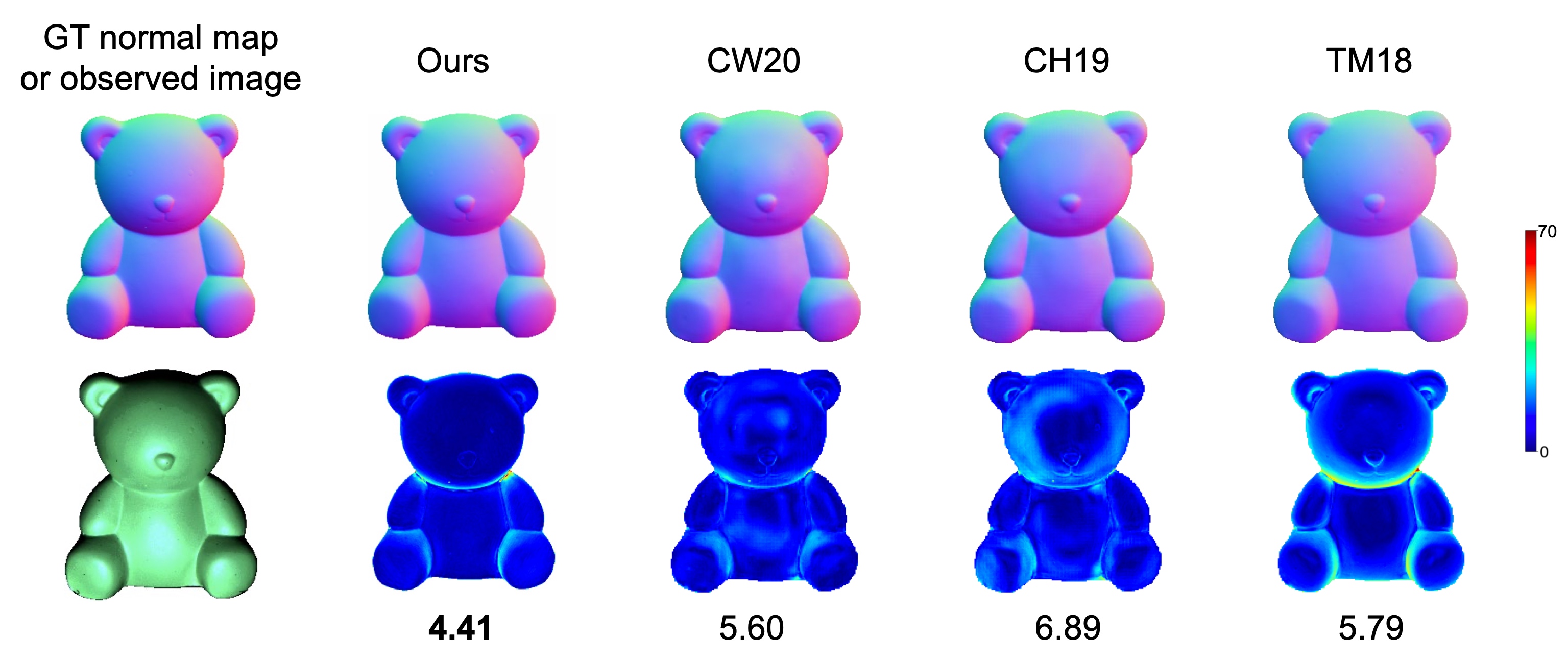}
    \caption{\sc Bear}
\end{subfigure}
\begin{subfigure}{\linewidth}
    \centering
    \includegraphics[width=0.8\linewidth]{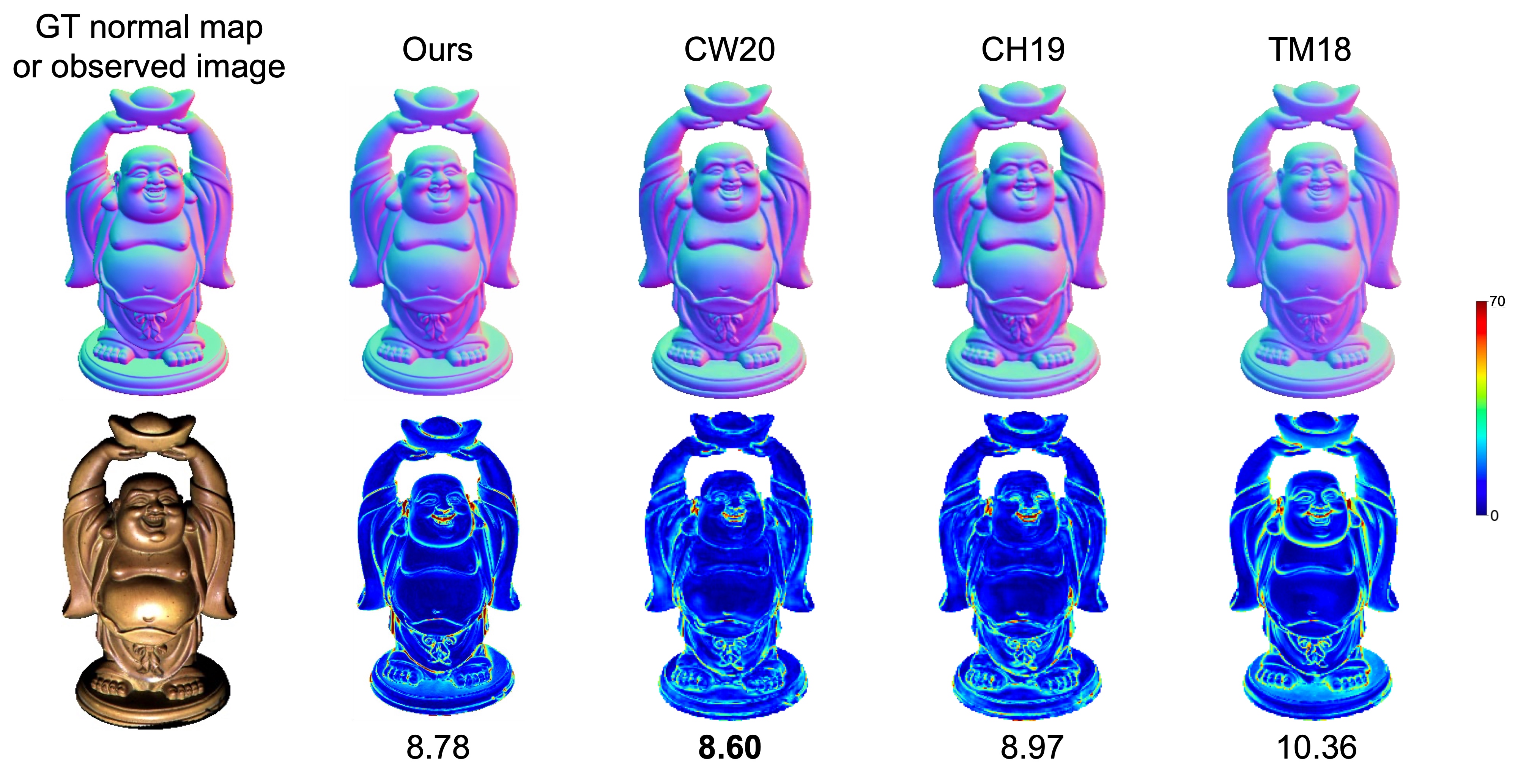}
    \caption{\sc Buddha}
\end{subfigure}
\captionsetup{justification   = raggedright, singlelinecheck = false}
\caption{Error maps and surface normal of {\sc Ball}, {\sc Bear}, {\sc Buddha}}
\label{fig:error map 1}
\end{figure}

\begin{figure}[H]
\begin{subfigure}{\linewidth}
    \centering
    \includegraphics[width=0.8\linewidth]{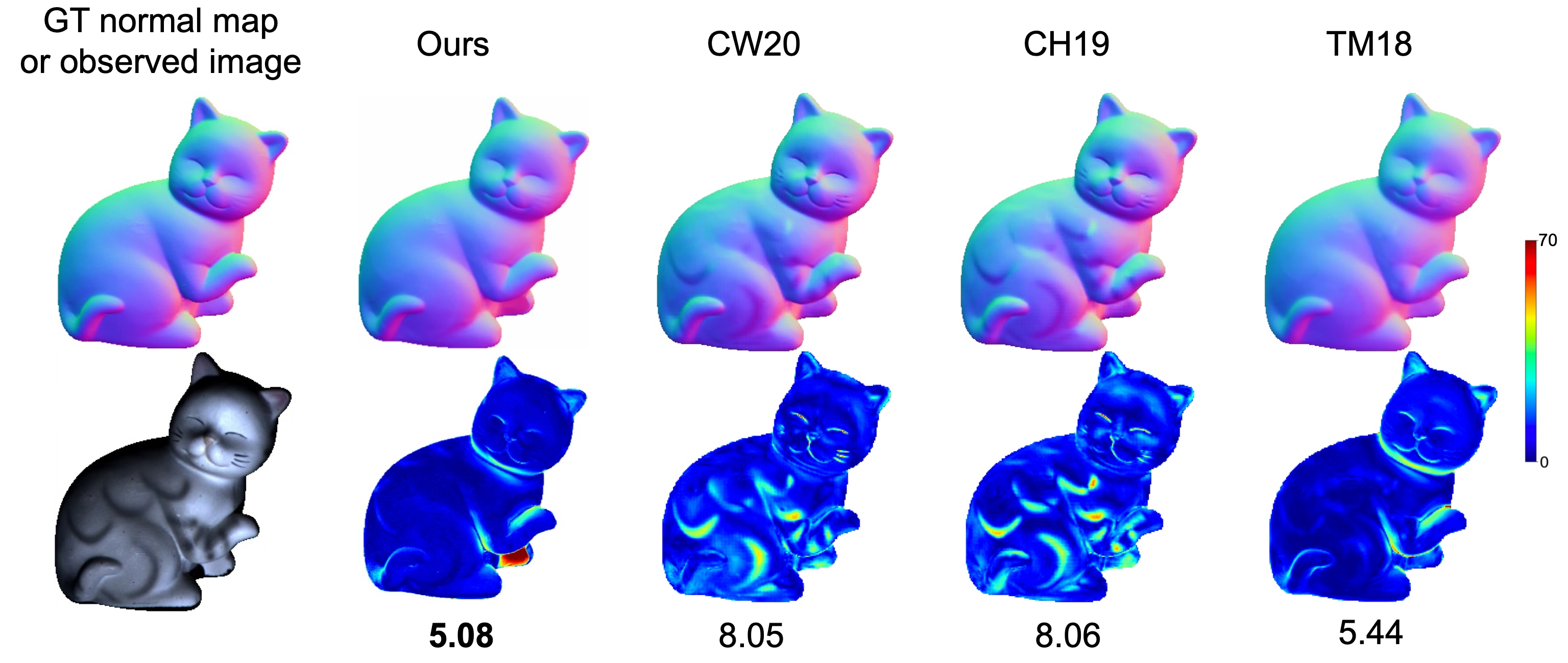}
    \caption{\sc Cat}
\end{subfigure}
\centering
\begin{subfigure}{\linewidth}
    \centering
    \includegraphics[width=0.8\linewidth]{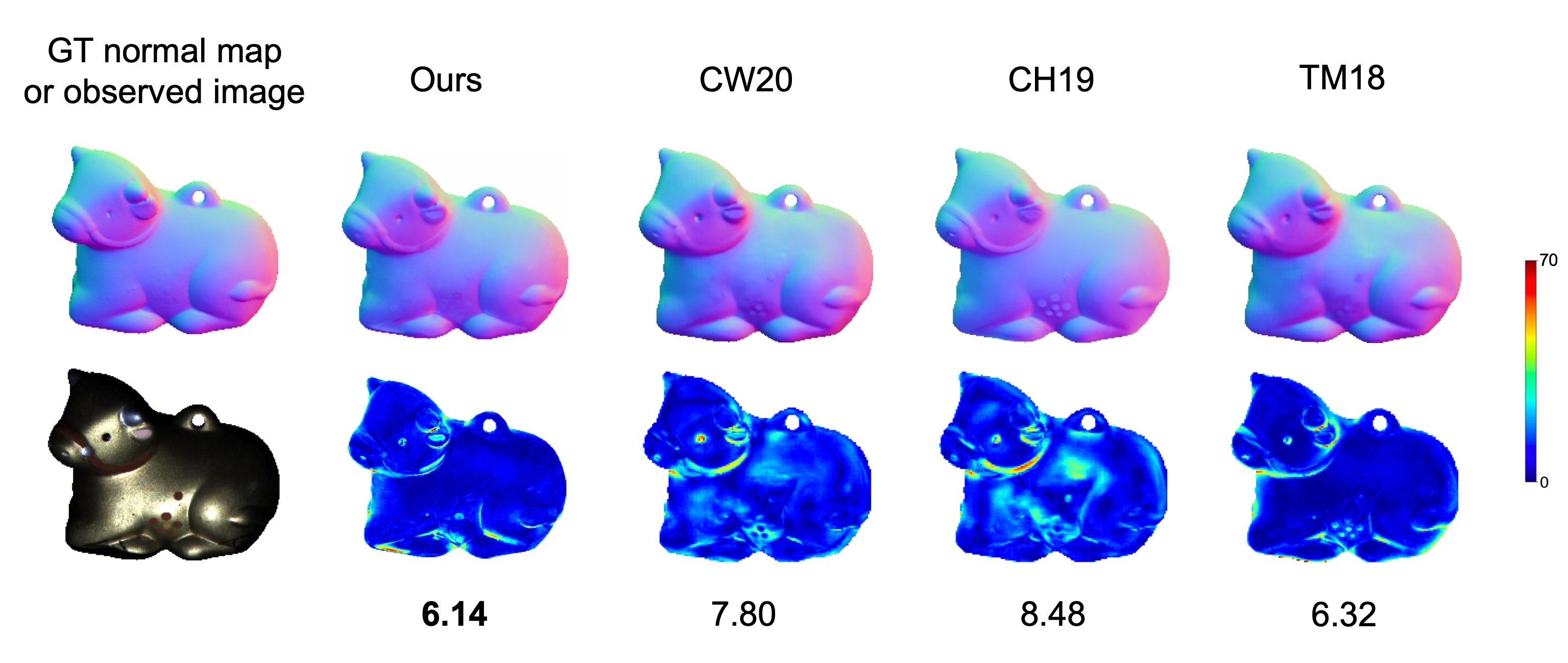}
    \caption{\sc Cow}
\end{subfigure}
\begin{subfigure}{\linewidth}
    \centering
    \includegraphics[width=0.8\linewidth]{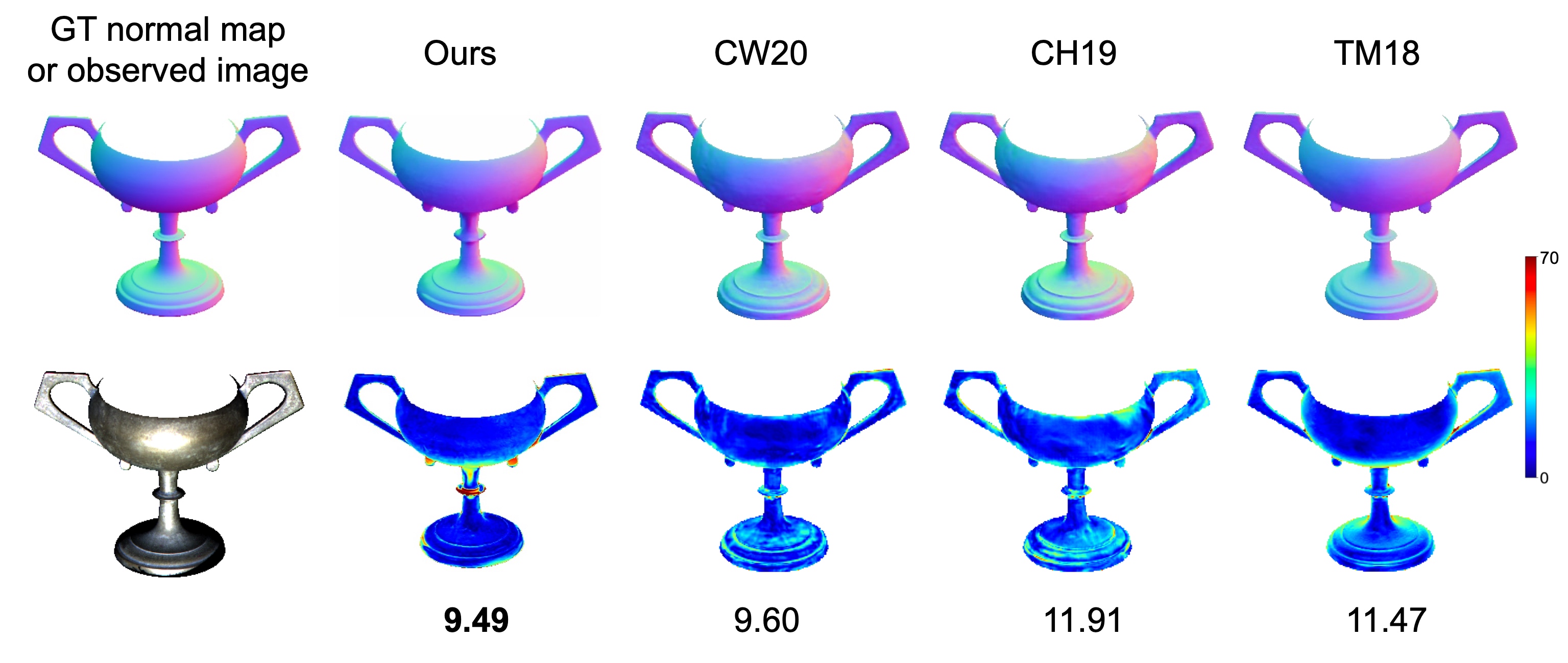}
    \caption{\sc Goblet}
\end{subfigure}
\begin{subfigure}{\linewidth}
    \centering
    \includegraphics[width=0.8\linewidth]{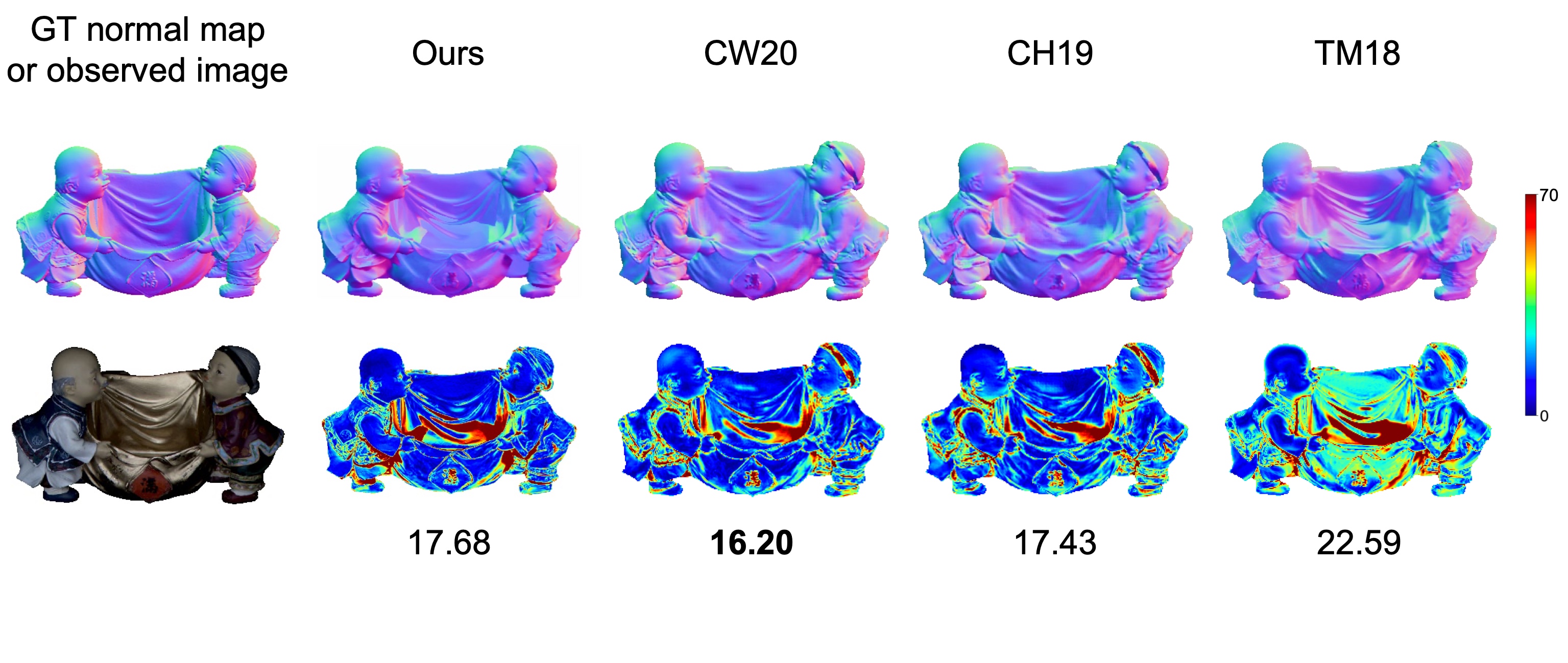}
    \caption{\sc Harvest}
\end{subfigure}
\captionsetup{justification   = raggedright, singlelinecheck = false}
\caption{Error maps and surface normal of {\sc Cat}, {\sc Cow}, {\sc Goblet} and {\sc Harvest}}
\label{fig:error map 2}
\end{figure}

\begin{figure}[H]
\centering
\begin{subfigure}{\linewidth}
    \centering
    \includegraphics[width=0.8\linewidth]{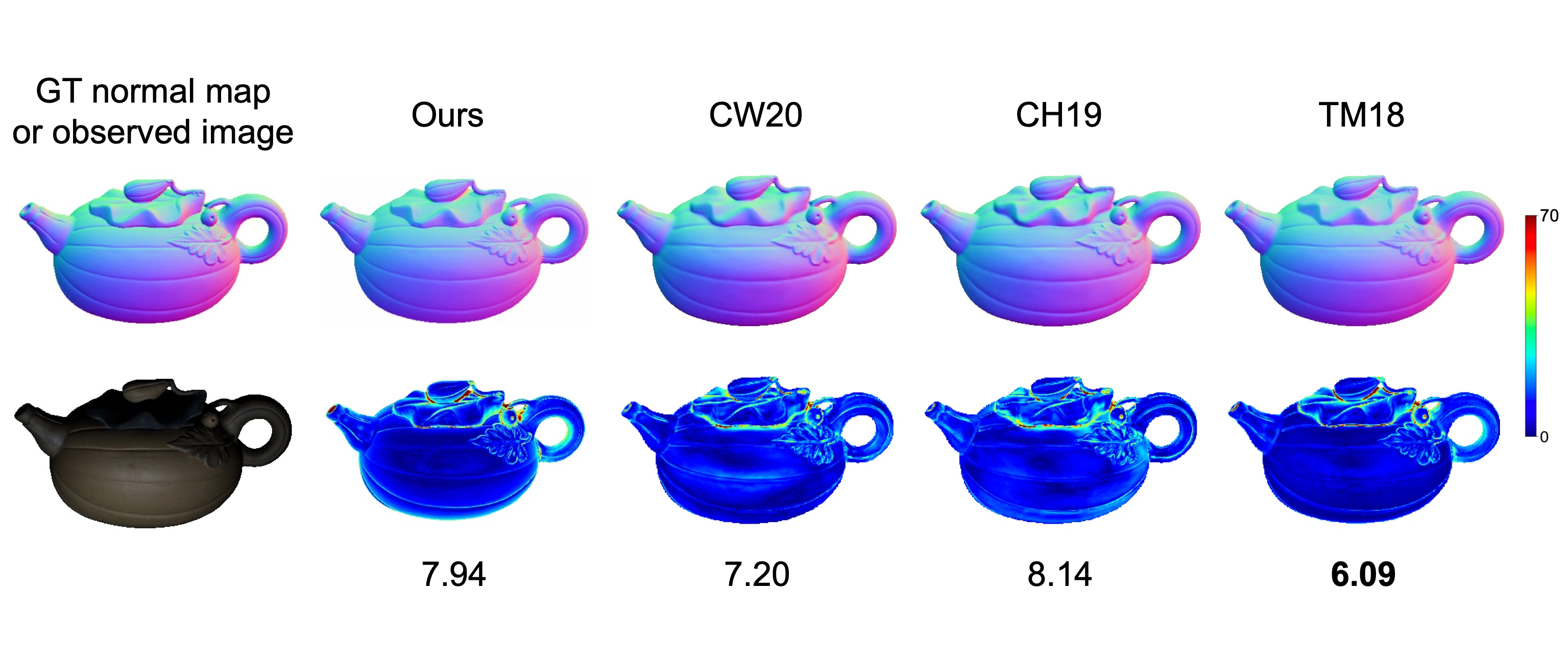}
    \caption{\sc Pot1}
\end{subfigure}
\begin{subfigure}{\linewidth}
    \centering
    \includegraphics[width=0.8\linewidth]{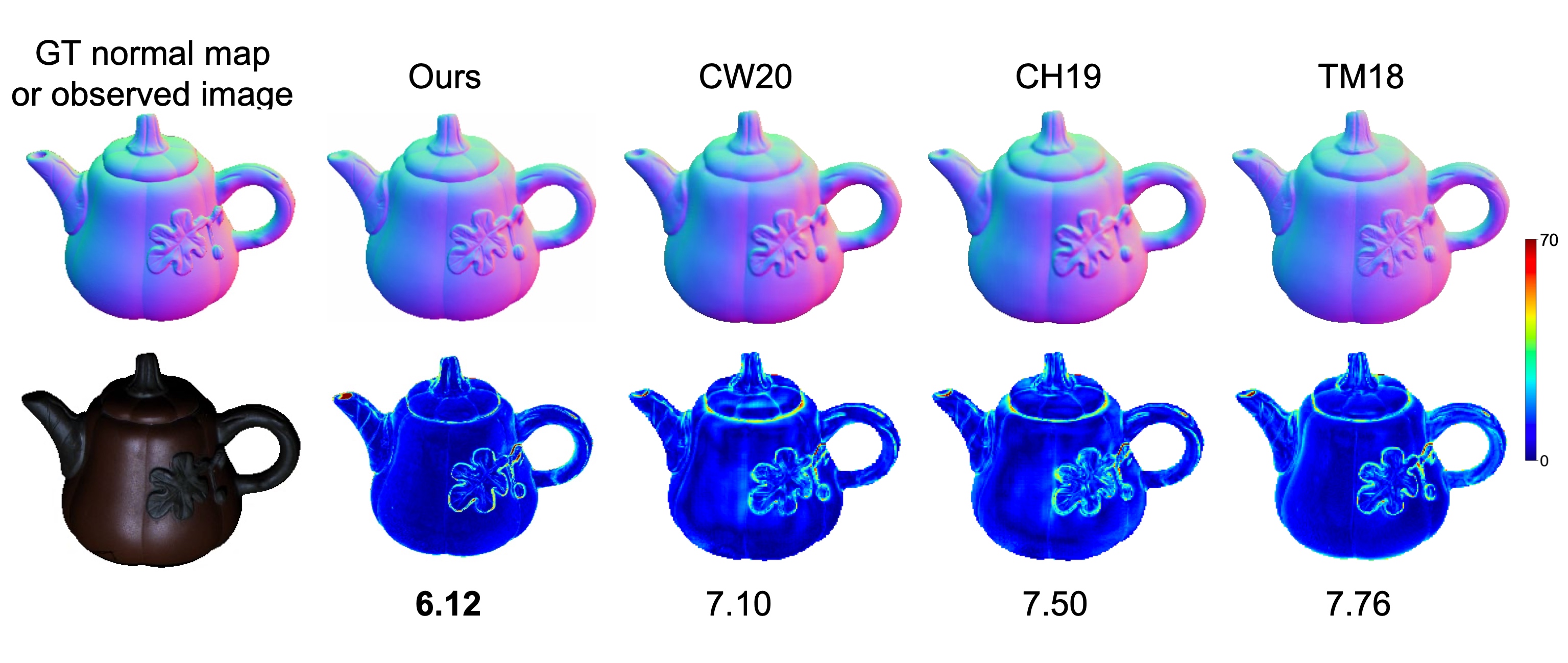}
    \caption{\sc Pot2}
\end{subfigure}
\begin{subfigure}{\linewidth}
    \centering
    \includegraphics[width=0.8\linewidth]{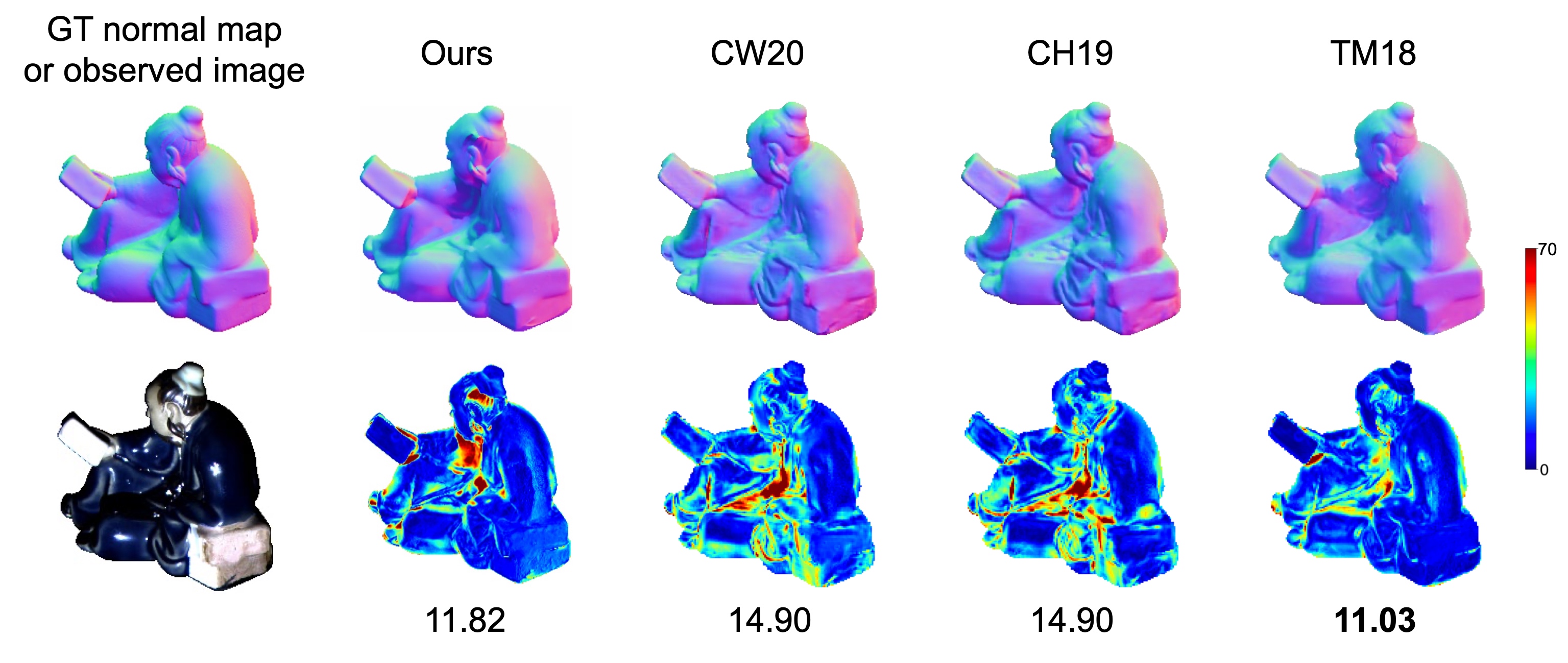}
    \caption{\sc Reading}
\end{subfigure}
\captionsetup{justification   = raggedright, singlelinecheck = false}
\caption{Error maps and surface normal of {\sc Pot1}, {\sc Pot2}, {\sc Reading}}
\label{fig:error map 3}
\end{figure}

\textbf{Surface normal.}
\fref{fig:error map 1} to \fref{fig:error map 3} show the normal and error maps comparison with CW20~\cite{chen2020learned}, CH19~\cite{chen2019self}, and TM18~\cite{taniai2018neural}. As TM18~\cite{taniai2018neural} did not provide any pre-trained models, we retrained the whole network and obtained a result that is slightly different from the original paper. Here we use the retrained models for visual quality comparison. 
The difference between the retrained and reported versions is shown in \Tref{tab:retrained}.

\newpage

\newpage

\begin{table}[h]
  \vspace{-1em}
  \centering
  \caption{Quantitative comparison in terms of mean angular error for estimated surface normal map on {\sc DiLiGenT} dataset~\cite{shi2016benchmark}. }
  \resizebox{\linewidth}{!}{
    \begin{tabular}{c|ccccccccccc}
    \hline
          & {\sc Ball} & {\sc Bear} & {\sc Buddha} & {\sc Cat} & {\sc Cow} & {\sc Goblet} & {\sc Harvest} & {\sc Pot1} & {\sc Pot2} & {\sc Reading} & AVG \\
    \hline
    TM18~\cite{taniai2018neural} (retrained) & 1.51  & 7.74  & 10.75  & 5.52  & 6.11  & 11.23  & 22.46  & 6.62  & 7.69  & 10.87  & 9.05  \\
    TM18~\cite{taniai2018neural} (reported) & 1.47  & 5.79  & 10.36  & 5.44  & 6.32  & 11.47  & 22.59  & 6.09  & 7.76  & 11.03  & 8.83  \\
    Ours & 1.17  & 4.49  & 8.73  & 4.89  & 6.27  & 9.53  & 18.31  & 7.08  & 5.85  & 12.02  & 7.83  \\
    \hline
    \end{tabular}%
    }
  \label{tab:retrained}%
  \vspace{-1em}
\end{table}%

\begin{figure}[H]
\centering
\includegraphics[width=0.8\textwidth]{"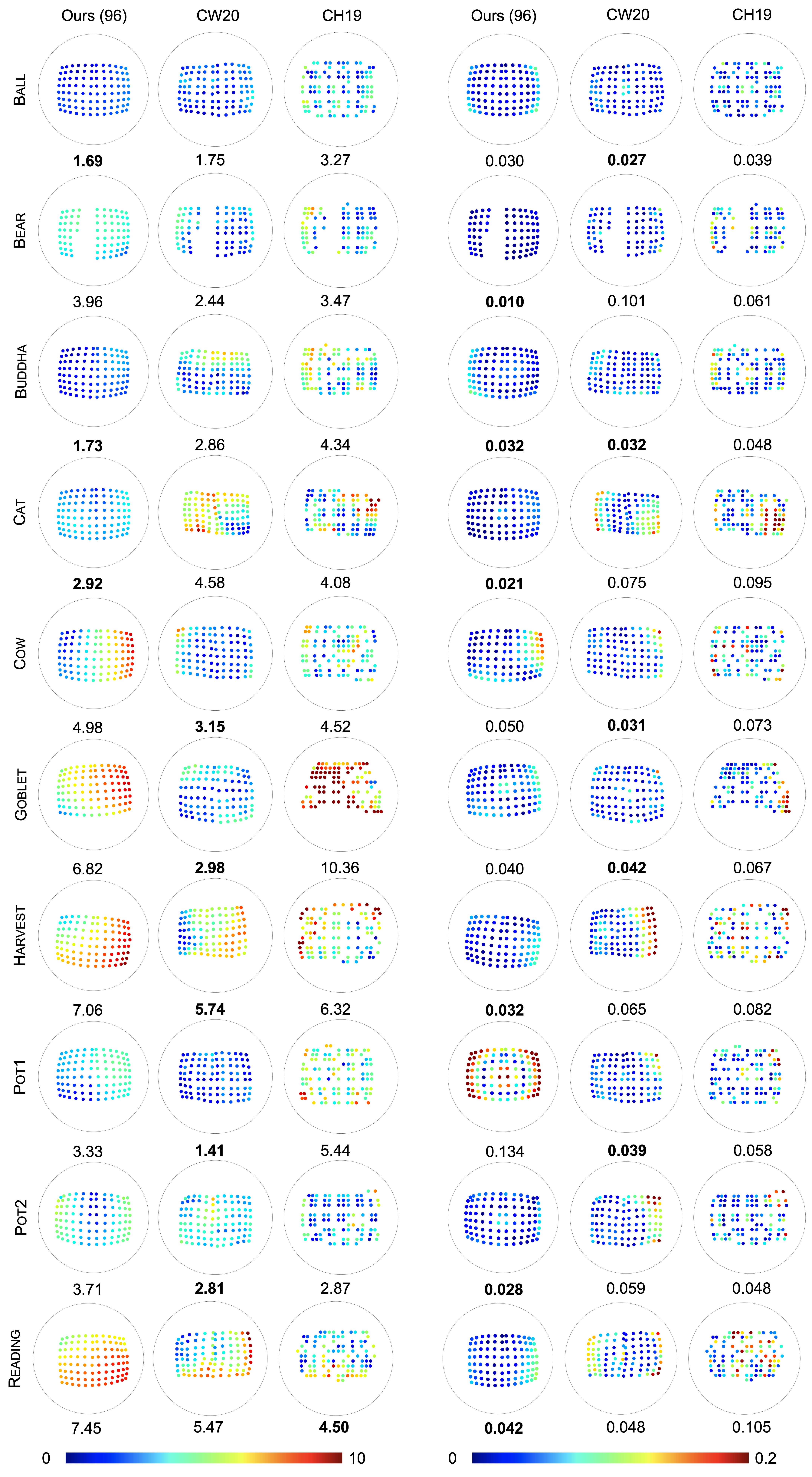"}
\captionsetup{justification   = raggedright, singlelinecheck = false}
\caption{Visual quality comparison of light direction and intensities' error maps on {\sc DiLiGenT} dataset~\cite{shi2016benchmark}.}
\label{fig:light}
\end{figure}

\newpage
\textbf{Light direction and intensity.} 
\Fref{fig:light} shows the results of light estimation for 10 objects.

\begin{figure}[H]
\centering
\includegraphics[width=0.8\textwidth]{"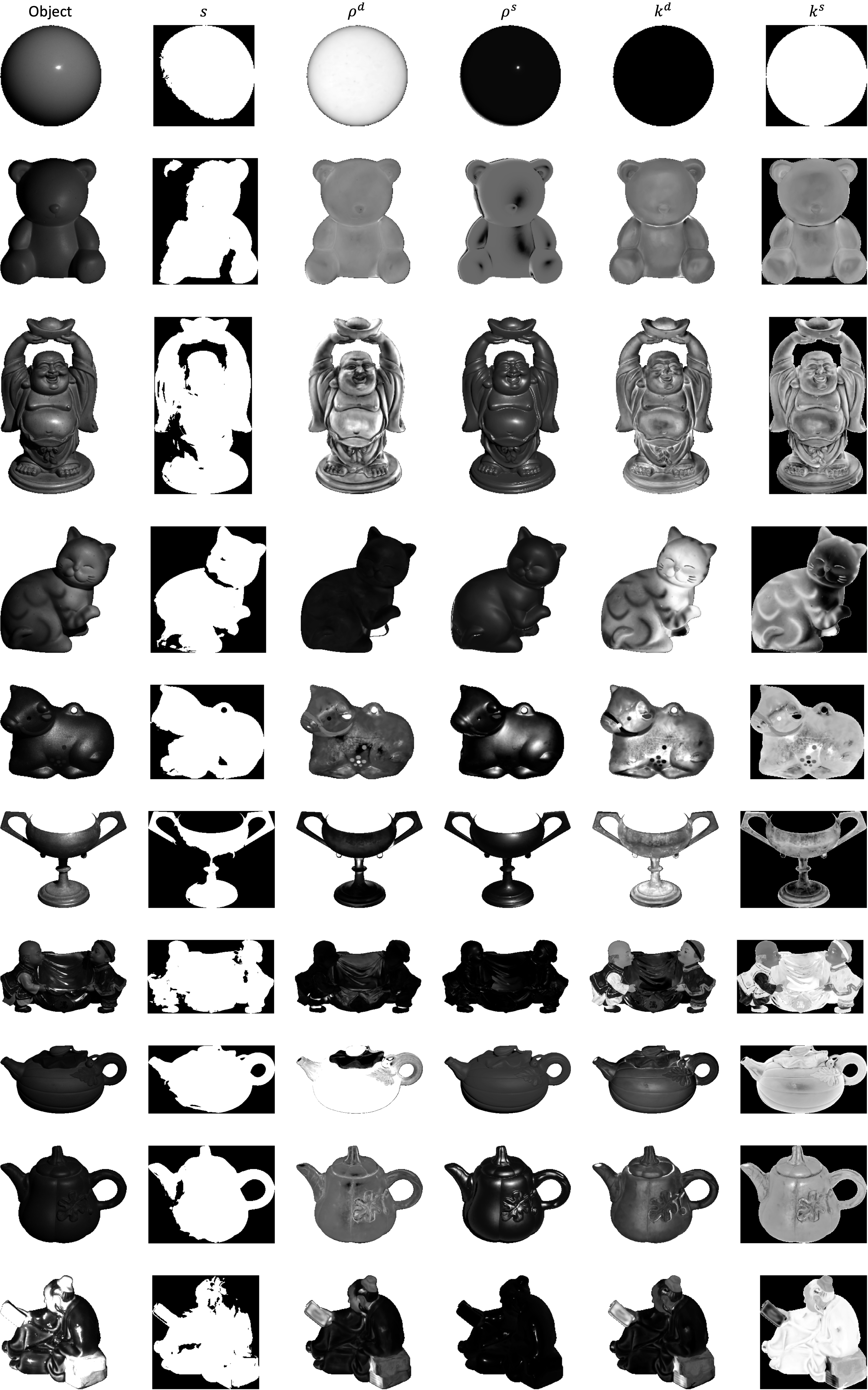"}
\captionsetup{justification   = raggedright, singlelinecheck = false}
\caption{Visual illustration of estimated intrinsics of the 10 objects from {\sc DiLiGenT} dataset~\cite{shi2016benchmark}. From left to right: images of objects, estimated shadow maps, diffuse reflectance maps, specular reflectance maps, diffuse coefficient maps, and specular coefficient maps. The diffuse reflectance maps are magnified 3 times, and all of the intrinsic maps are scaled to $[0, 1]$ for better visualization.}
\label{fig:intrinsics_1}
\end{figure}

\newpage
\textbf{Intrinsics visualization.} 
\Fref{fig:intrinsics_1} shows the results of estimated intrinsics for all 10 objects in {\sc DiLiGenT} dataset~\cite{shi2016benchmark}.

\newpage
\subsection{Results on {\sc Apple \& Gourd} dataset~\cite{alldrin2008photometric}}

\begin{figure}[h]
\centering
\includegraphics[width=\textwidth]{"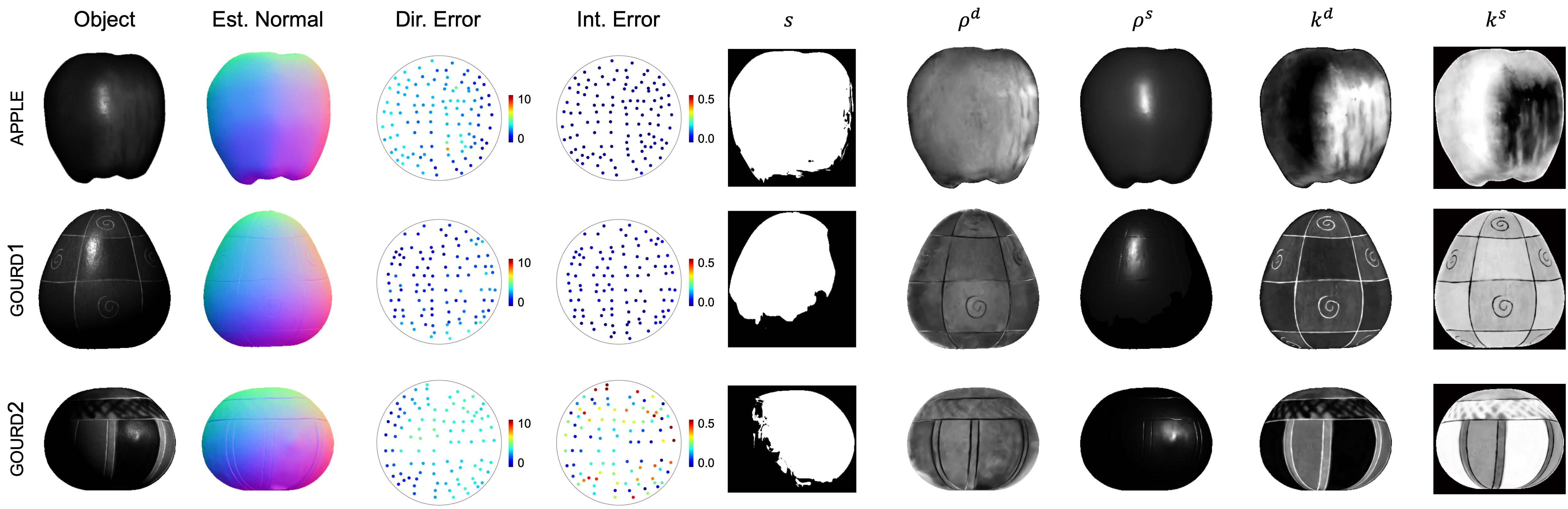"}
\captionsetup{justification   = raggedright, singlelinecheck = false}
\caption{Visual quality illustration of {\sc Standing Knight}, {\sc Helmet Side Left} and {\sc Plant} from {\sc Light Stage Data Gallery}~\cite{chabert2006relighting}. Left-right: images of objects, estimated normal maps,  light direction error maps, light intensity error maps, shadow maps, diffuse reflectance maps, specular reflectance maps, diffuse coefficient maps, and specular coefficient maps.}
\label{fig:apple}
\end{figure}

In this section, we show the intrinsics obtained by NeIF in {\sc Apple}, {\sc Gourd1} and {\sc Gourd2} in \Fref{fig:apple}. We also report the MAE of light directions in \Tref{tab:apple}. 

\begin{table}[h]
  \centering
  \caption{Quantitative comparison in terms of mean angular error for light direction and scale-invariant error for intensity on {\sc Apple \& Gourd}~\cite{alldrin2008photometric}.}
    \begin{tabular}{c|cccccc|cc}
    \hline
    & \multicolumn{2}{c}{\sc{Apple}} & \multicolumn{2}{c}{\sc{Gourd1}} & \multicolumn{2}{c|}{\sc{Gourd2}} & \multicolumn{2}{c}{AVG}\\
    Model & 
    \multicolumn{1}{c}{dirs.} & \multicolumn{1}{c}{ints.} & \multicolumn{1}{c}{dirs.} & \multicolumn{1}{c}{ints.} & \multicolumn{1}{c}{dirs.} & \multicolumn{1}{c|}{ints.} & \multicolumn{1}{c}{dirs.} & \multicolumn{1}{c}{ints.} \\
    \hline
    PF14~\cite{papadhimitri2014closed}  & 6.68  & 0.109 & 21.23 & 0.096 & 25.87 & 0.329 & 17.92 & 0.178\\
    CH19~\cite{chen2019self}   & 9.31  & 0.106 & 4.07  & 0.048 & 7.11  & \textbf{0.186} & 6.83 & 0.113\\
    CW20~\cite{chen2020learned}  & 10.91 & 0.094 & 4.29  & 0.042 & 7.13  & 0.199 & 7.44  & 0.112\\
    \hline
    Ours  & \textbf{2.65} & \textbf{0.011} & \textbf{1.76} & \textbf{0.029} & \textbf{3.21} & 0.230 & \textbf{2.54} & \textbf{0.090}\\
    \hline
    \end{tabular}%
    
  \label{tab:apple}%
\end{table}%

\subsection{Results on {\sc Light Stage Data Gallery}~\cite{chabert2006relighting}}
In this section, we show the intrinsics obtained by NeIF in {\sc Standing Knight}, {\sc Helmet Side Left} and {\sc Plant}. 
{\sc Light Stage Data Gallery} is challenging because of the complex details for each object. Although the given hyper-parameters work well in most cases, we do observe that for some cases, such as {\sc Plant}, which has complicated shapes and shadows, while with limited specular effects, the current setting may not be optimal. In such cases, we need to increase the update frequency of the reconstructed shadow maps from per epoch to per batch. 
\begin{figure}[h]
\vspace{-1em}
\centering
\includegraphics[width=\textwidth]{"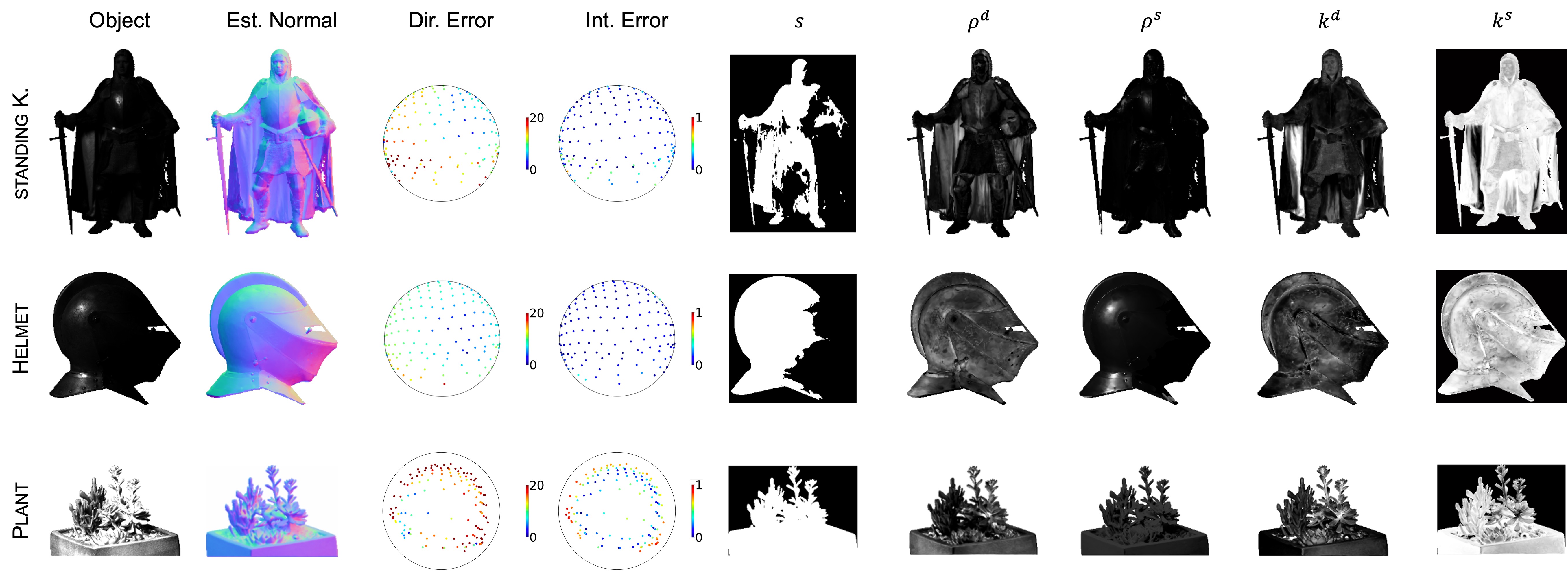"}
\captionsetup{justification   = raggedright, singlelinecheck = false}
\caption{Visual quality illustration of {\sc Standing Knight}, {\sc Helmet Side Left} and {\sc Plant} from {\sc Light Stage Data Gallery}~\cite{chabert2006relighting}. Left-right: images of objects, estimated normal maps,  light direction error maps, light intensity error maps, shadow maps, diffuse reflectance maps, specular reflectance maps, diffuse coefficient maps, and specular coefficient maps.}
\label{fig:light_stage}
\vspace{-1em}
\end{figure}
\begin{table}[h]
  \vspace{-1em}
  \centering
  \caption{Quantitative comparison in terms of light direction error and light intensity error on {\sc Standing Knight}, {\sc Helmet Side Left} and {\sc Plant} from {\sc Light Stage Data Gallery}~\cite{chabert2006relighting}.}
  
    \begin{tabular}{c|cccccc|cc}
    \hline
    & \multicolumn{2}{c}{{\sc \specialcell{Standing\\Knight}}} & \multicolumn{2}{c}{\sc Helmet} & \multicolumn{2}{c|}{\sc Plant} & \multicolumn{2}{c}{AVG} \\
    Model & 
    \multicolumn{1}{c}{dirs.} & \multicolumn{1}{c}{ints.} & \multicolumn{1}{c}{dirs.} & \multicolumn{1}{c}{ints.} & \multicolumn{1}{c}{dirs.} & \multicolumn{1}{c|}{ints.} & \multicolumn{1}{c}{dirs.} & \multicolumn{1}{c}{ints.} \\
    \hline
    PF14~\cite{papadhimitri2014closed}  & 33.81 & 1.311 & 25.4  & 0.576 & 20.56 & 0.227 & 26.59  & 0.705  \\
    CH19~\cite{chen2019self}  & 11.6  & 0.286 & 6.57  & 0.212 & 16.06 & 0.17  & 11.41  & 0.223  \\
    CW20~\cite{chen2020learned}  & \textbf{5.31} & 0.198 & \textbf{5.33} & 0.096 & \textbf{10.49} & \textbf{0.154} & \textbf{7.04 } & \textbf{0.149 } \\
    \hline
    Ours  & 13.38 & \textbf{0.189} & 8.12  & \textbf{0.082} & 18.04 & 0.464 & 13.18  & 0.245  \\
    \hline
    \end{tabular}%
    
  \label{tab:light_stage}%
\end{table}%
\section{Scale on the Input Images}
\label{random scale}
\begin{figure}[h]
\centering
\includegraphics[width=0.3\textwidth]{"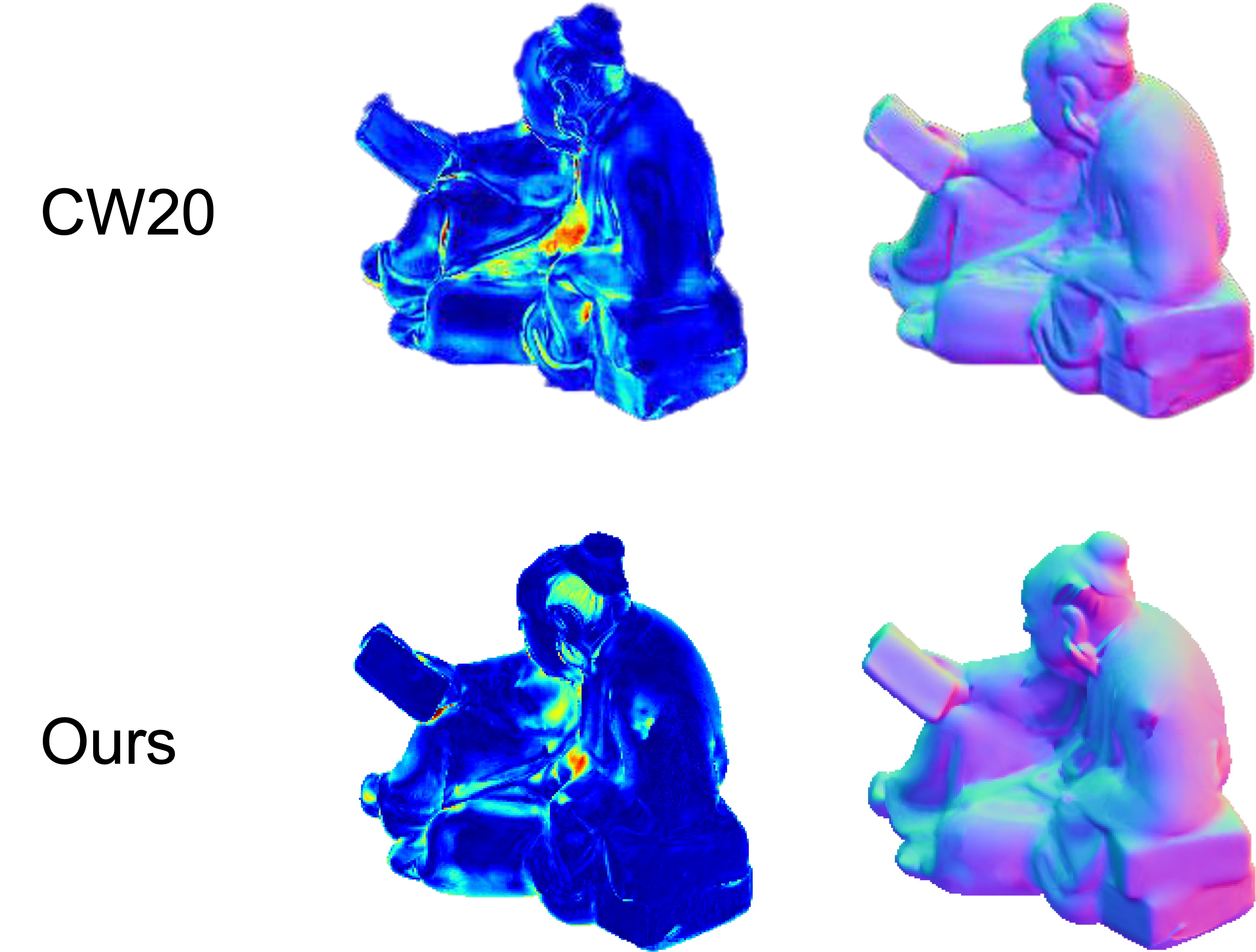"} 
\caption{Visual illustration of error maps of the normal. From left to right: error maps of {\sc Reading}, normal map of {\sc Reading}. Top-bottom: our result and CW20~\cite{chen2020learned}'s result.}
\label{fig:error map}
\end{figure}

The intensity of 96 images from {\sc Reading} are scaled by 96 variables generated by the the uniform distribution $\mathcal{U}(0.01, 1)$, respectively.
The new images are used as the training data to train our method.
The results on {\sc Reading} is shown in \Fref{fig:error map}, which further illustrates that our method can handle the varying light intensity and is free from data bias. While CW20~\cite{chen2020learned} (15.84 v.s. ours 11.05) fails because they have a pre-defined range on the intensities. 

\newpage
\section{Comparison on Data Collected under Casual Environments}
\label{NLUPS with ambient}

To further highlight the advantage of our method regarding unsupervised learning and light-normal joint optimization, we perform an experimental comparison on data collected in casual environments.

\begin{figure}[h]
\centering
\includegraphics[width=0.7\textwidth]{"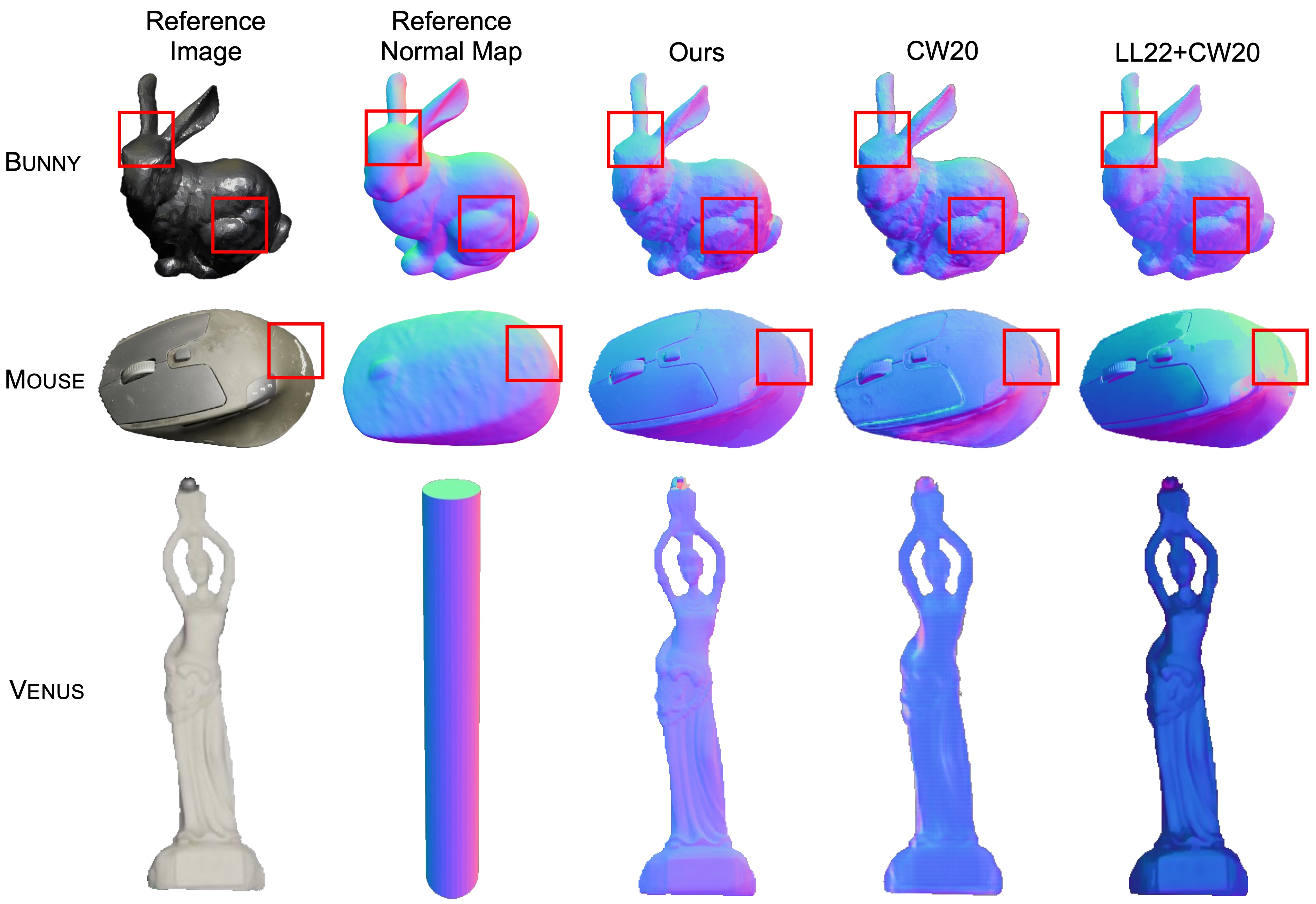"} 
\caption{Visual quality comparison in terms of normal map on {\sc Bunny} (row 1), {\sc Mouse} (row 2), and {\sc Venus} (row 3) from {\sc DiLiGenT}~\cite{shi2016benchmark}. For each subfig, from left to right: mean of the observed images, coarse normal map download in \url{https://sketchfab.com/}, that has similar shape with the objects for reference, normal map from our method, CW20~\cite{chen2020learned}, LL22~\cite{li2022neural} + CW20~\cite{chen2020learned}.} 
\label{fig:visual_natural_2}
\end{figure}

\textbf{Captured objects.}
We use three objects for this experiment, including {\sc Bunny}, {\sc Venus}, and {\sc Mouse}. {\sc Bunny} contains lots of fine details, with a broad specular lobe on a uniform material (phenolic resin).
{\sc Venus} is made up of glass for the pearl on the tray and gypsum for the body.
{\sc Mouse} has lots of defects on the shell, made up of a spatially varying but mostly diffuse material. 
The illustration of these objects can be found in \Fref{fig:visual_natural_2}.

\begin{figure}[h]
\centering
\includegraphics[width=0.7\textwidth]{"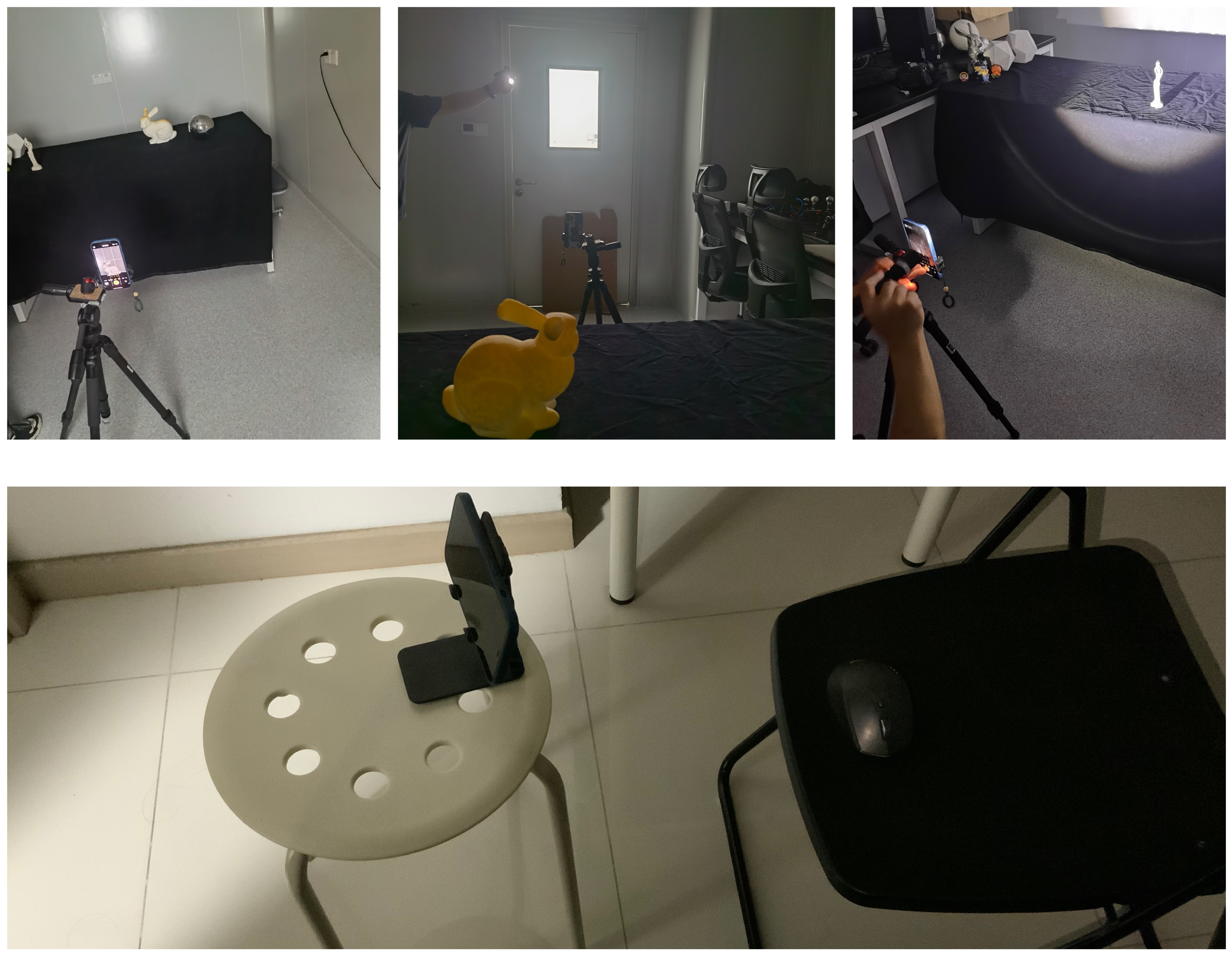"} 
\caption{Data collection process. From top to the bottom, Row 1: the equipment we use in scene 1 to capture the {\sc Venus} and {\sc Bunny}; Row 2: the equipment we capture {\sc Mouse} in scene 2.} 
\label{fig:scene}
\vspace{-1em}
\end{figure}

\textbf{Data capture}. We use the iPhone 13 Pro Max camera fixed on a phone tripod for data capture. 
We turn the HDR mode on and set the exposure compensation to -2 to avoid overexposure. We record a video with a raw resolution at $1920 \times 1080$. During video recording, we move the light source slowly to illuminate the object at different angles and keep the trajectory parallel to the image plane. 
\Fref{fig:scene} shows the capture equipment.
{\sc Bunny} and {\sc Venus} are shot at about 1 meter away with the light at about 1-1.5 meters away.
 {\sc Mouse} is shot at about 30cm away with light at about 40cm away. 
 
\textbf{Ambient light and light sources}. 
All data are captured with ambient light because the full-dark chamber is also expensive for casual users. 
For the data capture of {\sc Bunny} and {\sc Venus}, we control the impact of the ambient light by changing the intensity of the electric torch, \ie, weak impact (or strong light source intensity) for {\sc Bunny}, and strong impact (or weak light source intensity) for {\sc Venus}.
The capture environment for these objects is relatively controlled, and the ambient light is from a small window (see \Fref{fig:scene}).
For the data capture of {\sc Mouse}, we consider a more casual and challenging scenario (see \Fref{fig:scene}). That is, the ambient light is more dominant and uncontrolled. We use a flashlight from the mobile phone to further relax the assumption of directional light in photometric stereo. 

\textbf{Data processing.} We use MATLAB 2020 to extract 100 frames uniformly from each video and downsample them to $960 \times 540$. We extract the mask of each object via PhotoShop 2020. 

\textbf{Comparison methods.}
We compare our method with CW20~\cite{chen2020learned}.
As LL22~\cite{li2022neural} is not for UPS, we feed it by the estimated light from the state-of-the-art method CW20~\cite{chen2020learned} and denoted it as CW20~\cite{chen2020learned}+LL22~\cite{li2022neural}. 

\subsection{Comparison of Surface Normal}
The estimated normal map for each object is shown in \Fref{fig:visual_natural}. However, CW20~\cite{chen2020learned} and CW20~\cite{chen2020learned}+LL22~\cite{li2022neural} is sensitive to the data bias in supervised learning of light estimation model.
The error of estimated light dramatically degrades the performance of normal estimation.
In contrast, our method is robust to casual environments.
We have a more reasonable estimation on {\sc Bunny} with the ambient light and {\sc Mouse} in the challenging scene and much better results on {\sc Venus} as compared with CW20~\cite{chen2020learned} and CW20~\cite{chen2020learned}+LL22~\cite{li2022neural}.
Since our model adopts the directional light assumption, it produces inaccurate light. 
The superior performance advantage on surface normal estimation indicates that our method could better balance the accuracy of the surface normal and light due to its light-normal joint optimization manner.

\subsection{Comparison of Light Estimation}
\begin{figure}[h]
\centering
\includegraphics[width=0.4\textwidth]{"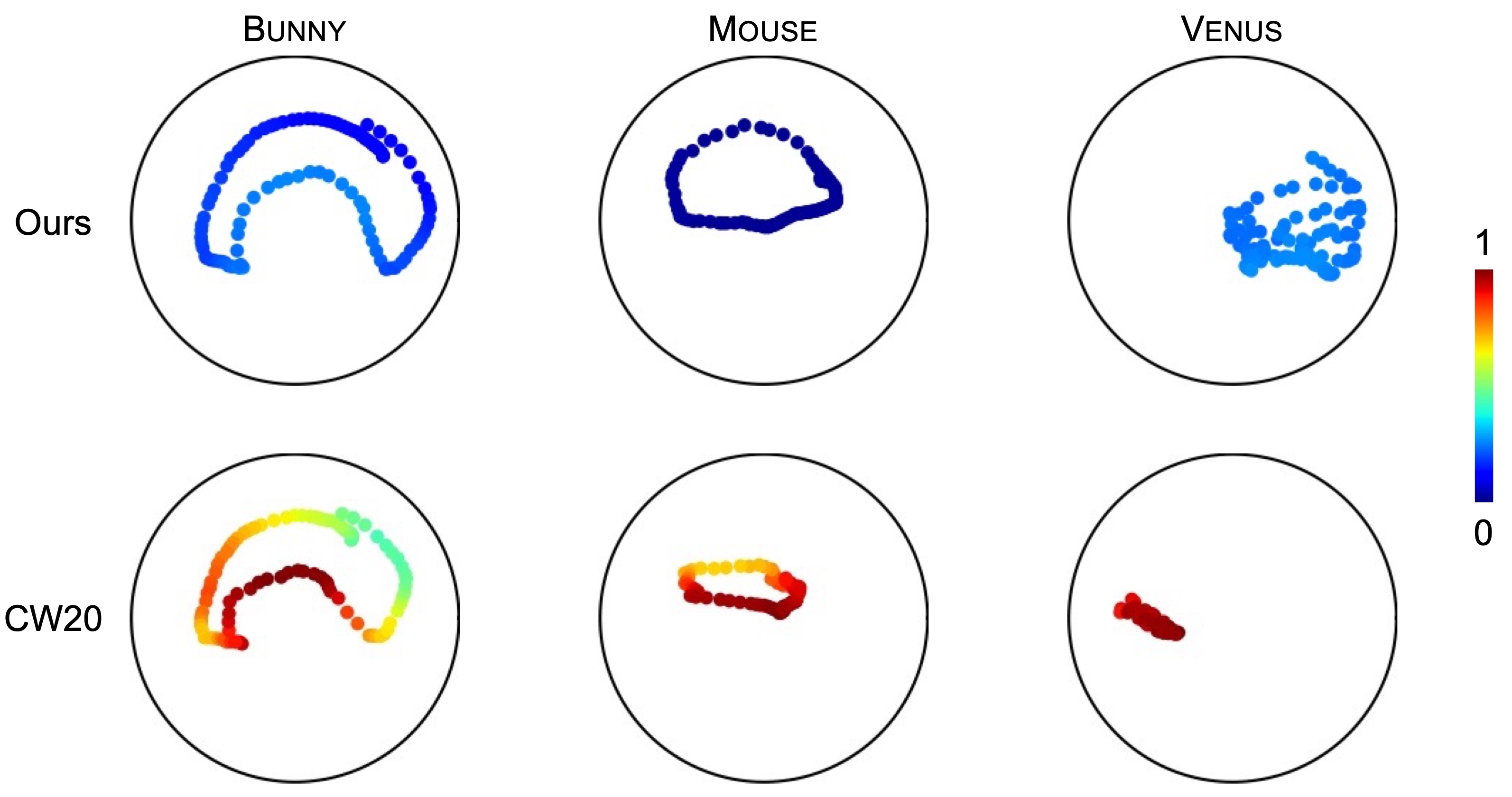"} 
\caption{Visual comparison in terms of the estimated light trajectory on {\sc Bunny}(column 1), {\sc Mouse} (column 2), and {\sc Venus} (column 3). For each sub-figure, top-bottom: Ours, CW20~\cite{chen2020learned}'s predicted light trajectory that projected onto the XY-plane, the color indicates the value of the light intensity.} 
\label{fig:light_trajectory}
\vspace{-1em}
\end{figure}

We further make a comparison on the estimation of light. The predicted light direction projected on the XY-plane is shown in \Fref{fig:light_trajectory} and represents the estimated light intensity by color (note that the light intensity is normalized for better visualization).
Since the intensity of our light source during data collection is unchanged, our method produces much more accurate light intensity than the comparison methods.
Besides, we change the light direction by moving the light source at a large distance. The trajectory recovered by our method is more reasonable than the comparison methods, especially for objects of {\sc Mouse} and {\sc Venus}.
These results indicate that our method is more robust to training data bias due to its unsupervised manner.

\newpage
\section{Sparse Uncalibrated Photometric Stereo}
\label{"tag:sparse ups"}
\subsection{Quantitative Results on {\sc DiLiGenT} dataset~\cite{shi2016benchmark}}
We evaluate the proposed NeIF for sparse uncalibrated photometric stereo. 
We use images under sparse lights (10 and 16 randomly selected lights with different seeds) for normal and light estimation. 

\begin{figure}[h] 
\vspace{-1em}
\centering
\includegraphics[width=0.8\textwidth]{"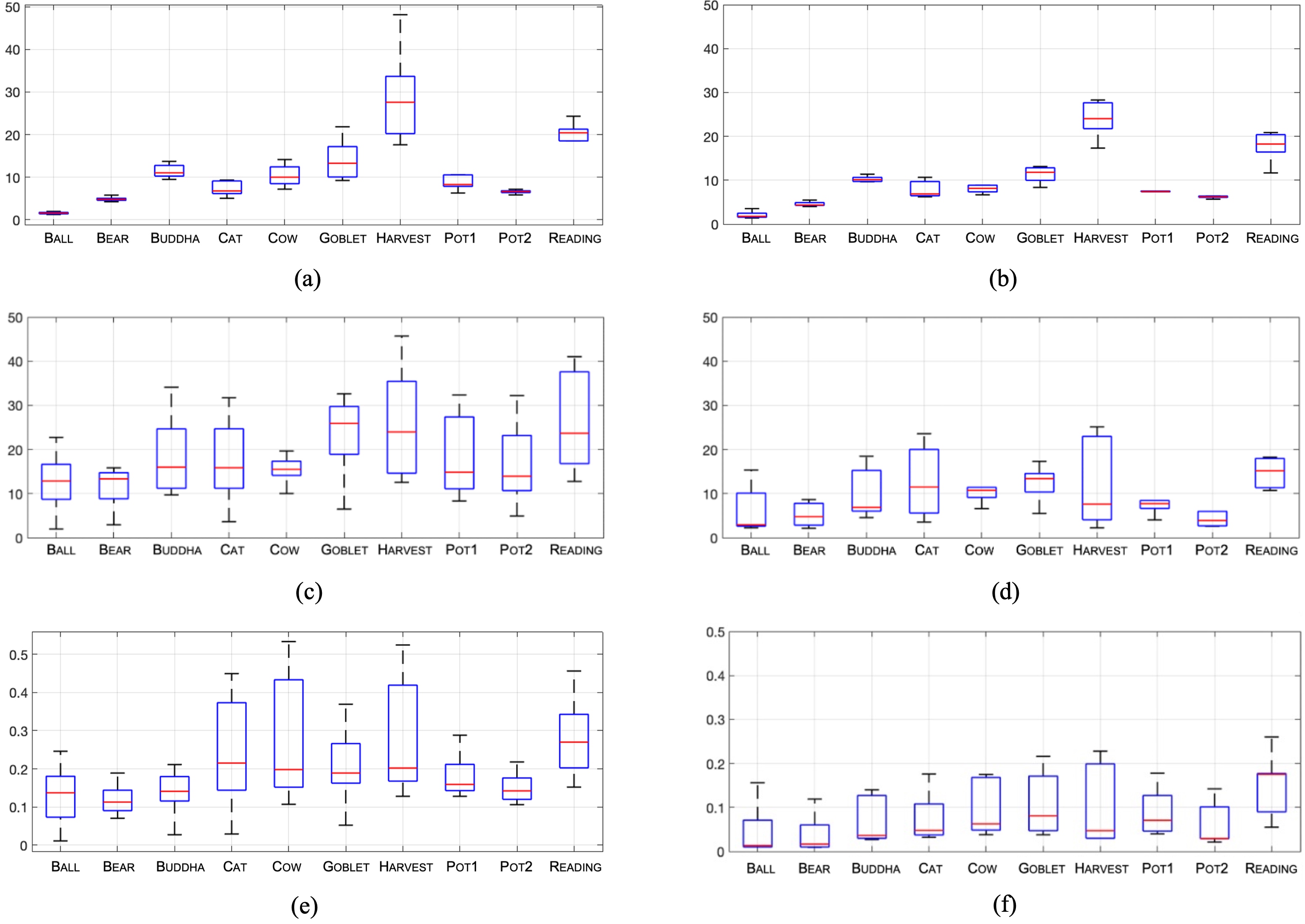"} 
\caption{Estimation results of NeIF given only 10 lights ((a), (c), (e)) and 16 lights ((b), (d), (f)) on {\sc DiLiGenT}~\cite{shi2016benchmark} benchmark regarding surface normal ((a), (b)), light direction ((c), (d)), and light intensity ((e), (f)). 
The $X$-axis indicates object names.
The $Y$-axis is the MAE (mean angular error, in degree) of the surface normal, light direction, or scale-invariant relative error of intensities, displayed using the box-and-whisker plot. The red line indicates the median value, the upper and the lower bound of the blue box are the first and third quartile values, respectively, and the top and bottom end of the black lines indicate the minimum and maximum.}
\label{fig:sparse_light}
\vspace{-1em}
\end{figure}

The results of 30 random trials are shown in \Fref{fig:sparse_light} and \Tref{tab:sparse_light}. As can be found, there is a large fluctuation of NeIF's performance for normal estimation on {\sc Harvest} that contains a large region of cast shadow under most lights, which has the most difficult geometries to estimate due to the limited information (\ie, the impact of the cast shadow is more significant under sparse lights). Compared to the calibrated and supervised method~\cite{zheng2019spline} and uncalibrated method~\cite{papadhimitri2014closed}, NeIF shows a promising performance on average for 10 lights (11.62 for normal estimation, 19.29 for light direction, and 0.199 for intensities) and 16 lights (10.38 for normal estimation, 7.21 for light direction, and 0.047 for intensities).
\begin{table}[h]
  \vspace{-1em}
  \centering
  \caption{Quantitative comparison in terms of normal map error, light direction error, and light intensity error on {\sc DiLiGenT} dataset~\cite{shi2016benchmark}. The numbers in brackets indicate the number of lights. For example, Ours (10) indicates that our method takes 10 images under different lights as the input.}
  \resizebox{\linewidth}{!}{
    \begin{tabular}{c|cccccccccccc}
    \hline
          &       & \multicolumn{1}{c}{\sc Ball} & \multicolumn{1}{c}{\sc Bear} & \multicolumn{1}{c}{\sc Buddha} & \multicolumn{1}{c}{\sc Cat} & \multicolumn{1}{c}{\sc Cow} & \multicolumn{1}{c}{\sc Goblet} & \multicolumn{1}{c}{\sc Harvest} & \multicolumn{1}{c}{\sc Pot1} & \multicolumn{1}{c}{\sc Pot2} & \multicolumn{1}{c}{\sc Reading} & \multicolumn{1}{c}{AVG} \\
    \hline
    ZJ19 (10)~\cite{zheng2019spline} \dag& norm.  & 4.96  & 5.99  & 10.07  & 7.52  & 8.80  & 10.43  & 19.05  & 8.77  & 11.79  & 16.13  & 10.35  \\
    ZJ19 (10)~\cite{zheng2019spline} \ddag& norm.  & 4.38  & 5.79  & 9.60  & 7.13  & 7.87  & 10.00 & 18.35  & 8.41  & 11.20  & 15.45  & 9.82  \\
    \hline
    \multirow{3}[0]{*}{Ours (10)} & norm.  & 2.49  & 4.90  & 11.58  & 7.47  & 10.21  & 14.45  & 27.30  & 11.41  & 6.57  & 19.85  & 11.62  \\
          & dirs.  & 14.31  & 13.38  & 18.32  & 18.37  & 16.48  & 24.43  & 25.59  & 22.80  & 17.14  & 26.60  & 19.74  \\
          & ints.  & 0.145  & 0.121  & 0.150  & 0.232  & 0.269  & 0.244  & 0.272  & 0.182  & 0.139  & 0.299  & 0.205  \\
    \multirow{3}[0]{*}{Ours (16)} & norm.  & 2.34  & 4.52  & 10.33  & 7.87  & 7.68  & 11.23  & 25.07  & 8.00  & 6.35  & 20.42  & 10.38  \\
      & dirs.  & 2.83  & 4.14  & 5.77  & 8.45  & 8.91  & 10.09  & 6.44  & 6.16  & 3.50  & 15.85  & 7.21  \\
      & ints.  & 0.012  & 0.014  & 0.033  & 0.038  & 0.047  & 0.062  & 0.041  & 0.062  & 0.026  & 0.136  & 0.047  \\
    \hline
    \multirow{3}[0]{*}{Ours (96)} & norm.  & 1.17  & 4.49  & 8.73  & 4.89  & 6.27  & 9.53  & 18.31  & 7.08  & 5.85  & 12.02  & 7.83\\
          & dirs. & 1.79  & 3.54  & 2.33  & 2.60  & 5.81  & 8.45  & 7.40  & 3.73  & 2.10  & 7.91  & 4.57\\
          & ints. & 0.014 & 0.011 & 0.034 & 0.023 & 0.196 & 0.045 & 0.032 	& 0.071 & 0.037 & 0.047 & 0.051 \\
    \hline
    \multicolumn{12}{l}{\small \dag Paper reported version, using different lights compared with us.}\\
    \multicolumn{12}{l}{\small \ddag Re-test version, using same lights compared with us.}\\
    \end{tabular}%
    }
  \label{tab:sparse_light}%
  \vspace{-2em}
\end{table}%

\newpage
\subsection{Qualitative Results on {\sc DiLiGenT} dataset~\cite{shi2016benchmark}}
\begin{figure}[H]
\centering
\begin{subfigure}{\linewidth}
    \centering
    \includegraphics[width=\linewidth]{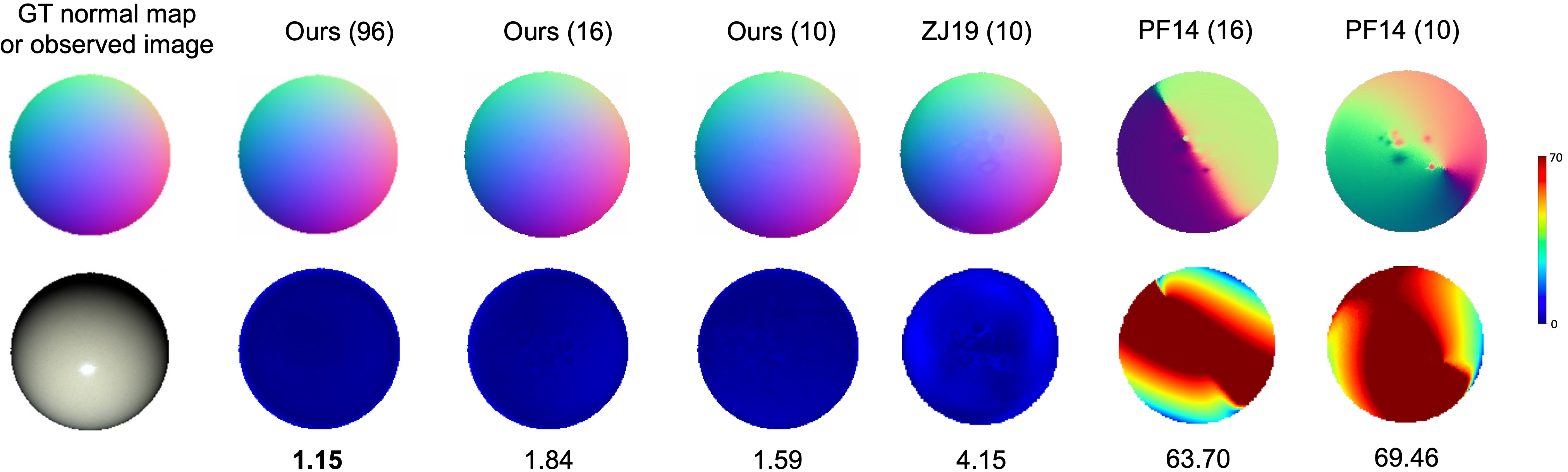}
    \caption{\sc Ball}
\end{subfigure}
\begin{subfigure}{\linewidth}
    \centering
    \includegraphics[width=\linewidth]{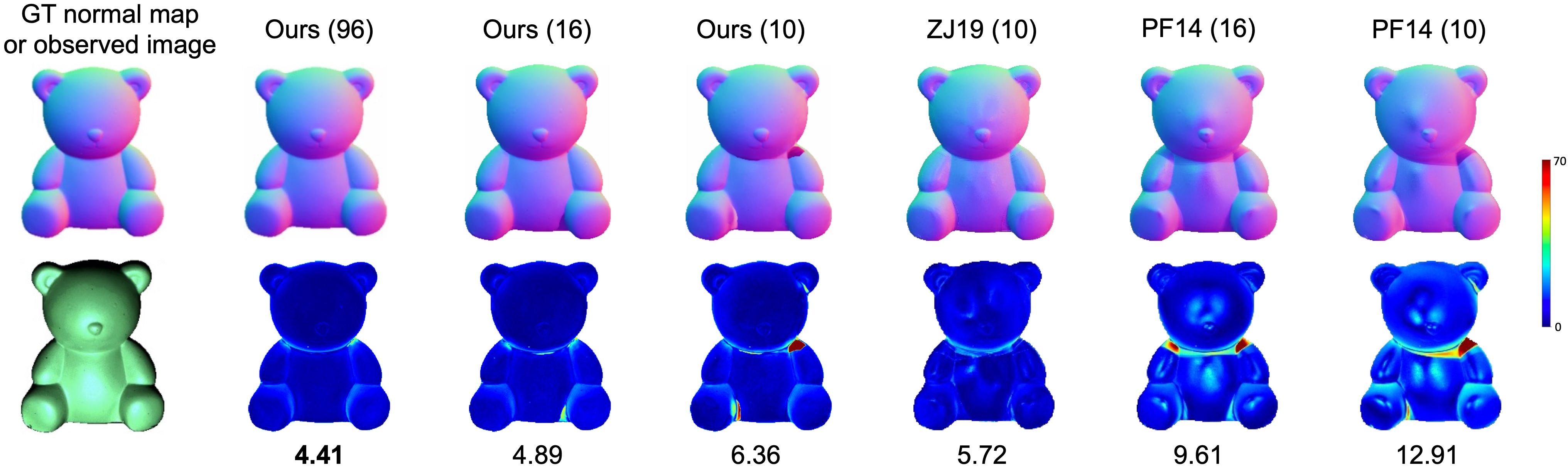}
    \caption{\sc Bear}
\end{subfigure}
\begin{subfigure}{\linewidth}
    \centering
    \includegraphics[width=\linewidth]{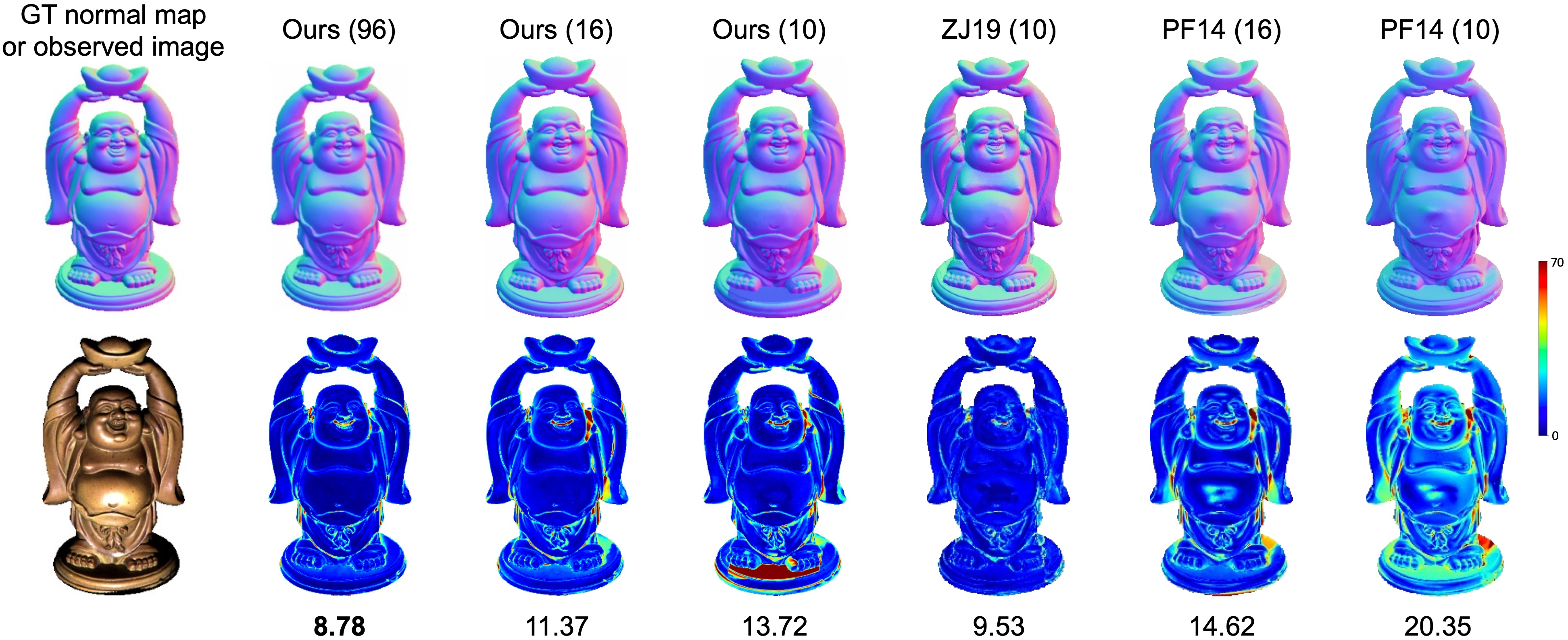}
    \caption{\sc Buddha}
\end{subfigure}
\caption{Error maps and surface normal of {\sc Ball}, {\sc Bear}, {\sc Buddha}}
\label{fig:sparse_1}
\end{figure}

\begin{figure}[H]
\begin{subfigure}{\linewidth}
    \centering
    \includegraphics[width=\linewidth]{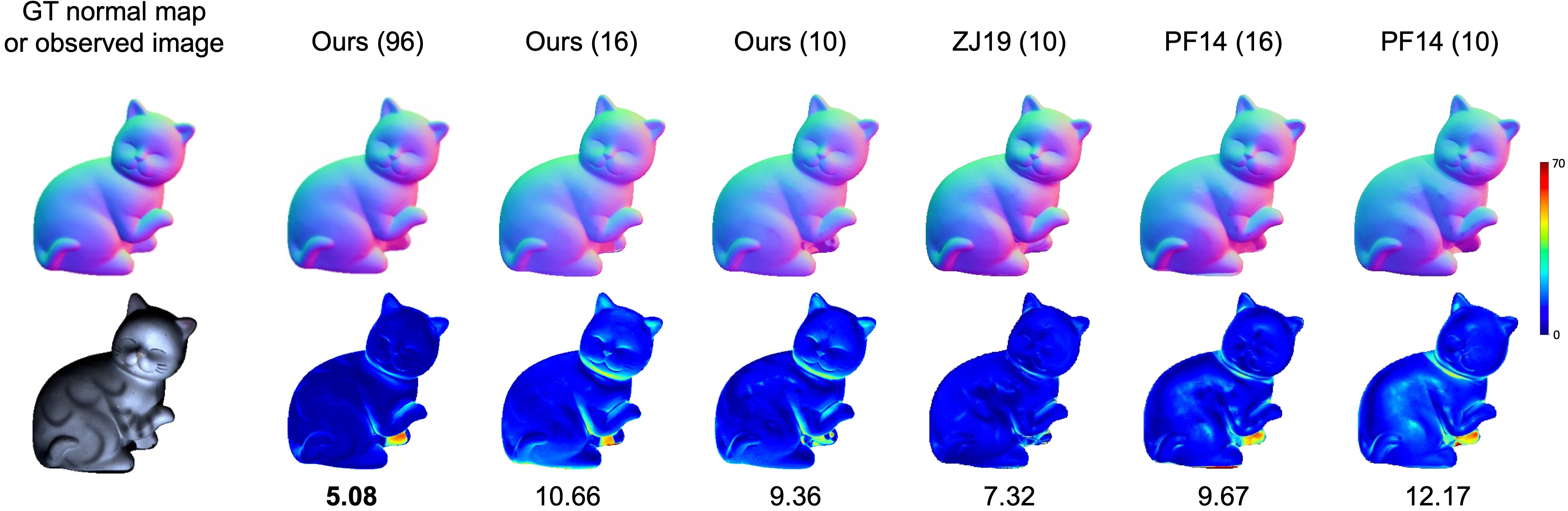}
    \caption{\sc Cat}
\end{subfigure}
\centering
\begin{subfigure}{\linewidth}
    \centering
    \includegraphics[width=\linewidth]{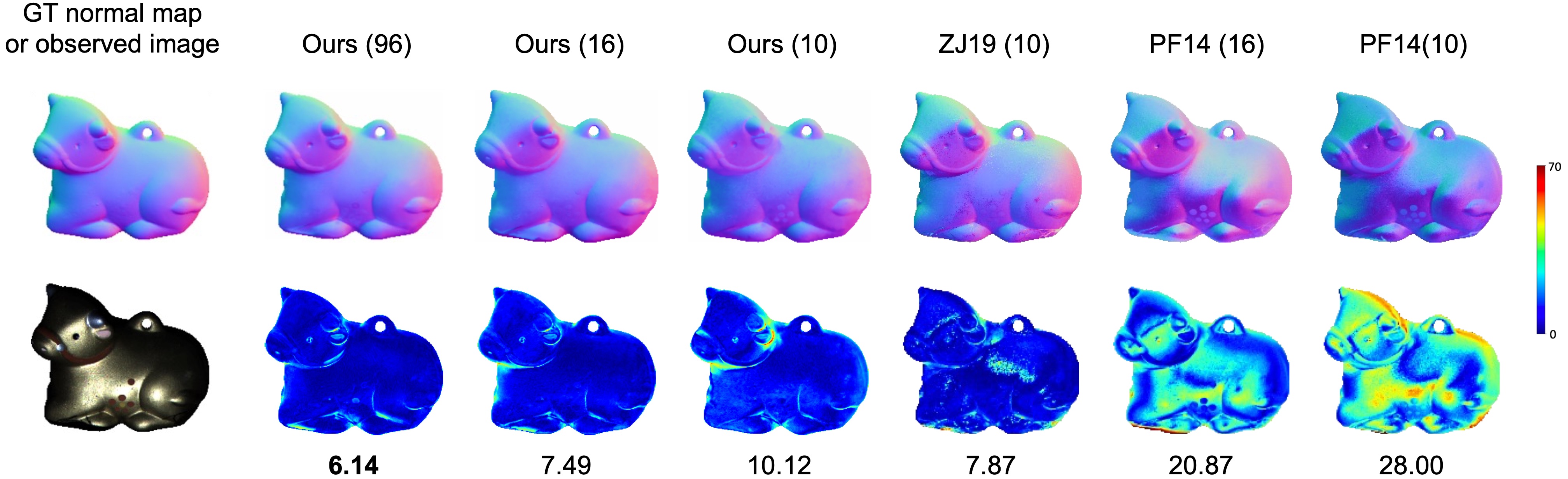}
    \caption{\sc Cow}
\end{subfigure}
\begin{subfigure}{\linewidth}
    \centering
    \includegraphics[width=\linewidth]{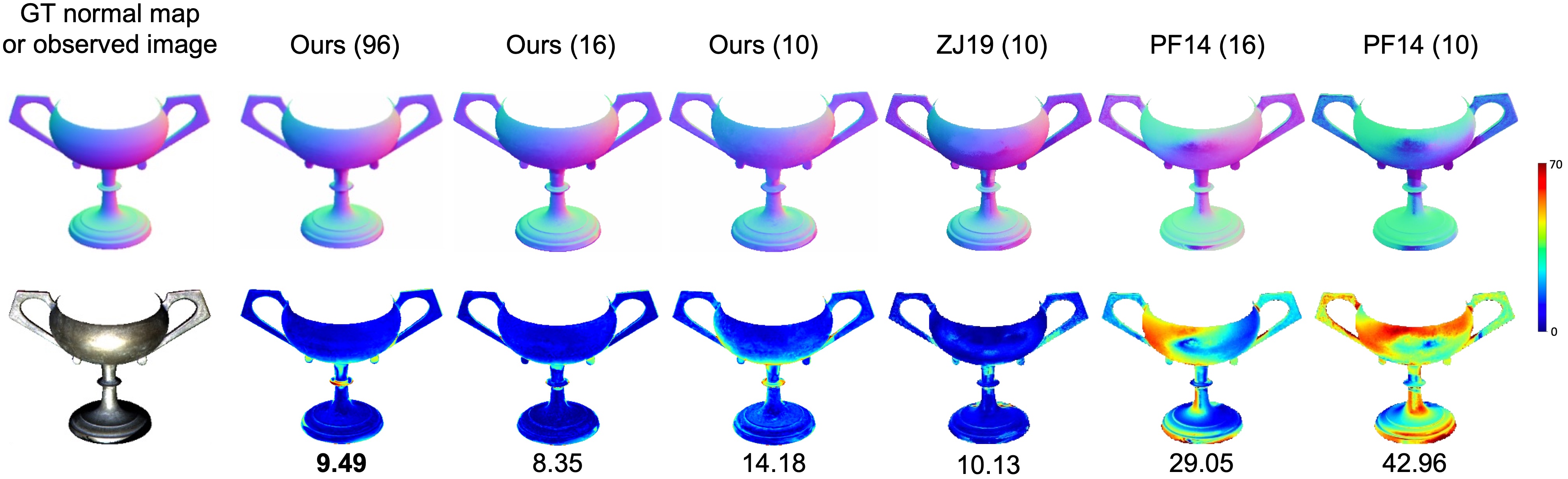}
    \caption{\sc Goblet}
\end{subfigure}
\begin{subfigure}{\linewidth}
    \centering
    \includegraphics[width=\linewidth]{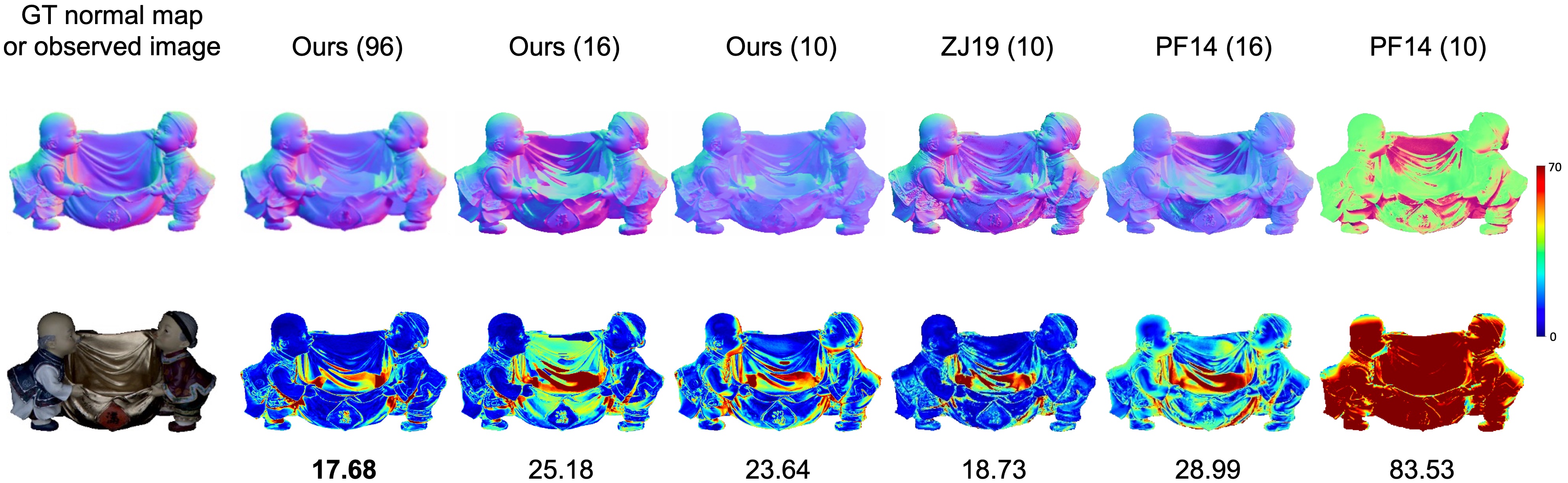}
    \caption{\sc Harvest}
\end{subfigure}
\caption{Error maps and surface normal of {\sc Cat}, {\sc Cow}, {\sc Goblet} and {\sc Harvest}}
\label{fig:sparse_2}
\end{figure}

\begin{figure}[H]
\centering
\begin{subfigure}{\linewidth}
    \centering
    \includegraphics[width=\linewidth]{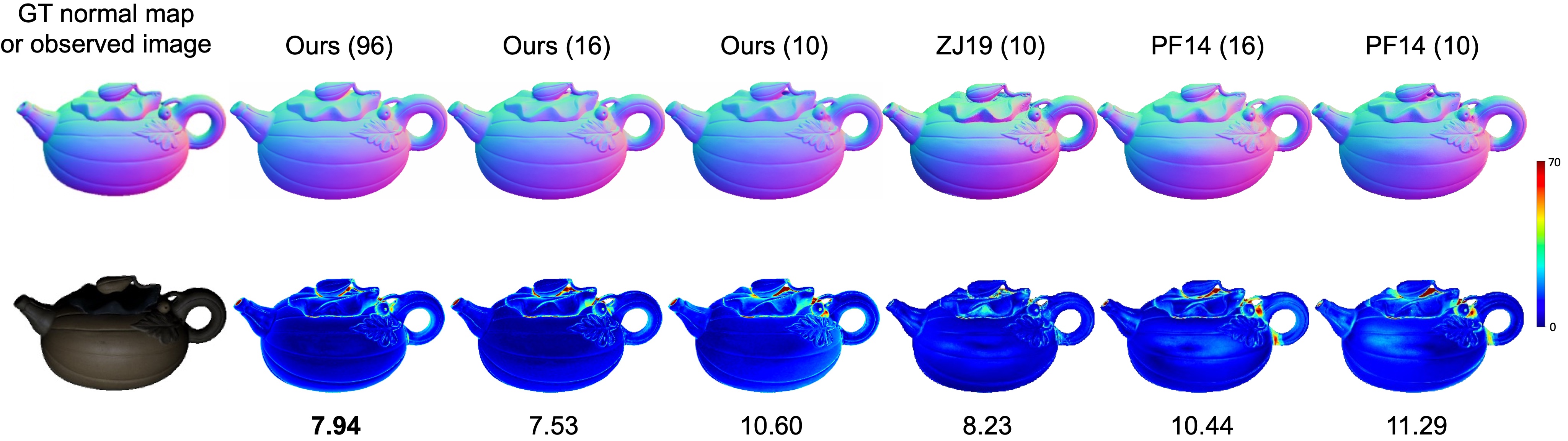}
    \caption{\sc Pot1}
\end{subfigure}
\begin{subfigure}{\linewidth}
    \centering
    \includegraphics[width=\linewidth]{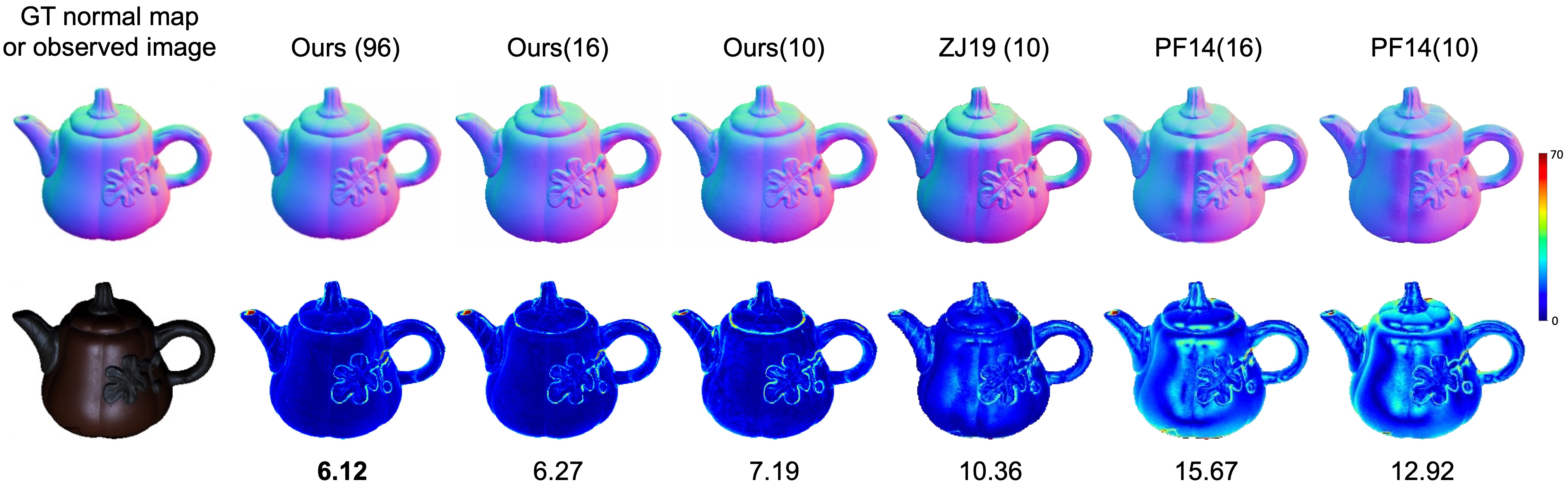}
    \caption{\sc Pot2}
\end{subfigure}
\begin{subfigure}{\linewidth}
    \centering
    \includegraphics[width=\linewidth]{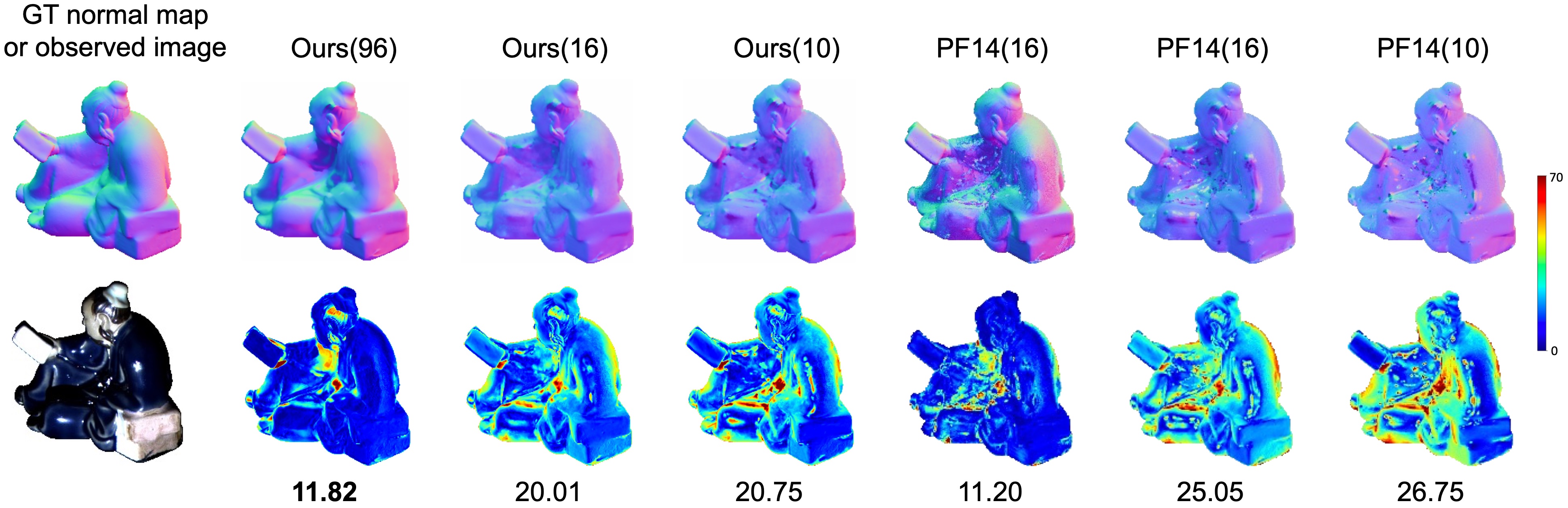}
    \caption{\sc Reading}
\end{subfigure}
\caption{Error maps and surface normal of {\sc Pot1}, {\sc Pot2}, {\sc Reading}}
\label{fig:sparse_3}
\end{figure}

We compared our 10 lights and 16 lights models with our 96 lights models and PF14~\cite{papadhimitri2014closed} under the same light setting, as shown in \Fref{fig:sparse_1}, \Fref{fig:sparse_2}, \Fref{fig:sparse_3}. Note that sparse uncalibrated photometric stereo results are from a set of random lights.

\newpage

\newpage

\newpage
\section{Time Analysis}
\label{computation}
KK21~\cite{kaya2021uncalibrated} did not release their code. Therefore, we report the test time (s) for TM18~\cite{taniai2018neural}, LL22~\cite{li2022neural} and ours in \Tref{tab:test time} under different numbers of input, respectively. Our method has the fastest test time, given 96 images as input and the competitive test time given 16 images as input among the mentioned methods. The tests are implemented on an RTX3090 GPU, and the batch size is full size for lighting and 2048 for spatially sampling. The larger the batch size for spatially sampling, the faster the test, but the GPU memory occupation will also increase. According to our experiments, the parameters we choose are the most cost-efficient, which will occupy around 3.2GB on RTX3090. 

The average training time for our method is about 4 hours on RTX 2080 Ti 12GB GPU, while \cite{taniai2018neural} has an average training time of 1 hour, \cite{kaya2021uncalibrated} has an average training time of 1 hour, and \cite{li2022neural} has an average training time of 6 minutes. The slow training time of our method is mainly because our depth reconstruction algorithm runs on the CPU and joint optimization of LightNet's CNN Encoder in a per-pixel way.

\begin{table}[h]
    \fontsize{6pt}{7pt}\selectfont
    \centering
    \caption{Quantitative comparison in terms of test time (s) on {\sc DiLiGenT} benchmark~\cite{shi2016benchmark}. Numbers in  brackets indicate the number of lights. For example, Ours (96) indicates that our method takes 96 images under different lights as the input.}
    \label{tab:test time}
    \resizebox{\linewidth}{!}{
    \begin{tabular}{c|cccccccccc|c}
    \hline
    & {\sc Ball}  & {\sc Bear}  & {\sc Buddha} & {\sc Cat}   & {\sc Cow}   & {\sc Goblet} & {\sc Harvest} & {\sc Pot1}  & {\sc Pot2}  & {\sc Reading} & AVG \\
    \hline
    TM18 (96)~\cite{taniai2018neural} & \textbf{0.21 } & \textbf{0.70 } & \textbf{0.79 } & 1.02 & \textbf{0.48 } & 1.07 & 1.07 & 1.18 & 0.84 & 0.54  & 0.79 \\
    LL22 (96)~\cite{li2022neural} & 1.22  & 0.92  & 1.17  & 1.36  & 0.85  & 1.26  & 1.23  & \textbf{1.08}  & 1.14  & 1.06  & 1.13  \\
    Ours (96) & 0.29  & \textbf{0.70}  & 0.83  & \textbf{0.82}  & 0.50  & \textbf{0.46}  & \textbf{1.04}  & 1.10  & \textbf{0.65}  & \textbf{0.52}  & \textbf{0.69}  \\
    \hline
    TM18 (16)~\cite{taniai2018neural} & \textbf{0.04 } & \textbf{0.11 } & \textbf{0.12 } & \textbf{0.17 } & \textbf{0.08 } & \textbf{0.22 } & \textbf{0.17 } & \textbf{0.17 } & \textbf{0.14 } & \textbf{0.13 } & \textbf{0.13 } \\
    LL22 (16)~\cite{li2022neural} & 0.62  & 0.69  & 0.69  & 0.68  & 0.67  & 0.66  & 0.71  & 0.66  & 0.69  & 0.77  & 0.68  \\
    Ours (16) & 0.27  & 0.67  & 0.80  & 0.83  & 0.49  & 0.49  & 1.08  & 1.09  & 0.63  & 0.52  & 0.69  \\
    \hline
    \end{tabular}%
    }
\end{table}%

\end{document}